\documentclass{article} 
\usepackage{iclr2025_conference,times}
\iclrfinalcopy

\usepackage{amsmath,amsfonts,bm}

\def\eqref#1{equation~\ref{#1}}

\def\1{\bm{1}}

\DeclareMathAlphabet{\mathsfit}{\encodingdefault}{\sfdefault}{m}{sl}
\SetMathAlphabet{\mathsfit}{bold}{\encodingdefault}{\sfdefault}{bx}{n}

\usepackage{graphicx}
\usepackage{amsmath}
\usepackage{amssymb}
\usepackage{booktabs}
\usepackage{bbm}
\usepackage{caption}
\usepackage{subcaption}
\usepackage{wrapfig}

\usepackage{hyperref}
\usepackage{url}

\title{Unveiling the Potential of Superexpressive Networks in Implicit Neural Representations}

\author{Uvini Balasuriya Mudiyanselage\\
SCAI, Arizona State University\\
\texttt{ubalasur@asu.edu}
\And 
Woojin Cho\\
TelePIX\\
\texttt{woojin@telepix.net}
\And Minju Jo\\
LG CNS\\
\texttt{minnju42@gmail.com}
\And Noseong Park\\
KAIST\\
\texttt{noseong@kaist.ac.kr}
\And Kookjin Lee\\
SCAI, Arizona State University\\
\texttt{kookjin.lee@asu.edu}
}

\begin{document}

\maketitle

\begin{abstract}
  In this study, we examine the potential of one of the ``superexpressive'' networks in the context of learning neural functions for representing complex signals and performing machine learning downstream tasks. Our focus is on evaluating their performance on computer vision and scientific machine learning tasks including signal representation/inverse problems and solutions of partial differential equations. Through an empirical investigation in various benchmark tasks, we demonstrate that superexpressive networks, as proposed by [Zhang et al. \textit{NeurIPS}, 2022], which employ a specialized network structure characterized by having an additional dimension, namely width, depth, and ``height'', can surpass recent implicit neural representations that use highly-specialized nonlinear activation functions. 
\end{abstract}

\section{Introduction}
\label{sec:intro}
Established on the universal approximation theorem \citep{hornik1989multilayer},  multi-layer perceptrons (MLPs)  (or, fully connected feed-forward neural networks) \citep{cybenko1989approximation,hastie2009elements} have long been served as a foundational component of modern deep learning architectures. These include recurrent neural networks (e.g., long short-term memory~\citep{hochreiter1997long}), graph neural networks~\citep{bronstein2021geometric}, residual networks~\citep{he2016deep}, and Transformers~\citep{vaswani2017attention,devlin2019bert,dosovitskiy2020image}, to name a few. Despite their demonstrated effectiveness, ongoing research continues to explore and develop new architectures that can outperform MLPs.

There have been some efforts to make MLPs more expressive; these approaches consider \textit{standard} MLPs, but bring superior expressive power by leveraging either \textit{nonstandard} nonlinear activation functions \citep{yarotsky2021elementary,shen2021neural} (which is based on the existence of ``superexpressive'' activation~\citep{maiorov1999lower,yarotsky2021elementary}) or \textit{nonstandard} network architectures with the \textit{standard} rectified linear unit (ReLU) activation function~\citep{zhang2022neural}.

In the field of implicit neural representations (INRs)~\citep{sitzmann2020implicit,tancik2020fourier}, the predominant choice for the base architecture has been MLPs, aligning with the alternative term ``coordinate-based MLPs." Recent advancements have centered on developing novel nonlinear activation functions to enhance the ability of INRs to capture high-frequency components of the target signal (i.e., details of the signals) \citep{sitzmann2020implicit,fathony2020multiplicative,ramasinghe2022beyond,saragadam2023wire}. These innovations have demonstrated improved expressivity and capabilities in many signal representation tasks and  computer vision downstream tasks.

In this study, we are interested in employing one of those superexperessive networks to perform tasks that are typically used to test INRs' expressivity and capabilities. To the best of our knowledge, there has been little to no effort in  investigating the practical usage of those superexpressive networks in the context of INRs. Specifically, we focus on MLPs with non-standard network architectures, but with the standard ReLU activation functions. 
This choice is made because MLPs with nonstandard nonlinear activation functions share a similar philosophy with the current literature of INRs, both utilizing specialized functions for nonlinear activation. Instead, we assess the performance of MLPs with the standard ReLU activation, but with nonstandard architecture, which makes the resulting nonlinear activation to be equipped with  additional ``learnable'' components, making resulting INRs to be more expressive and capable. 

\section{Method}
\label{sec:tech}
MLPs can be typically characterized by hyperparameters defining the width and the depth of the architecture and  standard nonlinear activation functions. To further improve expressivity of neural networks, recent studies further seek novel superexpressive architectures. In this study, we focus on the practical applications of the second type of superexpressive networks, i.e., the nonstandard network architectures, in the realm of INRs 
as the first type of superexpressive networks shares the similar underlying philosophy, i.e., developing novel activation functions, with the recent INR literature. We focus on expressivity of ReLU networks, but with special nested structures, allowing flexible representation of nonlienar activation functions. 

\paragraph{Nested Networks}
As opposed to standard two-dimensional MLPs with width and depth, an MLP with a nonstandard architecture, characterized by a nested  structure \citep{zhang2022neural}, introduces a three-dimensional feed-forward network architecture by introducing an additional dimension, ``height''. Neural networks with this new architecture is denoted as nested networks (NestNets) because hidden neurons of a NestNet of height $s$ are activated by a NestNet of height $s-1$. For a NestNet with $s=1$, it degenerates to a standard MLP. Hereinafter, a NestNet of height $s$ is denoted as NestNet($s$). 

\begin{figure}[t] 
    \centering
    \vspace{-3em} 

    \begin{subfigure}{0.55\textwidth}
        \centering
        \includegraphics[width=\linewidth]{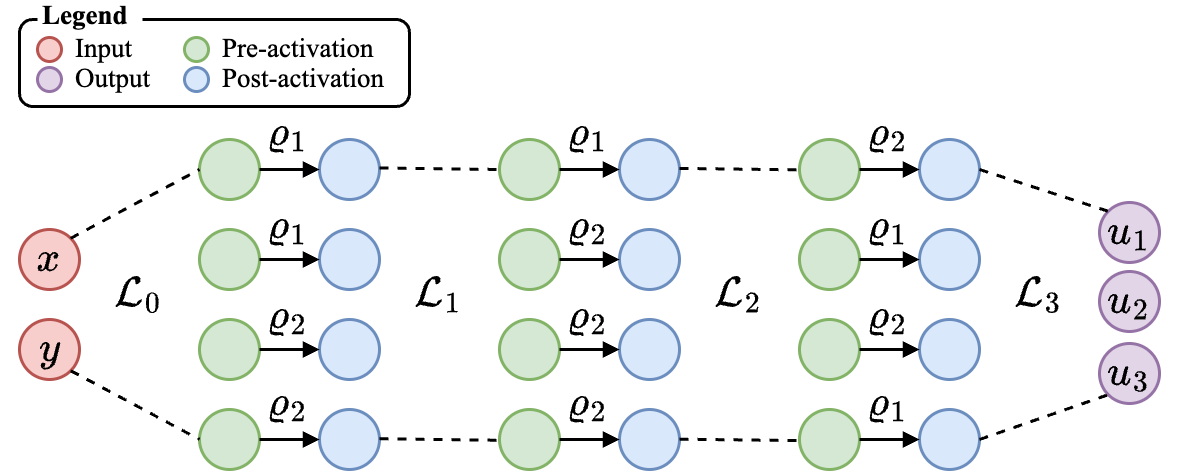}
        \caption{}
        \label{fig:archi}
    \end{subfigure}
    \hfill
    \begin{subfigure}{0.4\textwidth}
        \centering
        \includegraphics[width=\linewidth]{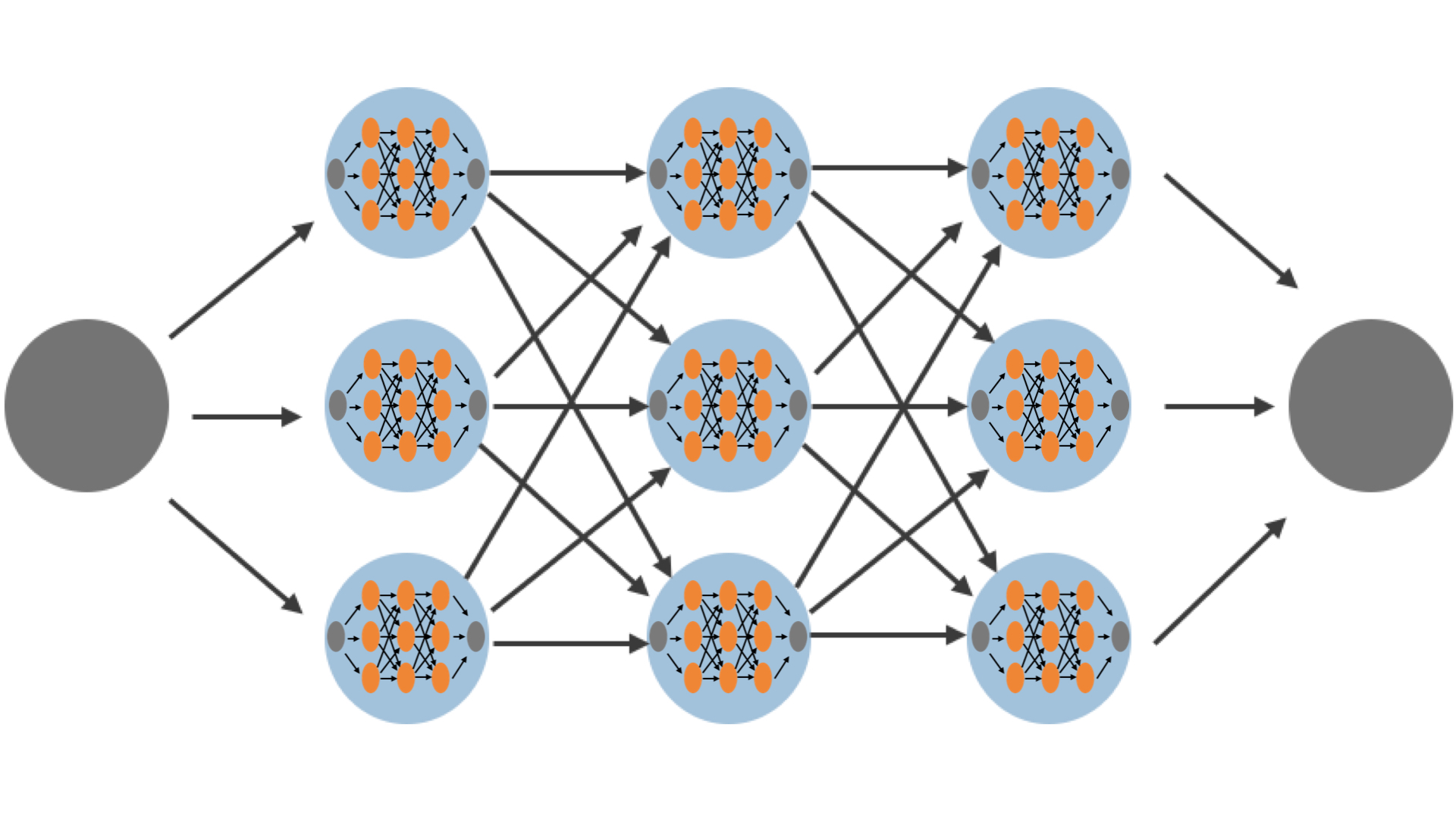}
        \caption{}
        \label{fig:archi_b}
    \end{subfigure}

    \vspace{-0.5em} 
    \caption{\textbf{A NestNet of height~2.}
    (a)~The input and output of the network are coordinates 
    $\pmb{x}=(x,y)$ and the signal at that coordinate 
    $\pmb{u}(\pmb{x})$. Two subnetworks 
    $\varrho_1$ and~$\varrho_2$ (which are regular MLPs) 
    serve as nonlinear activations.
    (b)~A high-level illustration of a NestNet with 
width = depth = 3 and height = 2. Each blue node represents a 
regular MLP, used as a learnable activation 
function applied element-wise to pre-activations in the main network.}
    \label{fig:1}
\end{figure}

Following the notations in the original paper \citet{zhang2022neural}, each hidden neuron of NestNet($s$) is activated by one of the $r$ subnetworks, $\varrho_1, \cdots, \varrho_r$, where each $\varrho_i: \mathbb{R} \mapsto \mathbb{R}$ is a trainable functions and applied to each neuron individually (i.e., element-wise activation). Figure~\ref{fig:archi} illustrates an instantiation of a NestNet of height 2 with two subnetworks $\varrho_1$ and $\varrho_2$. In Figure, $\mathcal L_i$ indicates affine transformation, which transforms the previous hidden representation as in standard MLPs such that $\mathcal L_i(h) = \mathbf{W}_i h + \mathbf{b}_i$, where $\mathbf{W}_i$ and $\mathbf{b}_i$ denote weight and bias of the $i$th layer, respectively. Then the pre-activation is activated by NestNets(1), $\varrho_1$ and $\varrho_2$.

\paragraph{Fourier Feature Mapping} In the context of INRs, we formulate NestNets to take coordinate information ($\pmb x = (x,y)$ in this illustration) and to output the signal at that coordinate, $u_1(x,y)$, $u_2(x,y)$, and $u_3(x,y)$ (i.e., multi-channel signal such as RGB channels for color images). Moreover, following the typical structure of the recent INRs~\citep{rahimi2007random,tancik2020fourier}, we add a layer at the beginning of NestNets to convert the coordinates to Fourier features  (i.e., sinusoidal signals) such that:
$[\alpha_1\cos(2\pi \pmb \beta_1^{\mathsf T} \pmb x), \alpha_1\sin(2\pi \pmb \beta_1^{\mathsf T} \pmb x), \alpha_2\cos(2\pi \pmb \beta_2^{\mathsf T} \pmb x), \alpha_2\sin(2\pi \pmb \beta_2^{\mathsf T} \pmb x), \ldots]$, 
where $\alpha_i$ and $\pmb \beta_i$ denote Fourier series coefficients and Fourier basis frequencies.

\paragraph{Training objectives}\label{sec:training_obj} We consider two cases, where the ground-truth target signal exists and the ground-truth target signal is implicitly matched via a set of equations. In the former case, the loss is set as a point-wise $\ell_2$-distance, $\frac{1}{|\mathcal D|}\sum_{i\in\mathcal D} \left(\pmb g(\pmb x_i) - \pmb f_{\theta}(\pmb x_i)\right)^2$, where $\pmb g(\pmb x)$ denotes the target signal. The later case is exemplified with a method of approximating solutions of PDEs, which is also known as the formalism of physics-informed neural networks (PINNs), and the loss is defined as $\sum_{i\in\mathcal D_{\text{IC}}} \left(\pmb g(\pmb x_i) - \pmb f_{\theta}(\pmb x_i)\right)^2 + \sum_{i\in\mathcal D_{\text{BC}}} \left(\pmb g(\pmb x_i) - \pmb f_{\theta}(\pmb x_i)\right)^2 + \sum_{i\in\mathcal D_{\text{PDE}}} \left(\mathcal R(\pmb x_i, \pmb f_{\theta}(\pmb x_i), \nabla_{\pmb x} \pmb f_{\theta}(\pmb x_i),\nabla_{\pmb x}^2(\pmb x_i) \pmb f_{\theta}(\pmb x_i),\ldots)\right)^2$; that is, the first and the second term aim to minimizing the errors in initial and boundary conditions and the last term aims to minimize the errors in the physical laws defined by the differential equations~\citep{raissi2019physics}.

\section{Experiments}
\label{sec:exp}
To demonstrate the performance of NestNets in the context of INR applications, we follow experimental procedures shown in \citet{saragadam2023wire} that test a neural network's capability of (1) representing a signal, (2) solving inverse problems in computer vision tasks, and (3) solving PDEs.  
To this end, we base our implementation on the work 
of \citet{saragadam2023wire}. The code is written in \textsc{PyTorch} \citep{paszke2019pytorch}. 
For all tasks, we repeat the same experiments for five varying random seeds and report the best results among those. 

\paragraph{NestNet specification.} For all experiments, we consider NestNets of height 2; that is, the regular MLP structure with 2 hidden layers and 256 neurons is activated by a subnetwork (i.e., a NestNet of height 1); the subnetwork $\varrho(\cdot)$  applied in an element-wise fashion. Following the original reference~\citet{zhang2022neural}, we set the subnetwork to be a shallow MLP: 
\begin{equation}\label{eq:nonlin}
    \varrho(h) = w_2^{\intercal} \text{ReLU} ( w_1 h + b_1 ) + b_2,
\end{equation}
where $w_1, w_2, b_1 \in \mathbb{R}^3$ and $b_2$ are learnable parameters and initialized as $w_1 = [1,1,1], w_2=[1,1,-1], b_1 = [-0.2, -0.1, 0.0]$, and $b_2=0$.
For the Fourier features, we set $\alpha_i=1$ and $\beta_i=i$, which simply degenerates back to positional encoding used in NeRF \citep{mildenhall2021nerf}.

\paragraph{Baselines.} As baselines of comparisons, we consider the following models: WIREs~\citep{saragadam2023wire}, SIRENs~\citep{sitzmann2020implicit}, Gaussian~\citep{ramasinghe2022beyond}, MFNs~\citep{fathony2020multiplicative}, and FFNs~\citep{tancik2020fourier}. Unless otherwise specified, we consider each neural network with 2 hidden layers and 256 neurons to be consistent with NestNets. In case of WIREs, we follow the approach in \citet{saragadam2023wire}, which reduces  the number of hidden neurons by $\sqrt{2}$ to account for the doubling due to having real and imaginary parts.

\paragraph{Training.} 
For all tasks, training is done via minimizing the losses (either the direct matching or the implicit matching) defined in Section~\ref{sec:training_obj} with gradient-descent type algorithms, specifically the Adam optimizer \citep{DBLP:journals/corr/KingmaB14}. No additional regularization terms are included. 

\begin{table*}[!h]
\vspace{-4mm}
  \centering
  \setlength{\tabcolsep}{1pt} 
  \begin{tabular}{ccccccc}
    Ground Truth & NestNet & WIRE & SIREN & Gaussian & MFN & FFN \\
    \includegraphics[width=0.12\textwidth]{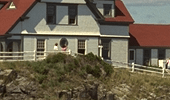} & 
    \includegraphics[width=0.12\textwidth]{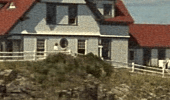}
    &
    \includegraphics[width=0.12\textwidth]{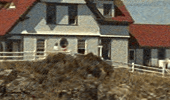}
    &
    \includegraphics[width=0.12\textwidth]{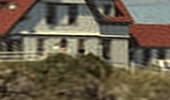}
    & 
    \includegraphics[width=0.12\textwidth]{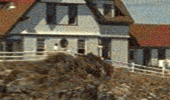}
    & 
    \includegraphics[width=0.12\textwidth]{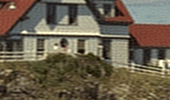}
    & 
    \includegraphics[width=0.12\textwidth]{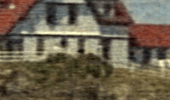}\\
    \includegraphics[width=0.13\textwidth]{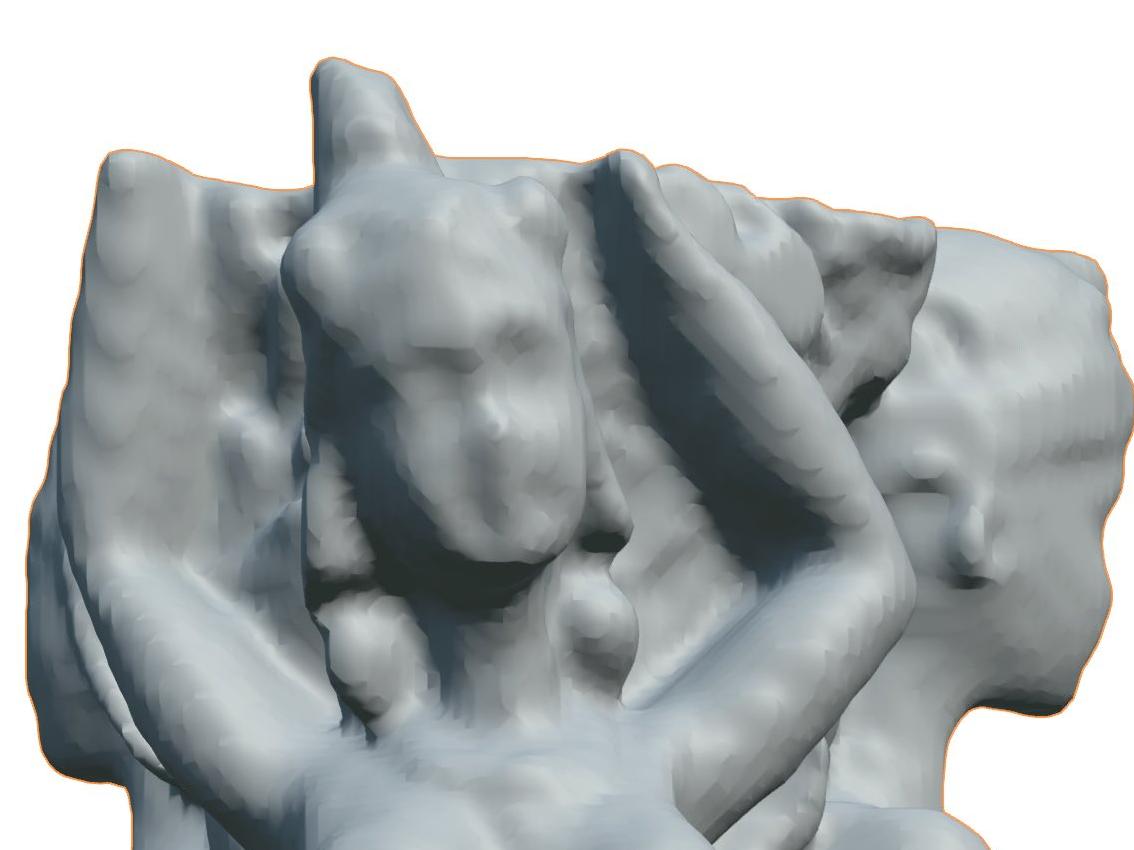}
     & 
    \includegraphics[width=0.13\textwidth]{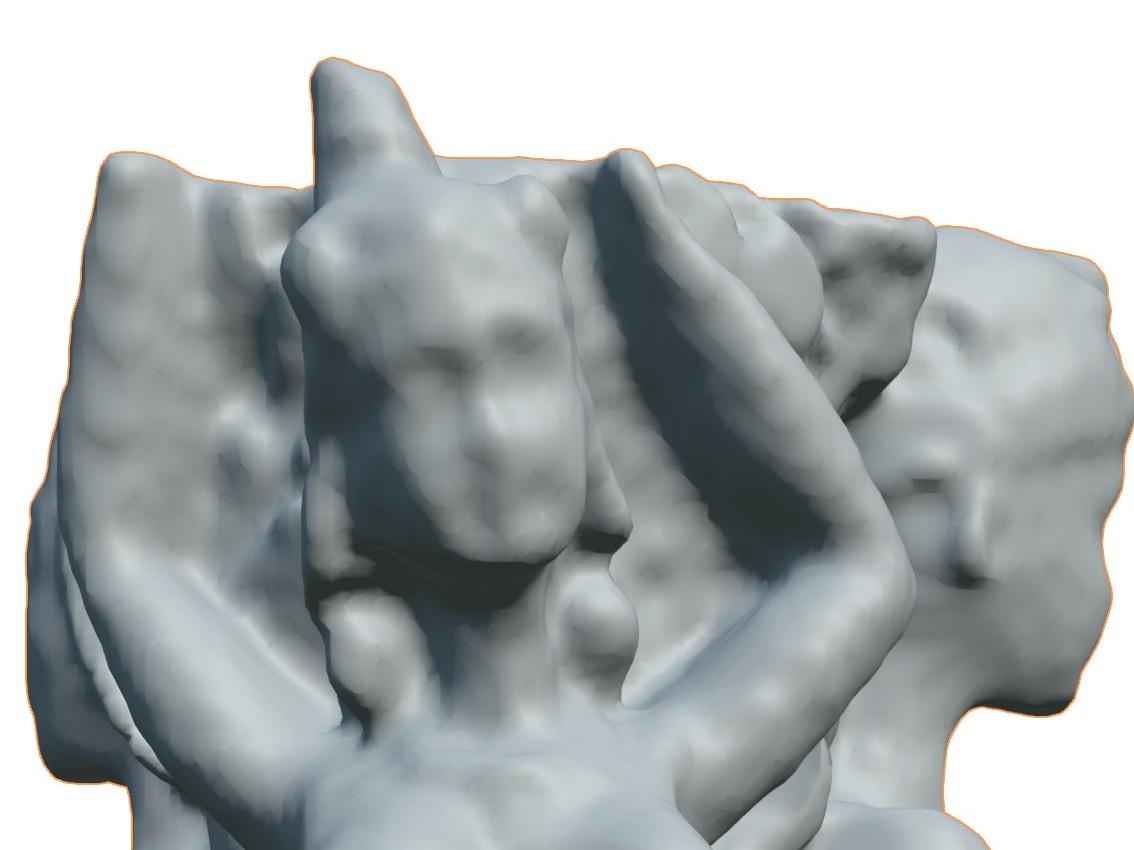}
     &
    \includegraphics[width=0.13\textwidth]{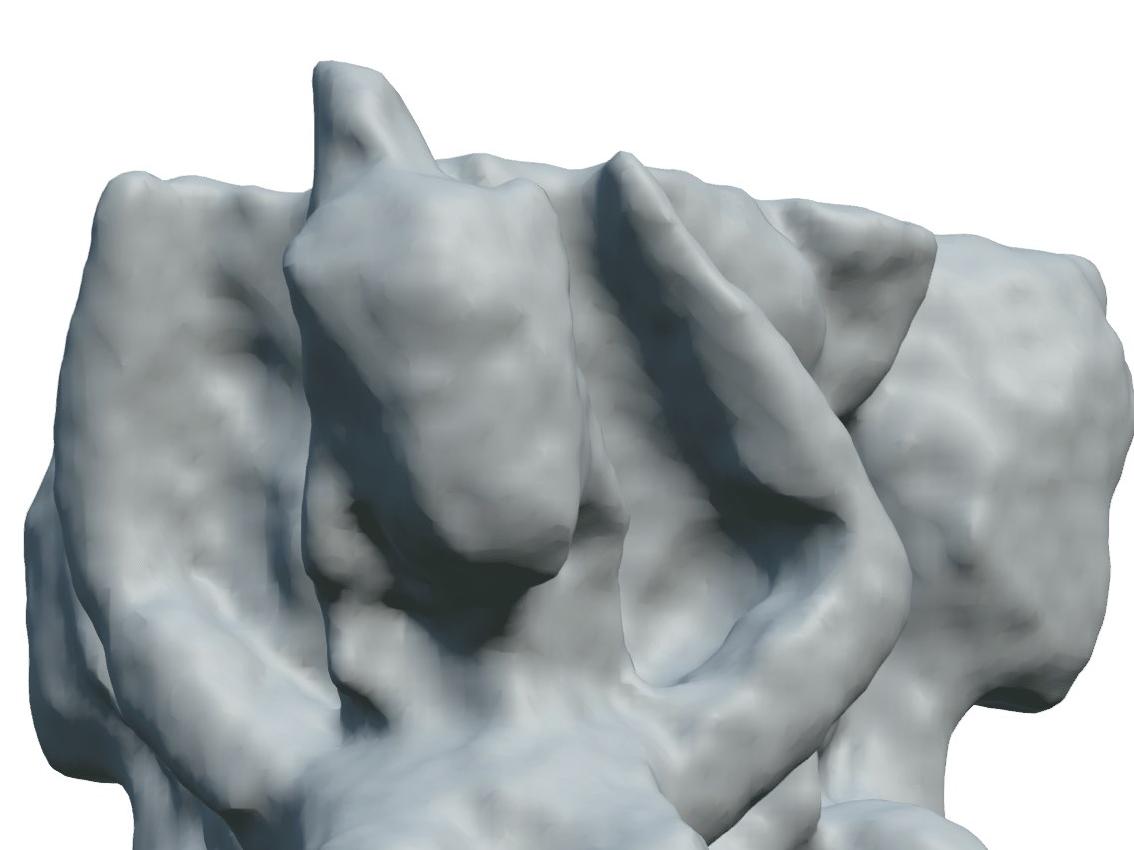}
     &
    \includegraphics[width=0.13\textwidth]{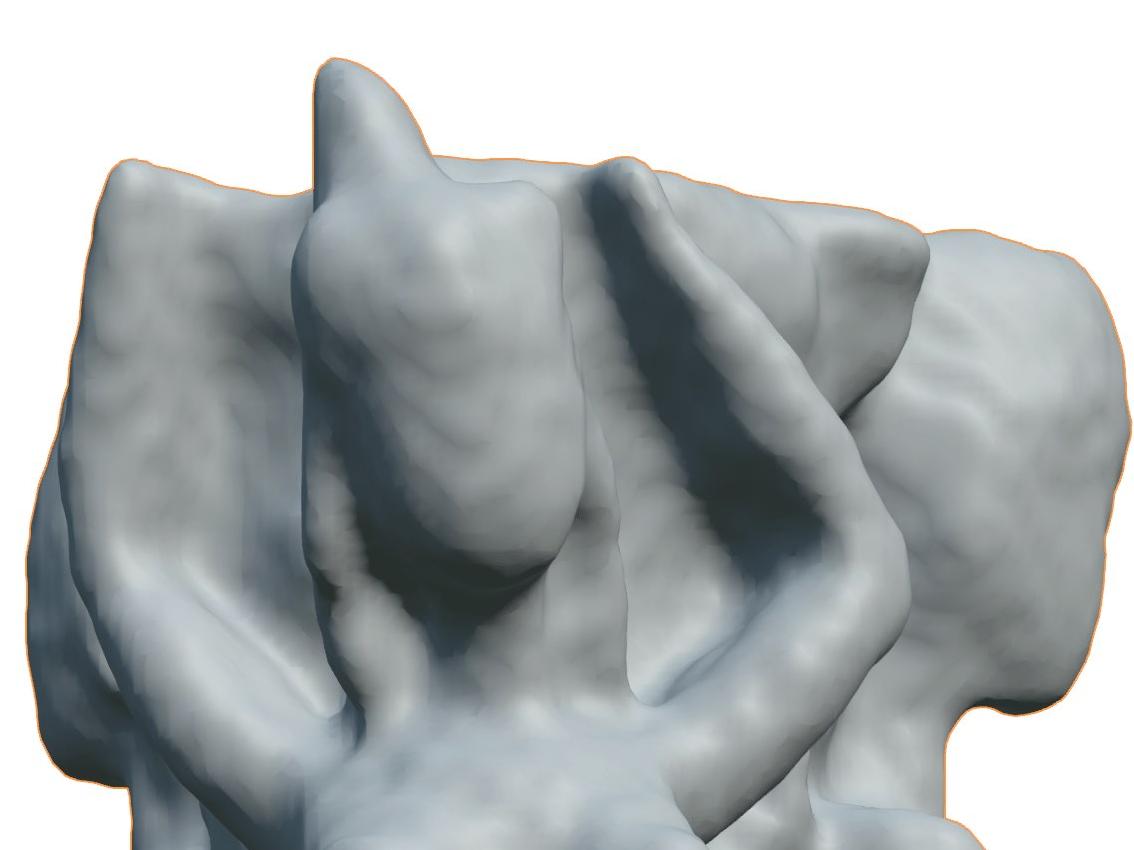} &
    \includegraphics[width=0.13\textwidth]{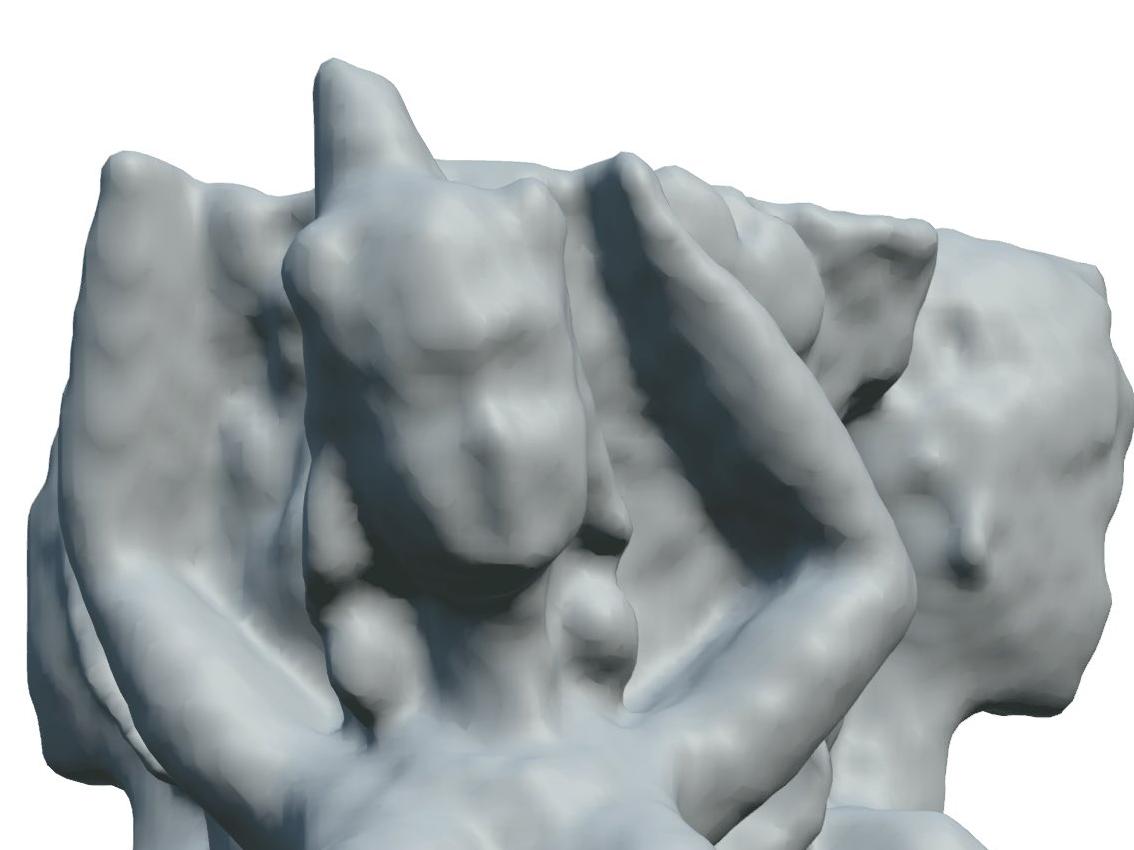}
     & 
    \includegraphics[width=0.13\textwidth]{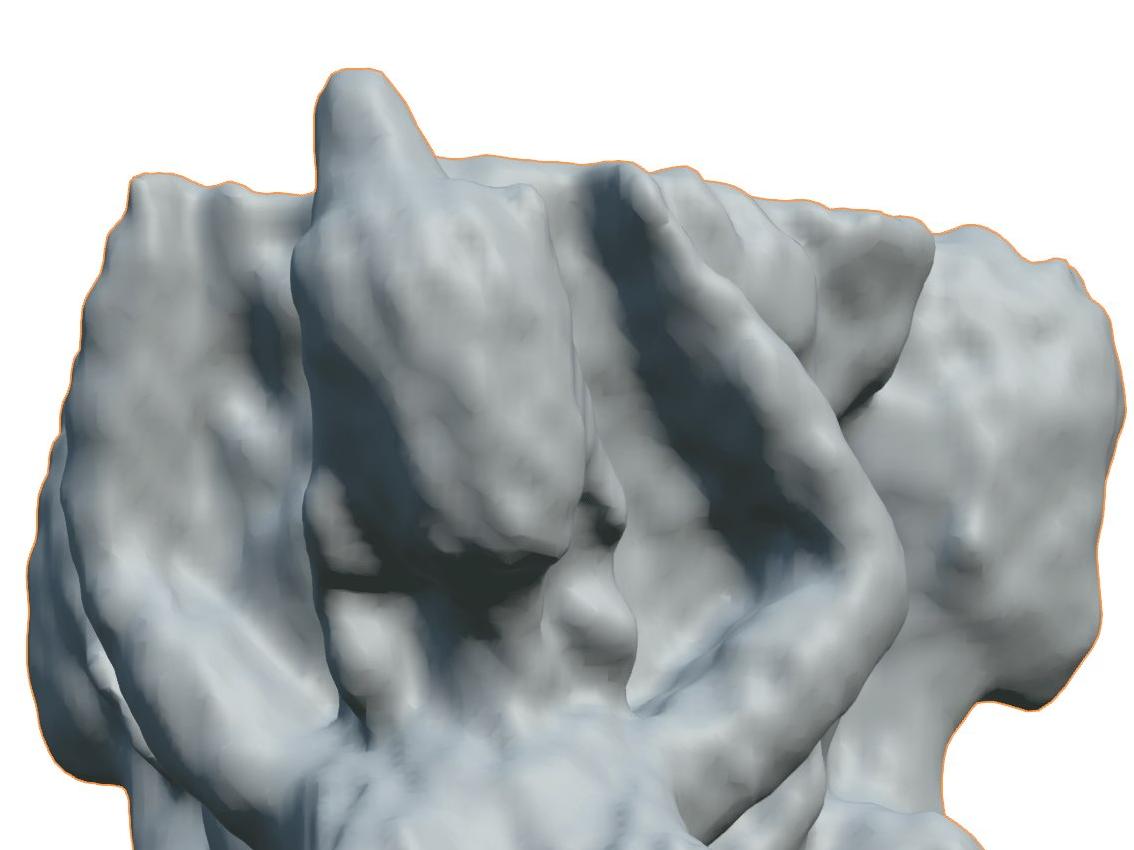}
     & 
    \includegraphics[width=0.13\textwidth]{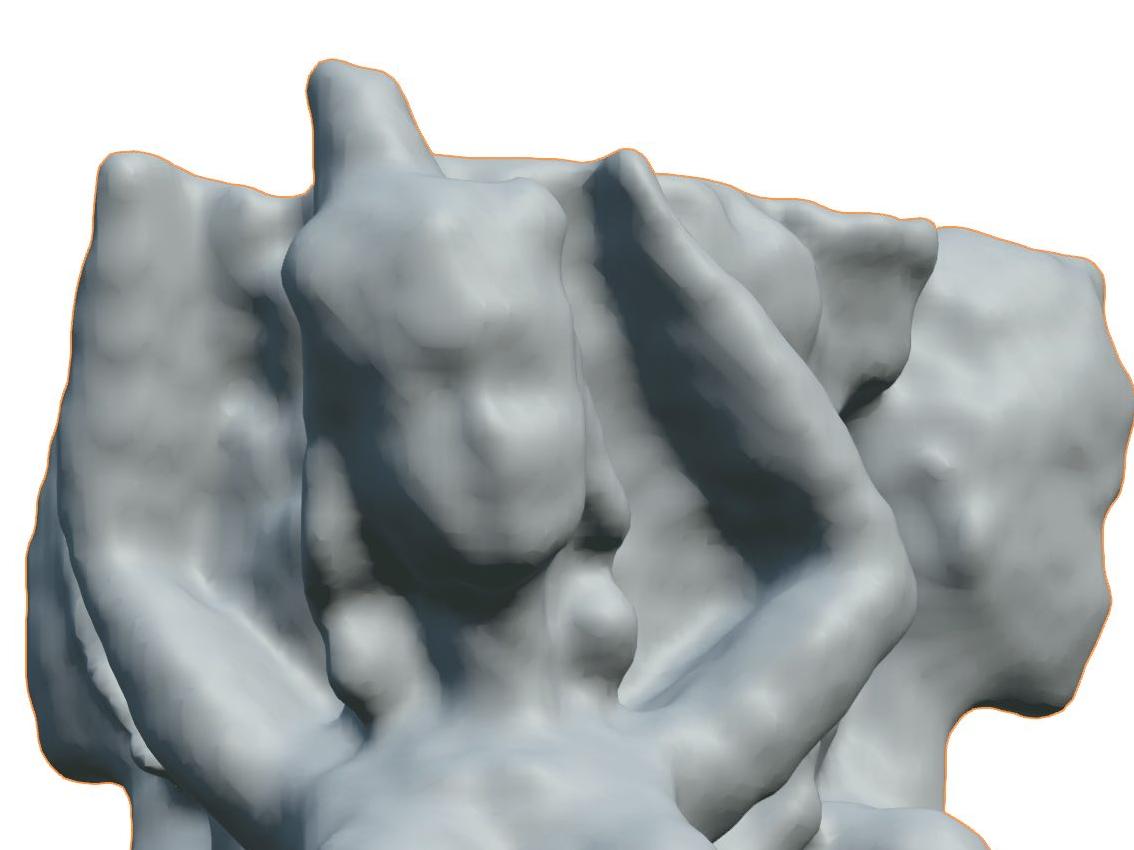}\\
    \includegraphics[width=0.13\textwidth]{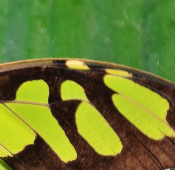}
     & 
    \includegraphics[width=0.13\textwidth]{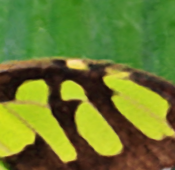}
     &
    \includegraphics[width=0.13\textwidth]{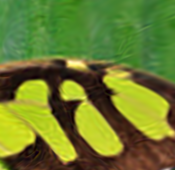} 
    & 
    \includegraphics[width=0.13\textwidth]{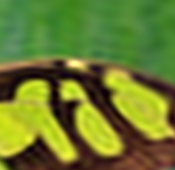}
     & 
    \includegraphics[width=0.13\textwidth]{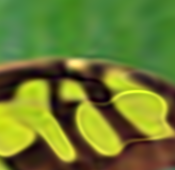}
     & 
    \includegraphics[width=0.13\textwidth]{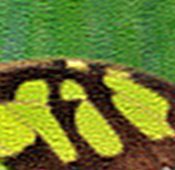}
    & 
    \includegraphics[width=0.13\textwidth]{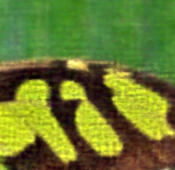}\\
    \includegraphics[width=0.13\textwidth]{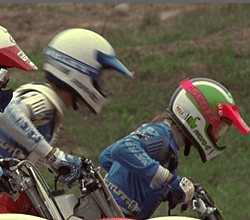}   & 
    \includegraphics[width=0.13\textwidth]{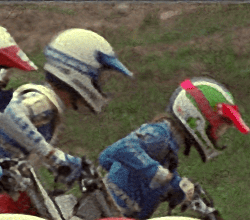} &
    \includegraphics[width=0.13\textwidth]{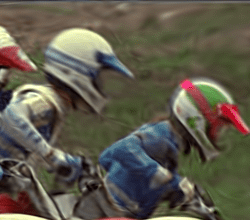} 
    &\includegraphics[width=0.13\textwidth]{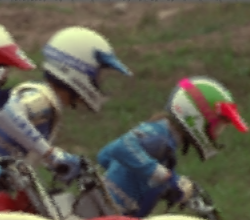} &
    \includegraphics[width=0.13\textwidth]{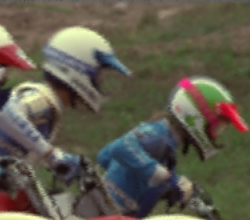} &
    \includegraphics[width=0.13\textwidth]{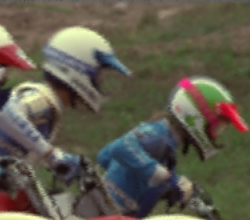} &
    \includegraphics[width=0.13\textwidth]{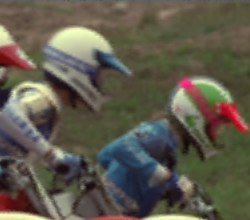} \\
    \includegraphics[width=0.13\textwidth]{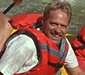}   & 
    \includegraphics[width=0.13\textwidth]{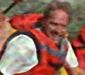} &
    \includegraphics[width=0.13\textwidth]{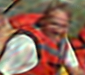} 
    &
    \includegraphics[width=0.13\textwidth]{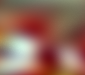} &
    \includegraphics[width=0.13\textwidth]{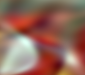} &
    \includegraphics[width=0.13\textwidth]{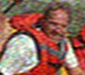} &
    \includegraphics[width=0.13\textwidth]{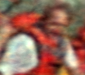} \\
    \includegraphics[width=0.12\textwidth]{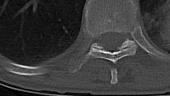} & 
    \includegraphics[width=0.12\textwidth]{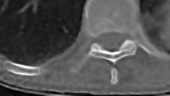} &
    \includegraphics[width=0.12\textwidth]{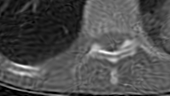} &
    \includegraphics[width=0.12\textwidth]{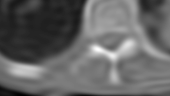} &
    \includegraphics[width=0.12\textwidth]{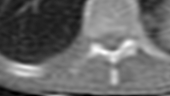} &
    \includegraphics[width=0.12\textwidth]{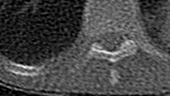} &
    \includegraphics[width=0.12\textwidth]{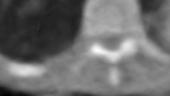}
  \end{tabular}
  \vspace{-1mm}
  \captionof{figure}{[Computer Vision tasks] (From top to bottom rows) image representation, occupancy volume presentation, single-image super resolution, multi-image super resolution, image-denoising, and CT reconstruction. NestNets produce consistently better results qualitatively as well as quantitatively (PSNR, SSIM, and IOU reported in Appendix~\ref{app:results}).} 
  \label{tab:occupancy_volume_rep}
\end{table*}


\subsection{Computer Vision Downstream Tasks}
We first test the NestNet architecture on a set of computer vision downstreams tasks, where the usage of INRs is considered to be beneficial: Image representation, occupancy volumes representation, and inverse problems including (single-image/multi-image) super resolution, image denoising, and computed tomography (CT) reconstruction.  The results are summarized in Figure~\ref{tab:occupancy_volume_rep}, which shows a segment of all considered images and 3d-rendering of volume occupancy (in the second row), highlighting the improved performance in each tasks. We evaluate the performance of models in  peak signal-to-noise ratio (PSNR) and structural similarity (SSIM)~\citep{wang2004image}. For learning point cloud occupancy volumes, intersection over union (IOU) is measured. Although not reported in the main body due to the page limit, NestNet outperforms in each tasks in terms of either PSNR, SSIM, and IOU. We provide the full description of the problem setup and additional results reporting PSNR, SSIM, and IOU in Appendices. Overall, NestNet consistently outperform baselines, suggesting improved performance for conducting computer vision downstream tasks using neural functional representations.
Additionally, in Appendix~\ref{app:img_rep}, we provide training trajectories of all methods, depict learned  activations over training epochs, and in Appendix~\ref{app:extra_exps}, we provide results for varying learning rates and super-resolution experimentation with varying downsampling rates.

\subsection{Physics-informed Neural Networks}
The task of solving PDEs with the PINNs formalism is to minimize the errors in the implicit problem formulation, $\mathcal R$. To compare the performance of different INRs, we choose a canonical benchmark problem \citep{krishnapriyan2021characterizing}: 1D convection equations, which are defined as the implicit formulation: $\mathcal R(u) = u_t + \beta u_x=0$, where $u_t$ and $u_x$ are the temporal and spatial derivatives of the solution function $u$. The convective term $\beta$ is set as 10.

\begin{figure}[h]
    \centering
    \begin{subfigure}{0.16\columnwidth}
    \includegraphics[width=1\columnwidth]{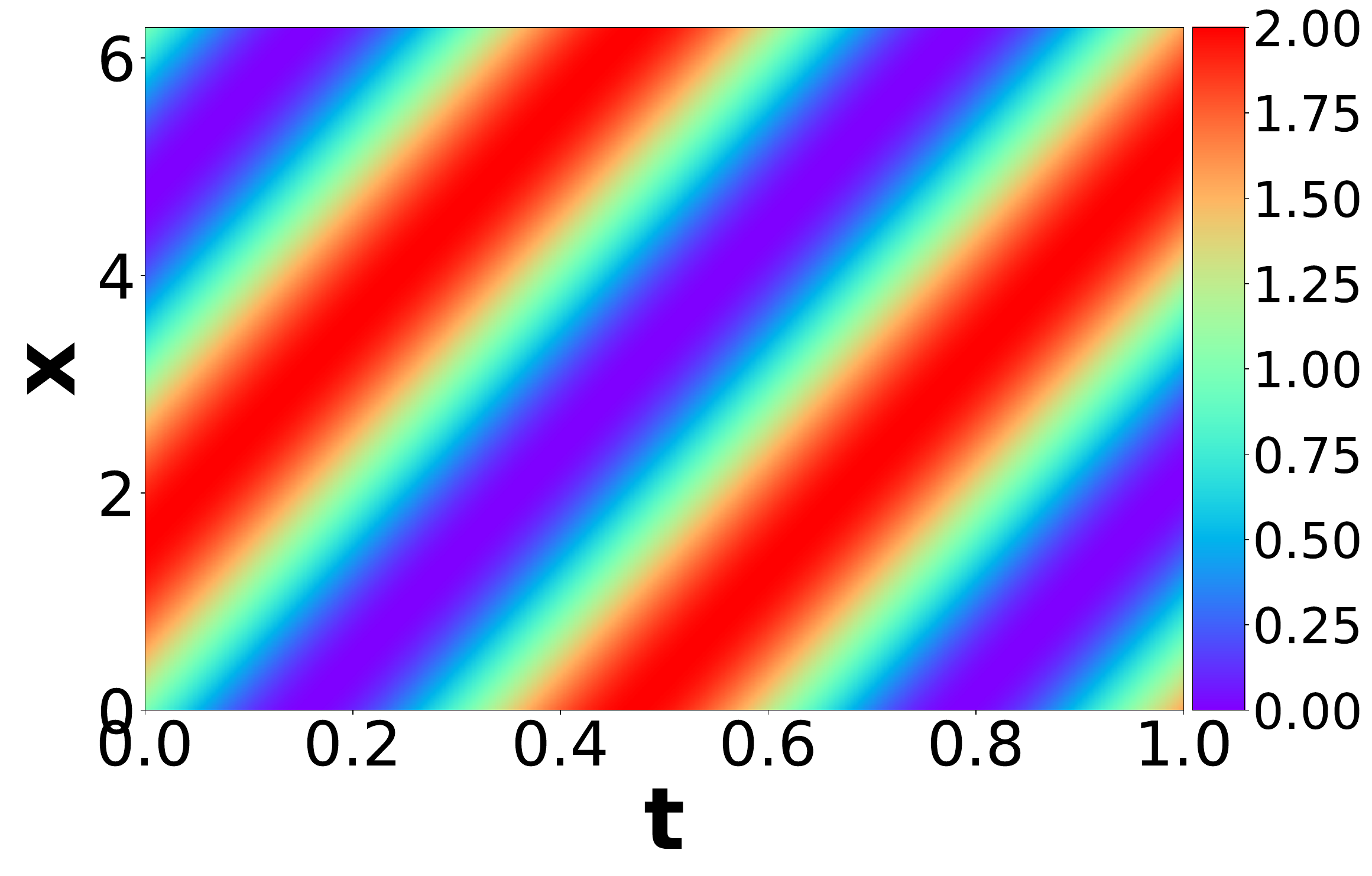}
    \caption{Ground Truth}
    \end{subfigure} 
    \begin{subfigure}{0.16\columnwidth}
    \includegraphics[width=1\columnwidth]{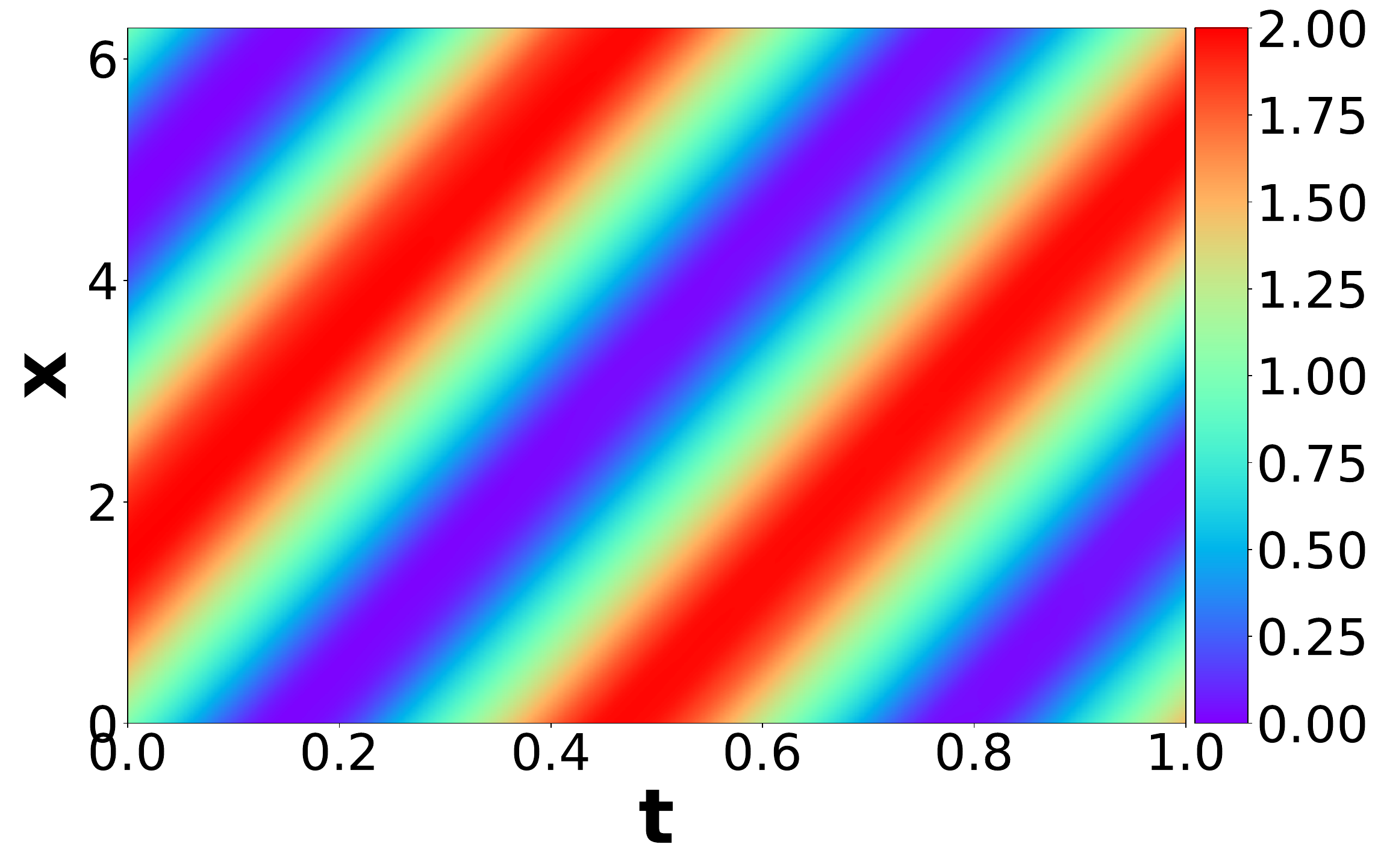}
    \caption{NestNet}
    \end{subfigure}
    \begin{subfigure}{0.16\columnwidth}
    \includegraphics[width=1\columnwidth]{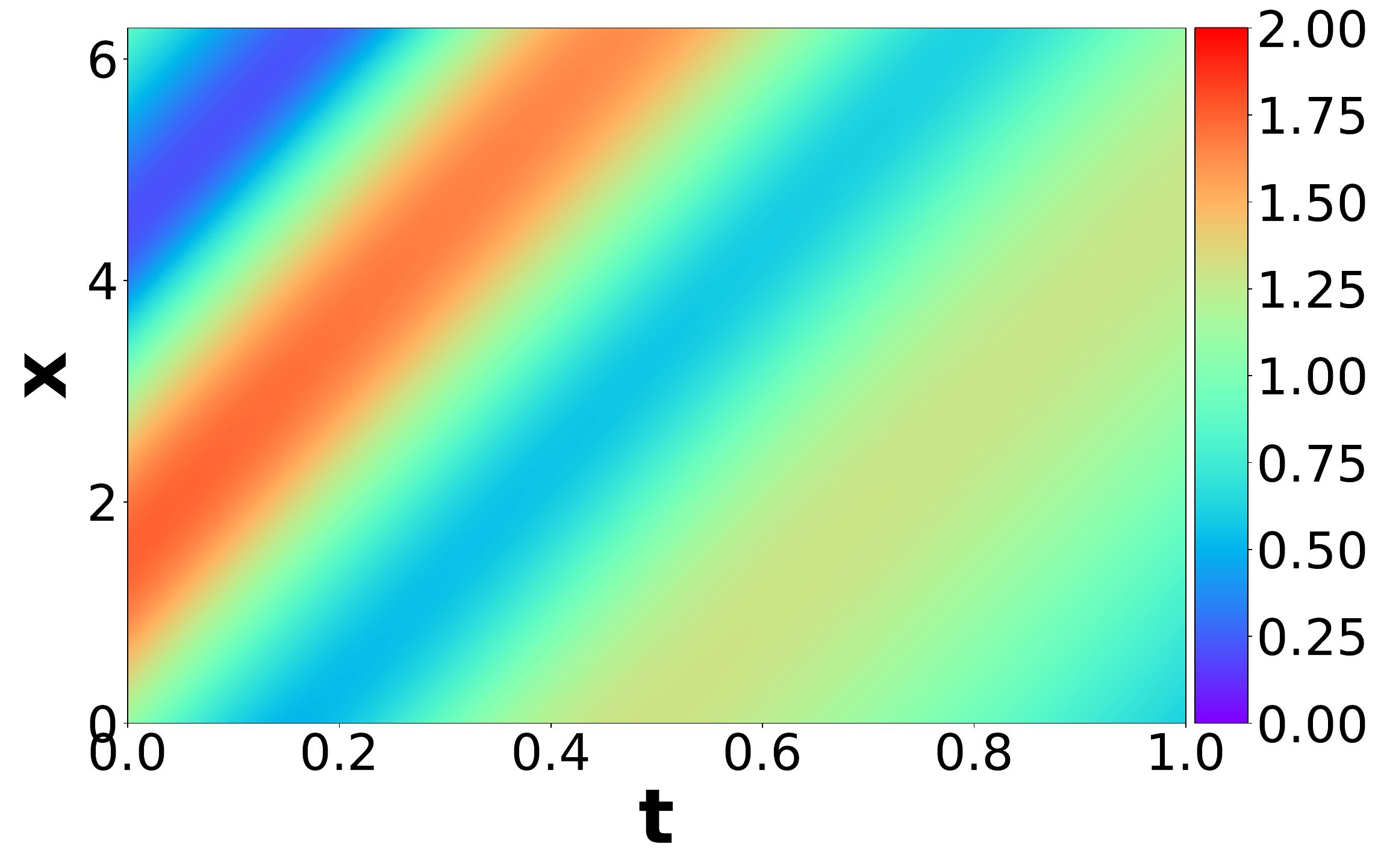}
    \caption{PINN (MLP)}
    \end{subfigure}
    \begin{subfigure}{0.16\columnwidth}
    \includegraphics[width=1\columnwidth]{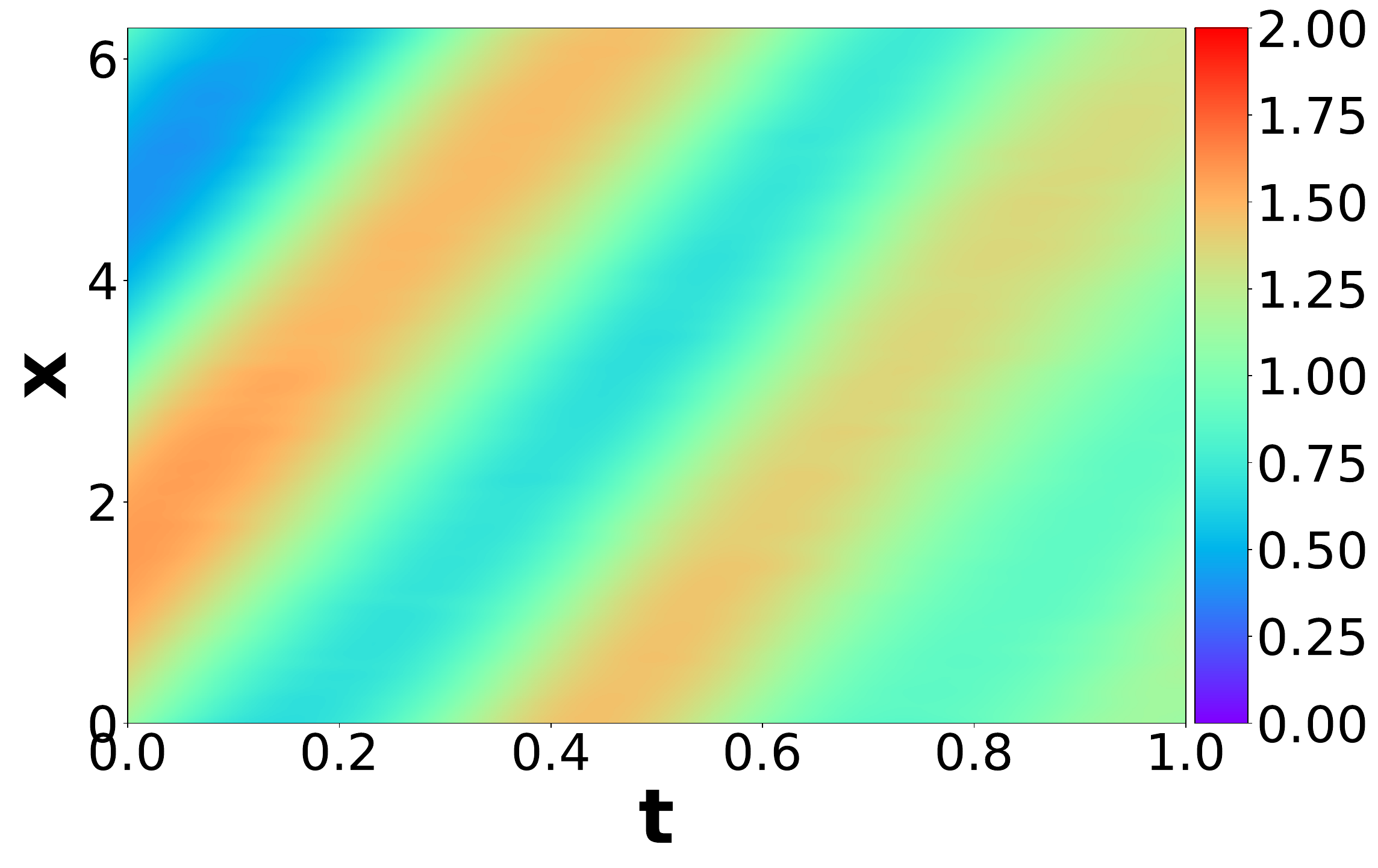}
    \caption{SIREN}
    \end{subfigure}
    \begin{subfigure}{0.16\columnwidth}
    \includegraphics[width=1\columnwidth]{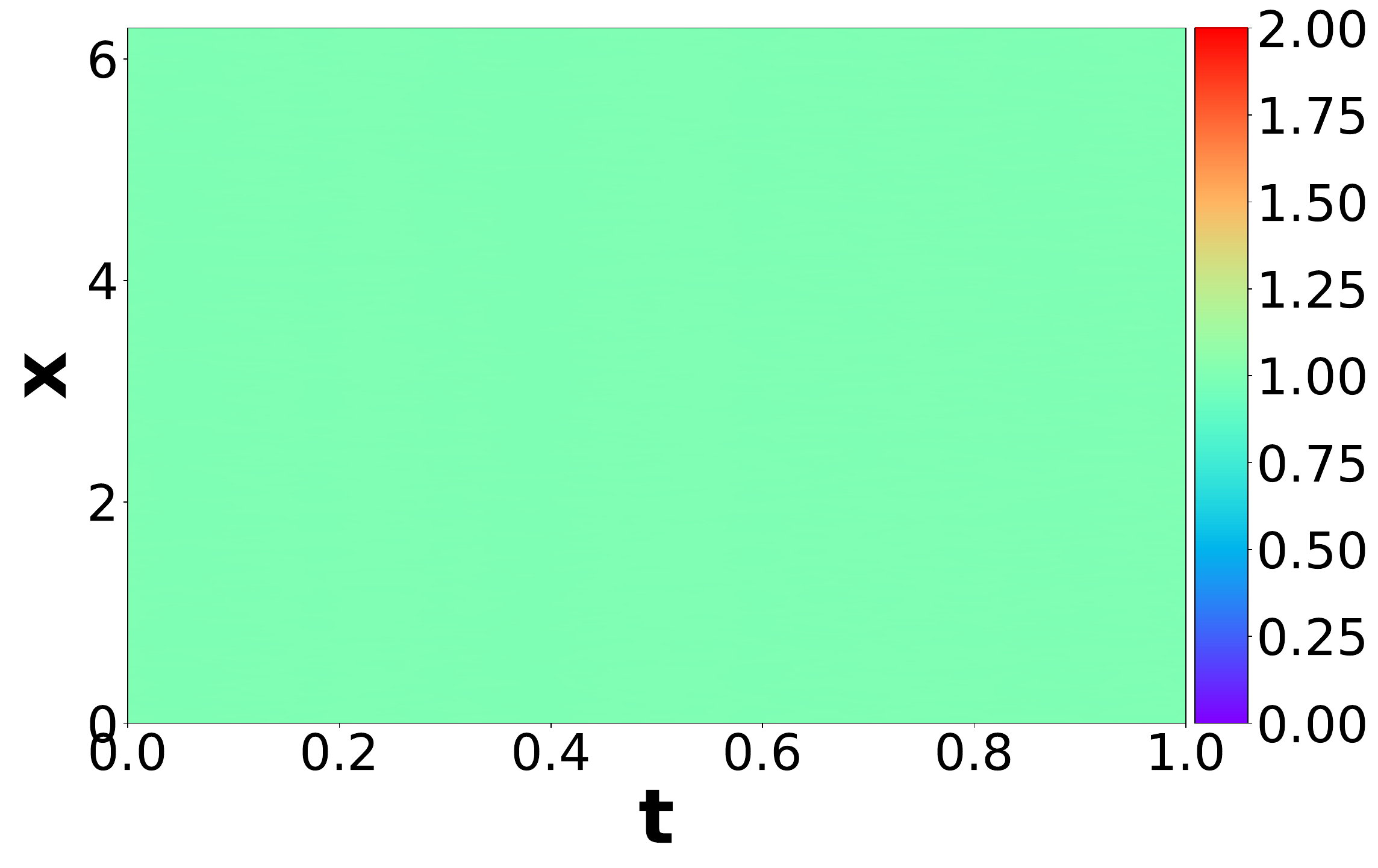}
    \caption{FFN}
    \end{subfigure} 
    \begin{subfigure}{0.16\columnwidth}
    \includegraphics[width=1\columnwidth]{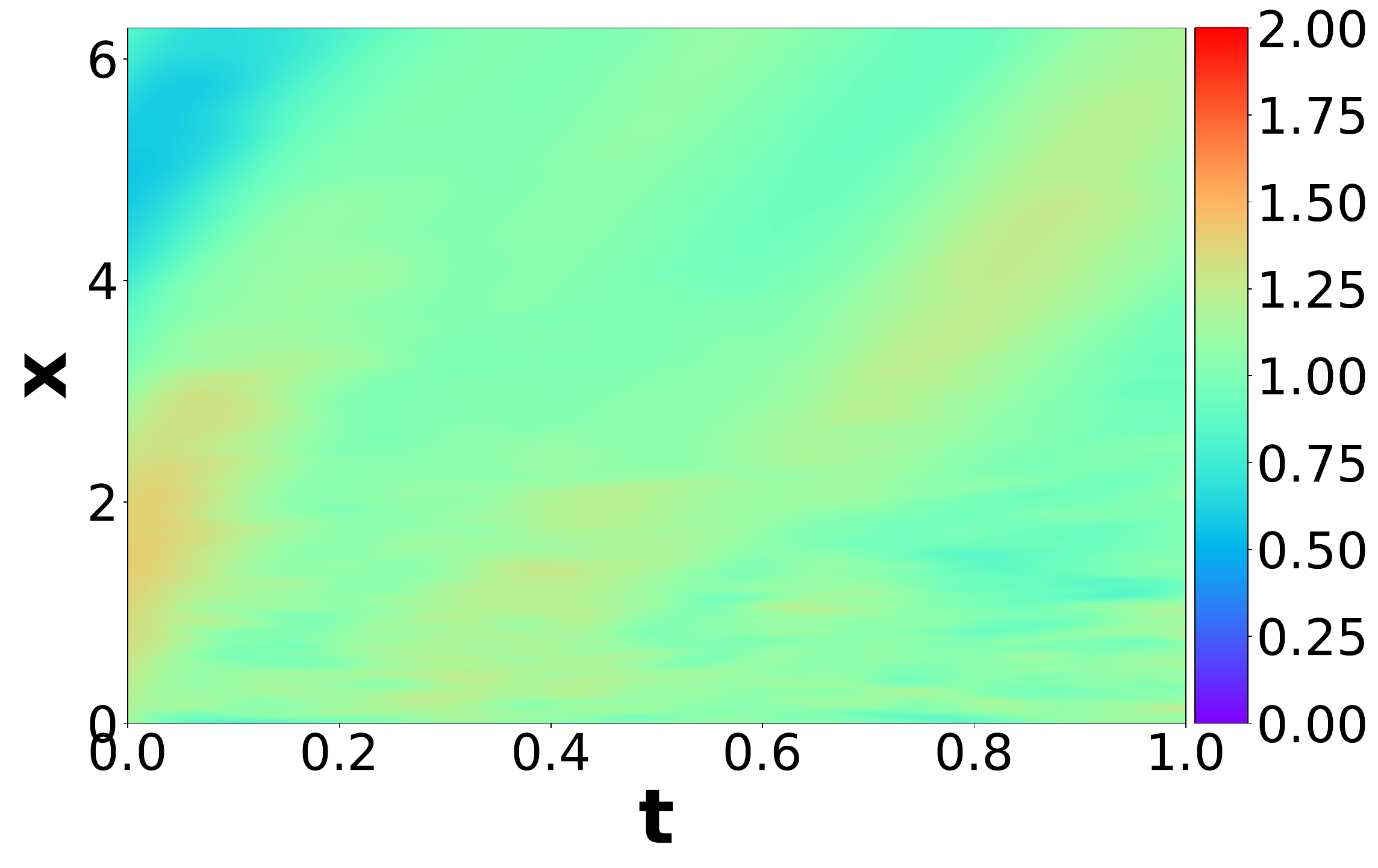}
    \caption{WIRE}
    \end{subfigure}
    \\
    \vspace{-1mm}
    \caption{[PINNs] The solution snapshots of the 1D convection equation ($\beta=10$). The relative errors for (NestNet, MLP, SIREN, FFN, WIRE) are (0.0342, 0.3719, 0.4031, 0.5918, 0.5525).}
    \label{fig:pinn_sol}
\end{figure}

Figure~\ref{fig:pinn_sol} shows the solutions of the equation depicted on a ($t$,$x$)-plane ($t$ in the horizontal and $x$ in the vertical axes) approximated by varying INRs. The solution approximated by using NestNet is evidently shown to be much more accurate than those of the baselines. While acknowledging the existence of advanced PINNs algorithms~ \citep{lau2023pinnacle,de2023operator,cho2024separable,cho2024hypernetwork,cho2024parameterized}, we focus our comparisons to standard PINNs with varying INRs as we compare the expressivity of each INR; any such INRs can be combined with those advanced algorithms and we leave this to future work. 


\vspace{-2mm}
\vspace{-2mm}
\section{Conclusion}
\label{sec:conc}
We have explored the potential of NestNets, \textit{super-expressive} networks realized by invoking non-standard architectural design, for learning neural functions to represent complex signals. Through extensive experiments on critical benchmark tasks for evaluating implicit neural representations (INRs), we demonstrate that NestNets exhibit superior expressivity and performance compared to state-of-the-art INRs. The results highlight that
the non-standard nested structure of NestNets allows for the learning of nonlinear activation functions with greater flexibility, enabling the representation of more intricate functions than those achievable with conventional nonlinear activation functions.

\section{Acknowledgment}
K. Lee acknowledges support from the U.S. National Science Foundation under grant IIS 2338909. K. Lee also acknowledges Research Computing at Arizona State University for providing HPC resources that have contributed to the partial research results reported within this paper.


\bibliography{./iclr2025_conference}

\begin{thebibliography}{37}
\providecommand{\natexlab}[1]{#1}
\providecommand{\url}[1]{\texttt{#1}}
\expandafter\ifx\csname urlstyle\endcsname\relax
  \providecommand{\doi}[1]{doi: #1}\else
  \providecommand{\doi}{doi: \begingroup \urlstyle{rm}\Url}\fi

\bibitem[Agustsson \& Timofte(2017)Agustsson and Timofte]{agustsson2017ntire}
Eirikur Agustsson and Radu Timofte.
\newblock {NTIRE} 2017 challenge on single image super-resolution: {D}ataset and study.
\newblock In \emph{Proceedings of the IEEE conference on computer vision and pattern recognition workshops}, pp.\  126--135, 2017.

\bibitem[Bronstein et~al.(2021)Bronstein, Bruna, Cohen, and Veli{\v{c}}kovi{\'c}]{bronstein2021geometric}
Michael~M Bronstein, Joan Bruna, Taco Cohen, and Petar Veli{\v{c}}kovi{\'c}.
\newblock Geometric deep learning: {G}rids, groups, graphs, geodesics, and gauges.
\newblock \emph{arXiv preprint arXiv:2104.13478}, 2021.

\bibitem[Cho et~al.(2024{\natexlab{a}})Cho, Nam, Yang, Yun, Hong, and Park]{cho2024separable}
Junwoo Cho, Seungtae Nam, Hyunmo Yang, Seok-Bae Yun, Youngjoon Hong, and Eunbyung Park.
\newblock Separable physics-informed neural networks.
\newblock \emph{Advances in Neural Information Processing Systems}, 36, 2024{\natexlab{a}}.

\bibitem[Cho et~al.(2024{\natexlab{b}})Cho, Jo, Lim, Lee, Lee, Hong, and Park]{cho2024parameterized}
Woojin Cho, Minju Jo, Haksoo Lim, Kookjin Lee, Dongeun Lee, Sanghyun Hong, and Noseong Park.
\newblock Parameterized physics-informed neural networks for parameterized pdes.
\newblock In \emph{Forty-first International Conference on Machine Learning}, 2024{\natexlab{b}}.

\bibitem[Cho et~al.(2024{\natexlab{c}})Cho, Lee, Rim, and Park]{cho2024hypernetwork}
Woojin Cho, Kookjin Lee, Donsub Rim, and Noseong Park.
\newblock Hypernetwork-based meta-learning for low-rank physics-informed neural networks.
\newblock \emph{Advances in Neural Information Processing Systems}, 36, 2024{\natexlab{c}}.

\bibitem[Clark et~al.(2013)Clark, Vendt, Smith, Freymann, Kirby, Koppel, Moore, Phillips, Maffitt, Pringle, et~al.]{clark2013cancer}
Kenneth Clark, Bruce Vendt, Kirk Smith, John Freymann, Justin Kirby, Paul Koppel, Stephen Moore, Stanley Phillips, David Maffitt, Michael Pringle, et~al.
\newblock The cancer imaging archive ({TCIA}): {M}aintaining and operating a public information repository.
\newblock \emph{Journal of digital imaging}, 26:\penalty0 1045--1057, 2013.

\bibitem[Cybenko(1989)]{cybenko1989approximation}
George Cybenko.
\newblock Approximation by superpositions of a sigmoidal function.
\newblock \emph{Mathematics of control, signals and systems}, 2\penalty0 (4):\penalty0 303--314, 1989.

\bibitem[De~Ryck et~al.(2023)De~Ryck, Bonnet, Mishra, and de~Bezenac]{de2023operator}
Tim De~Ryck, Florent Bonnet, Siddhartha Mishra, and Emmanuel de~Bezenac.
\newblock An operator preconditioning perspective on training in physics-informed machine learning.
\newblock In \emph{The Twelfth International Conference on Learning Representations}, 2023.

\bibitem[Devlin et~al.(2019)Devlin, Chang, Lee, and Toutanova]{devlin2019bert}
Jacob Devlin, Ming-Wei Chang, Kenton Lee, and Kristina Toutanova.
\newblock Bert: {P}re-training of deep bidirectional transformers for language understanding.
\newblock In \emph{Proceedings of the 2019 Conference of the North American Chapter of the Association for Computational Linguistics: Human Language Technologies, Volume 1 (Long and Short Papers)}, pp.\  4171--4186, 2019.

\bibitem[Dosovitskiy et~al.(2020)Dosovitskiy, Beyer, Kolesnikov, Weissenborn, Zhai, Unterthiner, Dehghani, Minderer, Heigold, Gelly, et~al.]{dosovitskiy2020image}
Alexey Dosovitskiy, Lucas Beyer, Alexander Kolesnikov, Dirk Weissenborn, Xiaohua Zhai, Thomas Unterthiner, Mostafa Dehghani, Matthias Minderer, Georg Heigold, Sylvain Gelly, et~al.
\newblock An image is worth 16x16 words: {T}ransformers for image recognition at scale.
\newblock \emph{arXiv preprint arXiv:2010.11929}, 2020.

\bibitem[Fathony et~al.(2020)Fathony, Sahu, Willmott, and Kolter]{fathony2020multiplicative}
Rizal Fathony, Anit~Kumar Sahu, Devin Willmott, and J~Zico Kolter.
\newblock Multiplicative filter networks.
\newblock In \emph{International Conference on Learning Representations}, 2020.

\bibitem[Franzen()]{Franzen}
Richard~W. Franzen.
\newblock Kodak lossless true color image suite.
\newblock URL \url{https://r0k.us/graphics/kodak/}.

\bibitem[Hastie et~al.(2009)Hastie, Tibshirani, Friedman, and Friedman]{hastie2009elements}
Trevor Hastie, Robert Tibshirani, Jerome~H Friedman, and Jerome~H Friedman.
\newblock \emph{The elements of statistical learning: data mining, inference, and prediction}, volume~2.
\newblock Springer, 2009.

\bibitem[He et~al.(2015)He, Zhang, Ren, and Sun]{he2015delving}
Kaiming He, Xiangyu Zhang, Shaoqing Ren, and Jian Sun.
\newblock Delving deep into rectifiers: {S}urpassing human-level performance on imagenet classification.
\newblock In \emph{Proceedings of the IEEE international conference on computer vision}, pp.\  1026--1034, 2015.

\bibitem[He et~al.(2016)He, Zhang, Ren, and Sun]{he2016deep}
Kaiming He, Xiangyu Zhang, Shaoqing Ren, and Jian Sun.
\newblock Deep residual learning for image recognition.
\newblock In \emph{Proceedings of the IEEE conference on computer vision and pattern recognition}, pp.\  770--778, 2016.

\bibitem[Hochreiter \& Schmidhuber(1997)Hochreiter and Schmidhuber]{hochreiter1997long}
Sepp Hochreiter and J{\"u}rgen Schmidhuber.
\newblock Long short-term memory.
\newblock \emph{Neural computation}, 9\penalty0 (8):\penalty0 1735--1780, 1997.

\bibitem[Hornik et~al.(1989)Hornik, Stinchcombe, and White]{hornik1989multilayer}
Kurt Hornik, Maxwell Stinchcombe, and Halbert White.
\newblock Multilayer feedforward networks are universal approximators.
\newblock \emph{Neural networks}, 2\penalty0 (5):\penalty0 359--366, 1989.

\bibitem[Kingma \& Ba(2015)Kingma and Ba]{DBLP:journals/corr/KingmaB14}
Diederik~P. Kingma and Jimmy Ba.
\newblock Adam: {A} method for stochastic optimization.
\newblock In Yoshua Bengio and Yann LeCun (eds.), \emph{3rd International Conference on Learning Representations, {ICLR} 2015, San Diego, CA, USA, May 7-9, 2015, Conference Track Proceedings}, 2015.
\newblock URL \url{http://arxiv.org/abs/1412.6980}.

\bibitem[Krishnapriyan et~al.(2021)Krishnapriyan, Gholami, Zhe, Kirby, and Mahoney]{krishnapriyan2021characterizing}
Aditi Krishnapriyan, Amir Gholami, Shandian Zhe, Robert Kirby, and Michael~W Mahoney.
\newblock Characterizing possible failure modes in physics-informed neural networks.
\newblock \emph{Advances in neural information processing systems}, 34:\penalty0 26548--26560, 2021.

\bibitem[Lau et~al.(2023)Lau, Hemachandra, Ng, and Low]{lau2023pinnacle}
Gregory Kang~Ruey Lau, Apivich Hemachandra, See-Kiong Ng, and Bryan Kian~Hsiang Low.
\newblock {PINNACLE: P}inn adaptive collocation and experimental points selection.
\newblock In \emph{The Twelfth International Conference on Learning Representations}, 2023.

\bibitem[Maiorov \& Pinkus(1999)Maiorov and Pinkus]{maiorov1999lower}
Vitaly Maiorov and Allan Pinkus.
\newblock Lower bounds for approximation by {MLP} neural networks.
\newblock \emph{Neurocomputing}, 25\penalty0 (1-3):\penalty0 81--91, 1999.

\bibitem[Mescheder et~al.(2019)Mescheder, Oechsle, Niemeyer, Nowozin, and Geiger]{mescheder2019occupancy}
Lars Mescheder, Michael Oechsle, Michael Niemeyer, Sebastian Nowozin, and Andreas Geiger.
\newblock Occupancy networks: {L}earning 3d reconstruction in function space.
\newblock In \emph{Proceedings of the IEEE/CVF conference on computer vision and pattern recognition}, pp.\  4460--4470, 2019.

\bibitem[Mildenhall et~al.(2021)Mildenhall, Srinivasan, Tancik, Barron, Ramamoorthi, and Ng]{mildenhall2021nerf}
Ben Mildenhall, Pratul~P Srinivasan, Matthew Tancik, Jonathan~T Barron, Ravi Ramamoorthi, and Ren Ng.
\newblock Nerf: {R}epresenting scenes as neural radiance fields for view synthesis.
\newblock \emph{Communications of the ACM}, 65\penalty0 (1):\penalty0 99--106, 2021.

\bibitem[Paszke et~al.(2019)Paszke, Gross, Massa, Lerer, Bradbury, Chanan, Killeen, Lin, Gimelshein, Antiga, et~al.]{paszke2019pytorch}
Adam Paszke, Sam Gross, Francisco Massa, Adam Lerer, James Bradbury, Gregory Chanan, Trevor Killeen, Zeming Lin, Natalia Gimelshein, Luca Antiga, et~al.
\newblock Pytorch: {A}n imperative style, high-performance deep learning library.
\newblock \emph{Advances in neural information processing systems}, 32, 2019.

\bibitem[Qiu et~al.(2018)Qiu, Xu, and Cai]{qiu2018frelu}
Suo Qiu, Xiangmin Xu, and Bolun Cai.
\newblock {FReLU}: {F}lexible rectified linear units for improving convolutional neural networks.
\newblock In \emph{2018 24th international conference on pattern recognition (icpr)}, pp.\  1223--1228. IEEE, 2018.

\bibitem[Rahimi \& Recht(2007)Rahimi and Recht]{rahimi2007random}
Ali Rahimi and Benjamin Recht.
\newblock Random features for large-scale kernel machines.
\newblock \emph{Advances in neural information processing systems}, 20, 2007.

\bibitem[Raissi et~al.(2019)Raissi, Perdikaris, and Karniadakis]{raissi2019physics}
Maziar Raissi, Paris Perdikaris, and George~E Karniadakis.
\newblock Physics-informed neural networks: {A} deep learning framework for solving forward and inverse problems involving nonlinear partial differential equations.
\newblock \emph{Journal of Computational physics}, 378:\penalty0 686--707, 2019.

\bibitem[Ramasinghe \& Lucey(2022)Ramasinghe and Lucey]{ramasinghe2022beyond}
Sameera Ramasinghe and Simon Lucey.
\newblock Beyond periodicity: Towards a unifying framework for activations in coordinate-{MLP}s.
\newblock In \emph{European Conference on Computer Vision}, pp.\  142--158. Springer, 2022.

\bibitem[Saragadam et~al.(2023)Saragadam, LeJeune, Tan, Balakrishnan, Veeraraghavan, and Baraniuk]{saragadam2023wire}
Vishwanath Saragadam, Daniel LeJeune, Jasper Tan, Guha Balakrishnan, Ashok Veeraraghavan, and Richard~G Baraniuk.
\newblock Wire: {W}avelet implicit neural representations.
\newblock In \emph{Proceedings of the IEEE/CVF Conference on Computer Vision and Pattern Recognition}, pp.\  18507--18516, 2023.

\bibitem[Shen et~al.(2021)Shen, Yang, and Zhang]{shen2021neural}
Zuowei Shen, Haizhao Yang, and Shijun Zhang.
\newblock Neural network approximation: {T}hree hidden layers are enough.
\newblock \emph{Neural Networks}, 141:\penalty0 160--173, 2021.

\bibitem[Sitzmann et~al.(2020)Sitzmann, Martel, Bergman, Lindell, and Wetzstein]{sitzmann2020implicit}
Vincent Sitzmann, Julien Martel, Alexander Bergman, David Lindell, and Gordon Wetzstein.
\newblock Implicit neural representations with periodic activation functions.
\newblock \emph{Advances in neural information processing systems}, 33:\penalty0 7462--7473, 2020.

\bibitem[Tancik et~al.(2020)Tancik, Srinivasan, Mildenhall, Fridovich-Keil, Raghavan, Singhal, Ramamoorthi, Barron, and Ng]{tancik2020fourier}
Matthew Tancik, Pratul Srinivasan, Ben Mildenhall, Sara Fridovich-Keil, Nithin Raghavan, Utkarsh Singhal, Ravi Ramamoorthi, Jonathan Barron, and Ren Ng.
\newblock Fourier features let networks learn high frequency functions in low dimensional domains.
\newblock \emph{Advances in neural information processing systems}, 33:\penalty0 7537--7547, 2020.

\bibitem[Trottier et~al.(2017)Trottier, Giguere, and Chaib-Draa]{trottier2017parametric}
Ludovic Trottier, Philippe Giguere, and Brahim Chaib-Draa.
\newblock Parametric exponential linear unit for deep convolutional neural networks.
\newblock In \emph{2017 16th IEEE international conference on machine learning and applications (ICMLA)}, pp.\  207--214. IEEE, 2017.

\bibitem[Vaswani et~al.(2017)Vaswani, Shazeer, Parmar, Uszkoreit, Jones, Gomez, Kaiser, and Polosukhin]{vaswani2017attention}
Ashish Vaswani, Noam Shazeer, Niki Parmar, Jakob Uszkoreit, Llion Jones, Aidan~N Gomez, {\L}ukasz Kaiser, and Illia Polosukhin.
\newblock Attention is all you need.
\newblock \emph{Advances in neural information processing systems}, 30, 2017.

\bibitem[Wang et~al.(2004)Wang, Bovik, Sheikh, and Simoncelli]{wang2004image}
Zhou Wang, Alan~C Bovik, Hamid~R Sheikh, and Eero~P Simoncelli.
\newblock Image quality assessment: {F}rom error visibility to structural similarity.
\newblock \emph{IEEE transactions on image processing}, 13\penalty0 (4):\penalty0 600--612, 2004.

\bibitem[Yarotsky(2021)]{yarotsky2021elementary}
Dmitry Yarotsky.
\newblock Elementary superexpressive activations.
\newblock In \emph{International Conference on Machine Learning}, pp.\  11932--11940. PMLR, 2021.

\bibitem[Zhang et~al.(2022)Zhang, Shen, and Yang]{zhang2022neural}
Shijun Zhang, Zuowei Shen, and Haizhao Yang.
\newblock Neural network architecture beyond width and depth.
\newblock \emph{Advances in Neural Information Processing Systems}, 35:\penalty0 5669--5681, 2022.

\end{thebibliography}
\bibliographystyle{iclr2025_conference}

\appendix
\section{Detailed description on experimental setup and additional results}\label{app:results}
\subsection{Signal Representation}\label{app:img_rep}
As a canonical test for INRs, we evaluate the model's expressivity on signal representation tasks: Image representation and occupancy volume representation. 
\subsubsection{Image Representation}
The first signal type is 2-dimensional image, where a test image is chosen from the Kodak dataset~\citep{Franzen}. Figure~\ref{tab:img_rep_results} shows the ground-truth image and reconstructed images by querying trained INRs on the same original mesh grid. For all methods, we train INRs for 2000 epochs with the initial learning rate 0.005. We set $s_0=30.0$ for Gaussian, $\omega_0=30$ for SIREN, and $s_0=30$, $\omega_0=20$ for WIRE. 

\begin{table}[h]
  \centering
  \setlength{\tabcolsep}{1pt}
  \begin{tabular}{cccc}
    Ground Truth &  NestNet (\textbf{32.9091}, \textbf{0.95}) & WIRE (31.2952, 0.92) & SIREN (27.0472, 0.88) \\
    \includegraphics[width=0.24\textwidth]{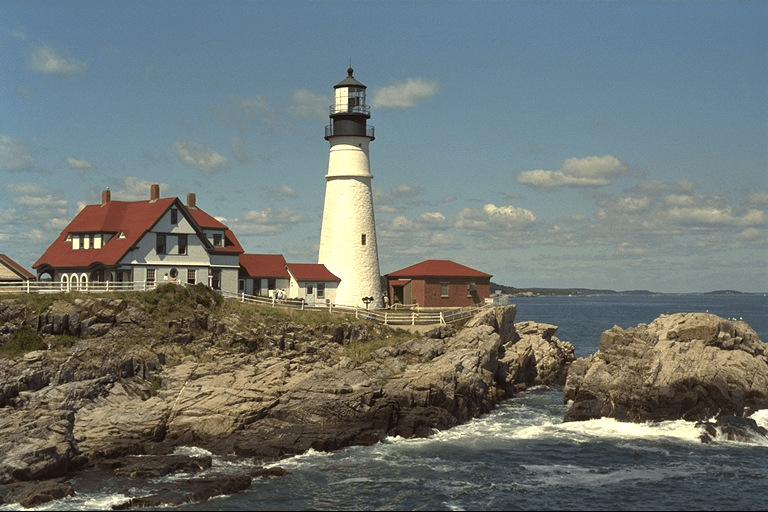} & 
    \includegraphics[width=0.24\textwidth]{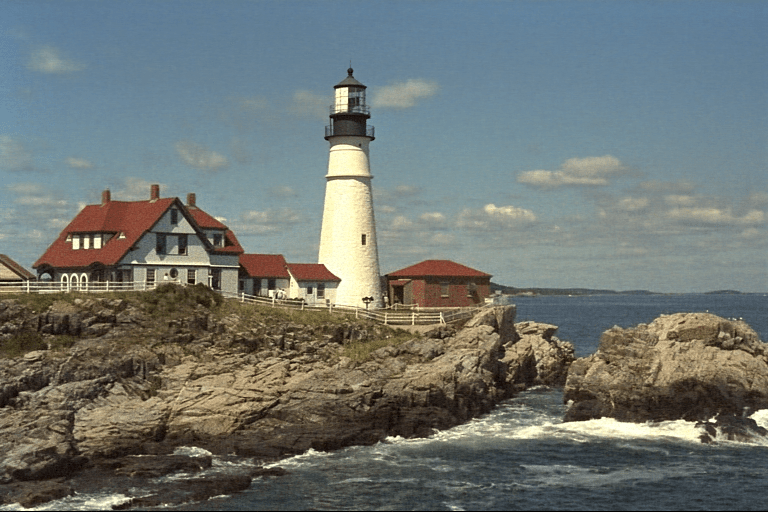} & \includegraphics[width=0.24\textwidth]{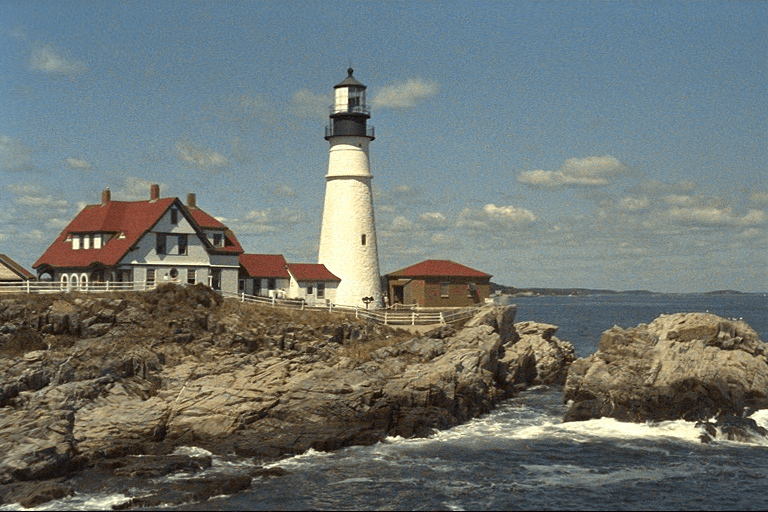} & \includegraphics[width=0.24\textwidth]{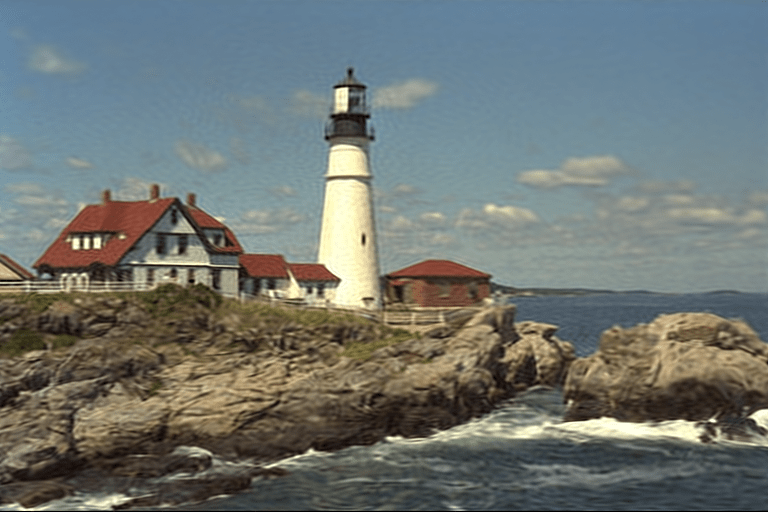}\\
    \includegraphics[width=0.12\textwidth]{figs/img_rep_GT_part_one-min.png}
    \includegraphics[width=0.12\textwidth]{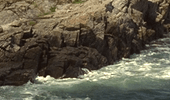} & 
    \includegraphics[width=0.12\textwidth]{figs/img_rep_nestmlp_4_part_one-min.png}
    \includegraphics[width=0.12\textwidth]{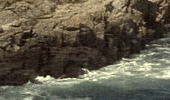} &
    \includegraphics[width=0.12\textwidth]{figs/img_rep_wire_3_part_one-min.png}
    \includegraphics[width=0.12\textwidth]{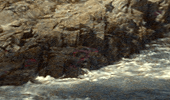} &
    \includegraphics[width=0.12\textwidth]{figs/img_rep_siren_0_part_one-min.png}
    \includegraphics[width=0.12\textwidth]{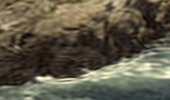} \\

    Gaussian (28.8719, 0.88) &  (MFN 28.9065, 0.92) & FFN (25.9388, 0.84)\\
    \includegraphics[width=0.24\textwidth]{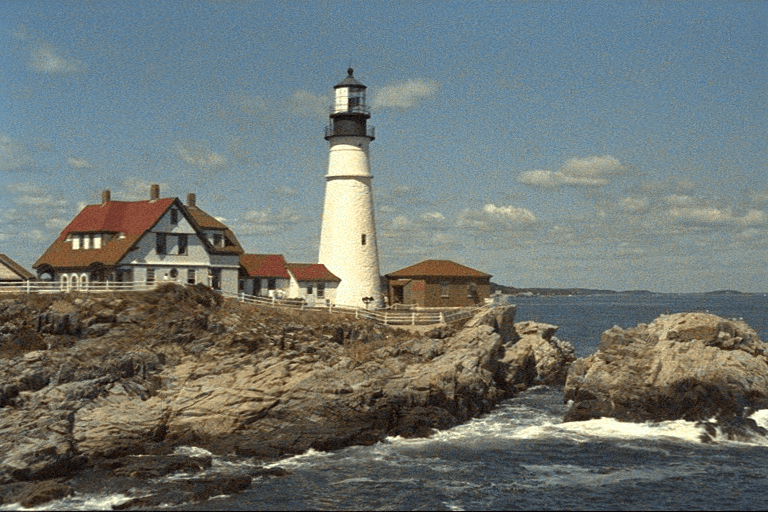} & 
    \includegraphics[width=0.24\textwidth]{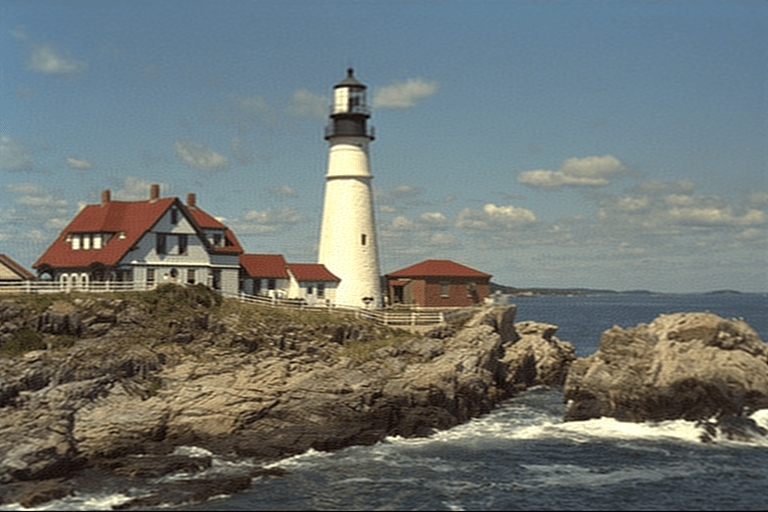} & 
    \includegraphics[width=0.24\textwidth]{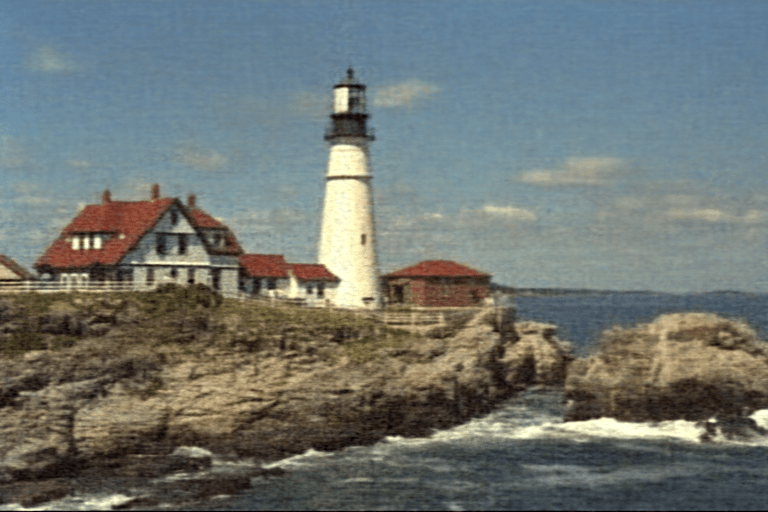}\\
    \includegraphics[width=0.12\textwidth]{figs/img_rep_gauss_3_part_one-min.png}
    \includegraphics[width=0.12\textwidth]{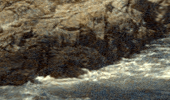} & 
    \includegraphics[width=0.12\textwidth]{figs/img_rep_mfn_2_part_one-min.png}
    \includegraphics[width=0.12\textwidth]{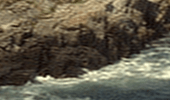} & 
    \includegraphics[width=0.12\textwidth]{figs/img_rep_posenc_3_part_one-min.png}
    \includegraphics[width=0.12\textwidth]{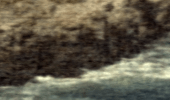}\\

  \end{tabular}
  \vspace{1ex}
  \captionof{figure}{[Image representation] The numbers above the figures report PSNR and SSIM (in the parenthesis).}
  \label{tab:img_rep_results}
\end{table}

\paragraph{Performance comparisons.} NestNet demonstrates the improved performance than other baselines both in PSNR and SSIM. Compared to the second best method (i.e., WIRE), the PSNR and SSIM are improved by +1.614 and +0.03, respectively. 

Differences in reconstruction qualities are more pronounced in zoomed-in plots. The green parts of the images (i.e., the colors of bushes and grasses) are accurately depicted in the reconstruction of NestNet whereas the baselines (e.g., WIRE and Gaussian) result in images of bushes and grasses colored more in brown. Also, in some baselines (again in WIRE and Gaussian), there are some red artifacts in rocks due to inaccurate reconstructions in baselines, while NestNet does not produce such artifact. Other baselines (SIREN and MFN) do not produce such artifact, but produce blurry representation of an image. 

We also report the size of the NestNet and WIRE that are used in producing the reported numbers: NestNet has 153,633 trainable parameters and the WIRE has 66,973 trainable parameters. The training times for NestNet and WIRE are 17 minutes and 43 seconds and 10 minutes and 22 seconds on NVIDIA RTX 3090 machine. Increasing the number of trainable parameters for WIRE, however, does not lead to improved performance; increasing the number of parameters by having more depth in WIRE (depth from 2 to 3 and 4, which results in models with 99,915 and 131,587 trainable parameters) indeed produces INRs with lower PSNR/SSIM with increased training time (15 minutes 37 seconds and  19 minutes 50 seconds, respectively).

\begin{figure}[h]
    \centering
    \vspace{-1mm}
    \includegraphics[scale=.7]{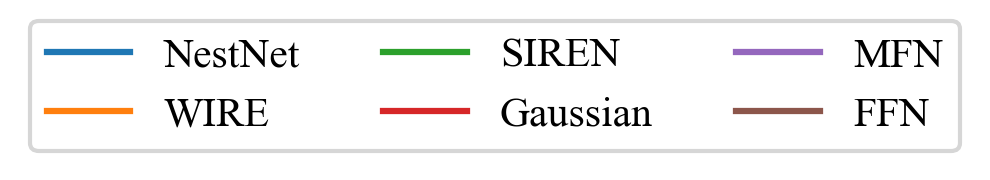}\vspace{-2.mm}\\ 
    \begin{subfigure}{0.39\columnwidth}
    \includegraphics[width=1\columnwidth]{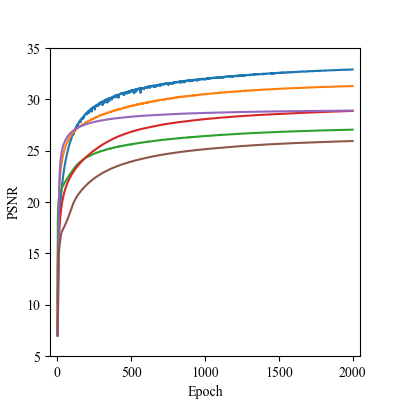}
    \caption{PSNR over epoch.}
    \label{fig:training_loss_psnr}
    \end{subfigure}
    \begin{subfigure}{0.39\columnwidth}
    \includegraphics[width=1\columnwidth]{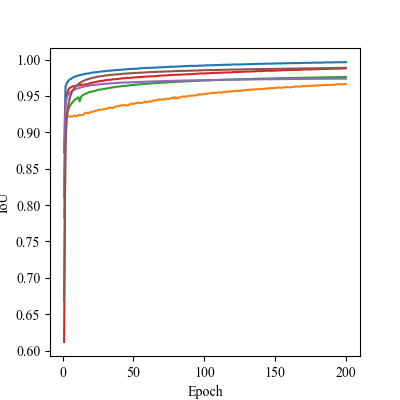}
    \caption{IoU over epoch.}
    \label{fig:training_loss_iou}
    \end{subfigure}
    \caption{[Signal representation] Image representation accuracy is measured over training epochs in terms of PSNR (left, \ref{fig:training_loss_psnr}) and occupancy volume representation accuracy is measured over training epochs in terms of IoU (right, \ref{fig:training_loss_iou}).}
    \label{fig:training_loss}
    \vspace{-1mm}
\end{figure}

\paragraph{Learning curve.} Figure~\ref{fig:training_loss_psnr} reports the PSNR of INRs (i.e., NestNet and the baselines) at each training epoch. Figure~\ref{fig:training_loss_psnr} shows that PSNR of one of the baselines (WIRE) increases faster at very early epochs (under 50 epochs), but NestNet quickly surpasses the compared baselines and achieves the highest PSNR. 

\begin{figure}[!h]
    \vspace{-2mm}
    \centering
    \includegraphics[width=0.7\columnwidth]{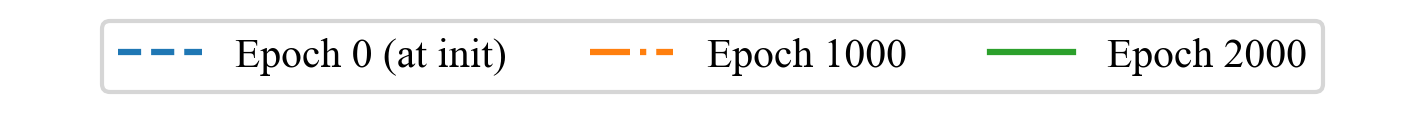}\\
    \begin{subfigure}{.32\columnwidth}
    \includegraphics[width=1.\columnwidth]{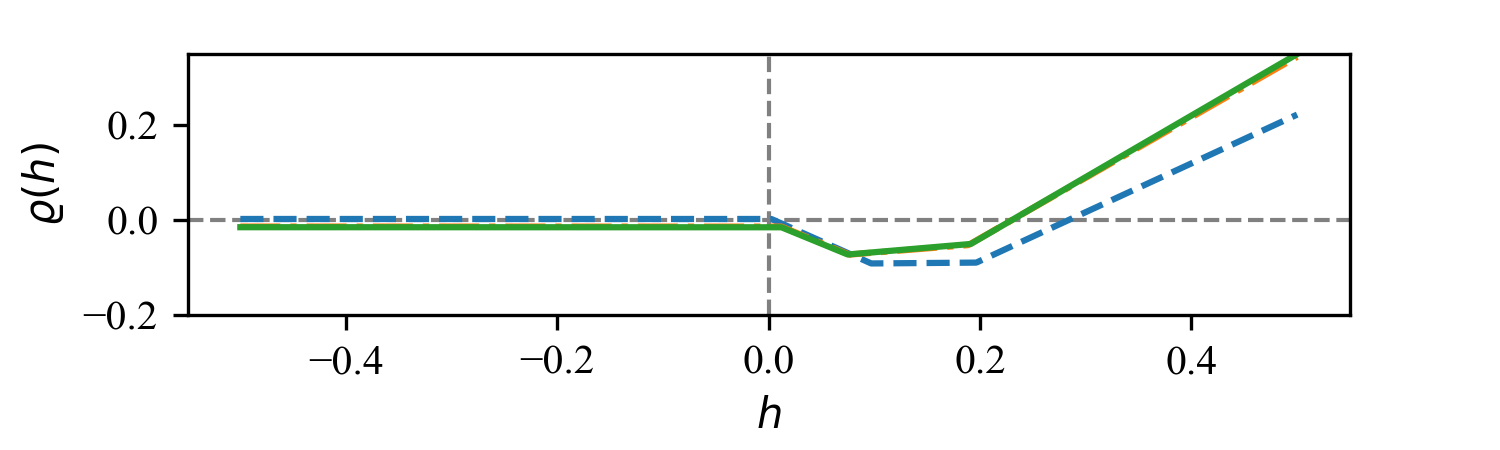}
    \caption{Layer 1}
    \label{fig:nonlin1}
    \end{subfigure}
    \begin{subfigure}{.32\columnwidth}
    \includegraphics[width=1.\columnwidth]{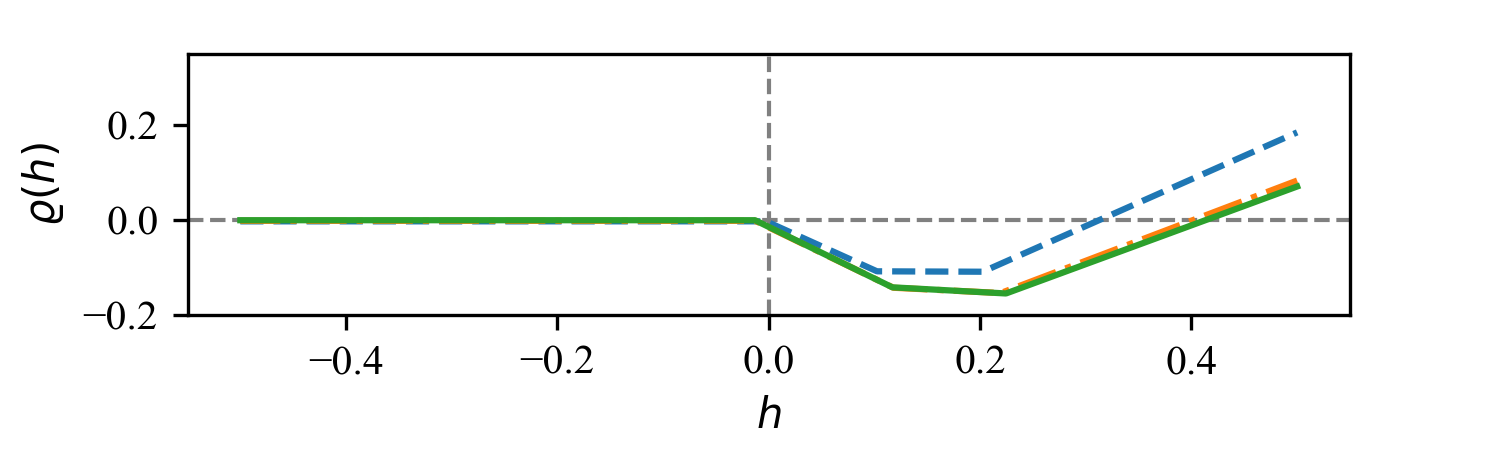}
    \caption{Layer 2}
    \label{fig:nonlin2}
    \end{subfigure}
    \begin{subfigure}{.32\columnwidth}
    \includegraphics[width=1.\columnwidth]{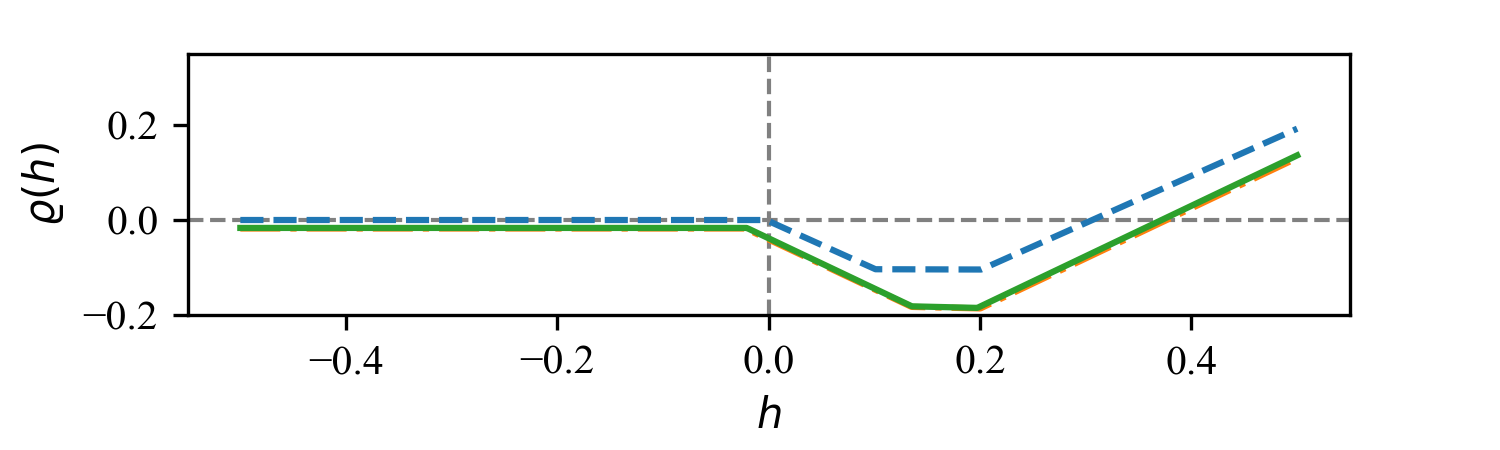}\vspace{-1mm}
    \caption{Layer 3}
    \label{fig:nonlin3}
    \end{subfigure}
    \caption{[Image representation] Nonlinear activation functions $\varrho(h)$ at the first, second, and third layers are instantiated at the initialization (dashed), 1000 epoch (dash-dotted), and 2000 epoch (the final epoch, solid line).}
    \label{fig:nonlin_nestnet}
    \vspace{-2mm}
\end{figure}

\paragraph{Learned nonlinear activation function.} Figure~\ref{fig:nonlin_nestnet} visualizes the learned nonlinear activation functions at each layer (i.e., the first, second, and third layers). Figure depicts the nonlinear activation function $\varrho(h)$ (as defined in Eq.~\eqref{eq:nonlin}) at the initialization (dashed line), at the training epoch 1000 (dash-dotted), and at the training epoch 2000 (the final epoch, solid line). The learned (and the initialized) nonlinear activation functions exhibit function shapes that are evidently different from well-known standard nonlinear activation functions (such as ReLU and its variants). 

Also, as pointed out in the original work~\citet{zhang2022neural}, the shapes of the nonlinear activation functions are nontrivial compared to existing nonlinear activation functions with learnable parameters. Examples of such functions include parametric ReLU \citep{he2015delving}, parametric ELU \citep{trottier2017parametric}, and flexible ReLU~\citep{qiu2018frelu}. All of these trainable nonlinear activation functions, however, can be seen as simple parameteric extensions of the original activation functions, which are still ``non-superexpressive'', and consist of very small learnable components (typically 1 or 2 parameters). On the contrary, subnetworks of NestNets serve as a much more flexible activation functions by randomly distributing the parameters in the affine linear maps and activation functions.

\subsubsection{Occupancy Volumes representation}
We test all considered INRs on representing occupancy volumes~\citep{mescheder2019occupancy}. Again, following the work of \citet{saragadam2023wire}, we consider ``Thai statue'' and sample grid voxels over a $512 \times 512 \times 512$
grid, where each voxel within the volume is  assigned to be 1, and
each voxel outside the volume assigned is assigned to be 0. For all methods, we train INRs for 200 epochs with the initial learning rate 0.005. We set $s_0=40.0$ for Gaussian, $\omega_0=10$ for SIREN, and $s_0=40$, $\omega_0=10$ for WIRE.

Figure~\ref{tab:occupancy_results} demonstrates accurate representations of occupancy volumes achieved by training a NestNet with the IOU value, 0.9964, outperforming the performances of the  baselines. Qualitatively, the reconstruction of NestNet provides finer and more accurate details (e.g., the foot of the statue). 

We further repeat the same experiments executed for the image representation experiment. 
Figure~\ref{fig:training_loss_iou} shows the IOU of INRs (NestNet and the baselines) at each training epoch and shows that the IoU of NestNet at the very beginning is significantly higher than other baselines and the final IOU achieved by NestNet shows a large performance gap with those of the baselines.

\begin{table}[h]
  \centering
  \setlength{\tabcolsep}{1pt}

  \begin{tabular}{ccccccc}
    Ground Truth &  NestNet & WIRE & SIREN & Gaussian &  MFN & FFN \\
    &  \textbf{0.9964}  & 0.9653  & 0.9759  & 0.9877 & 0.9736 & 0.9885  \\
    \includegraphics[width=0.14\textwidth]{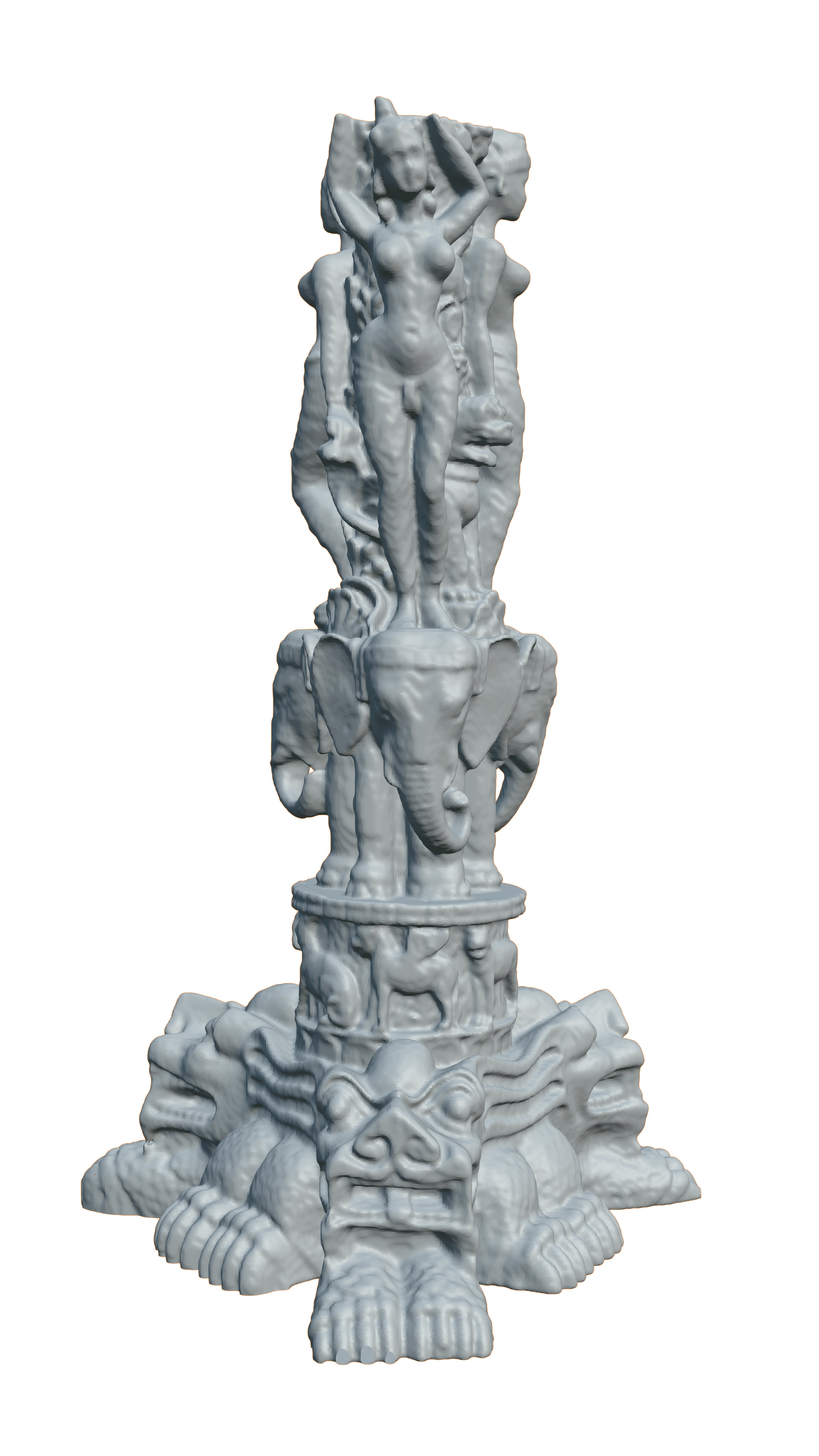} & 
    \includegraphics[width=0.14\textwidth]{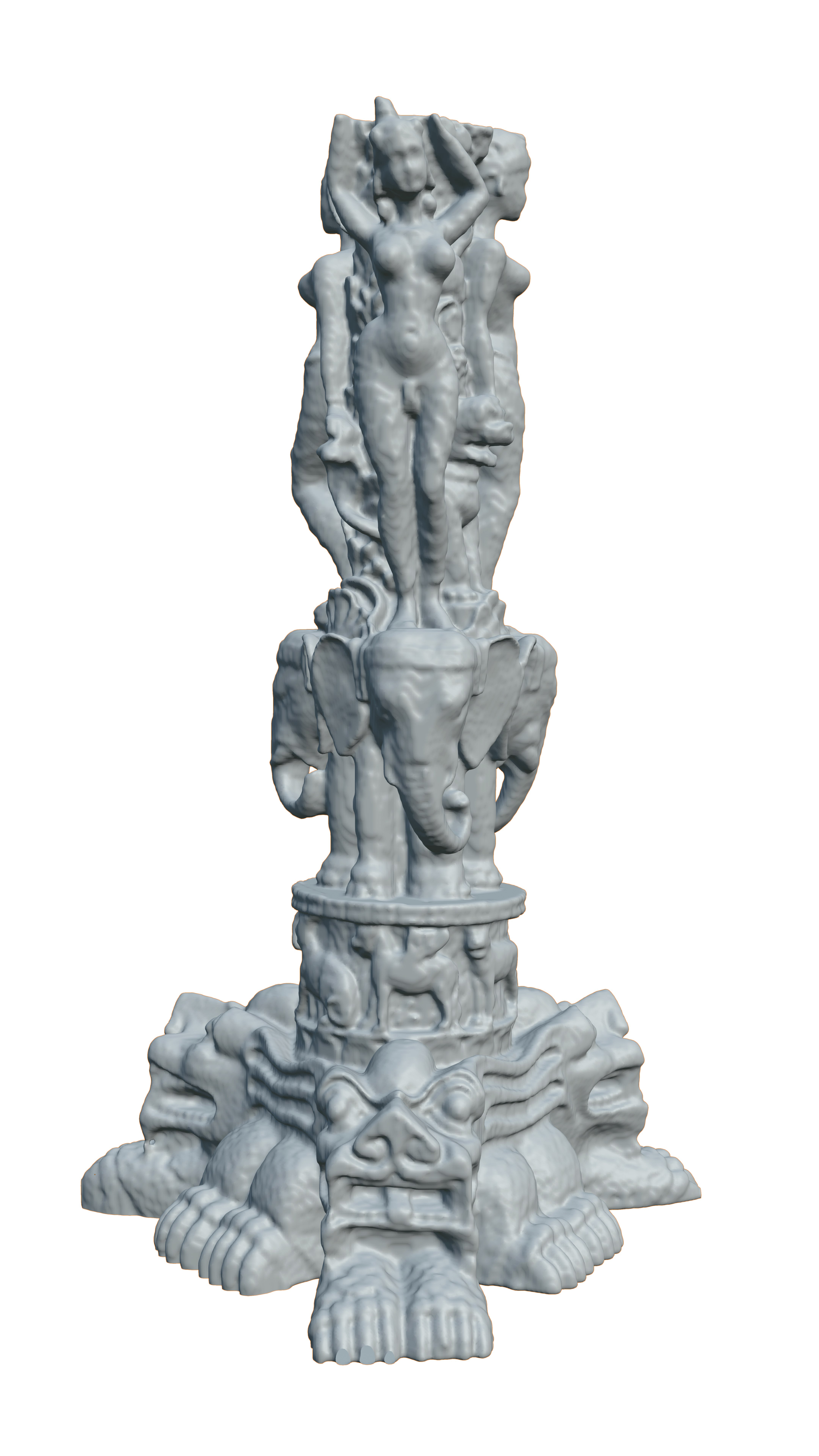} & \includegraphics[width=0.14\textwidth]{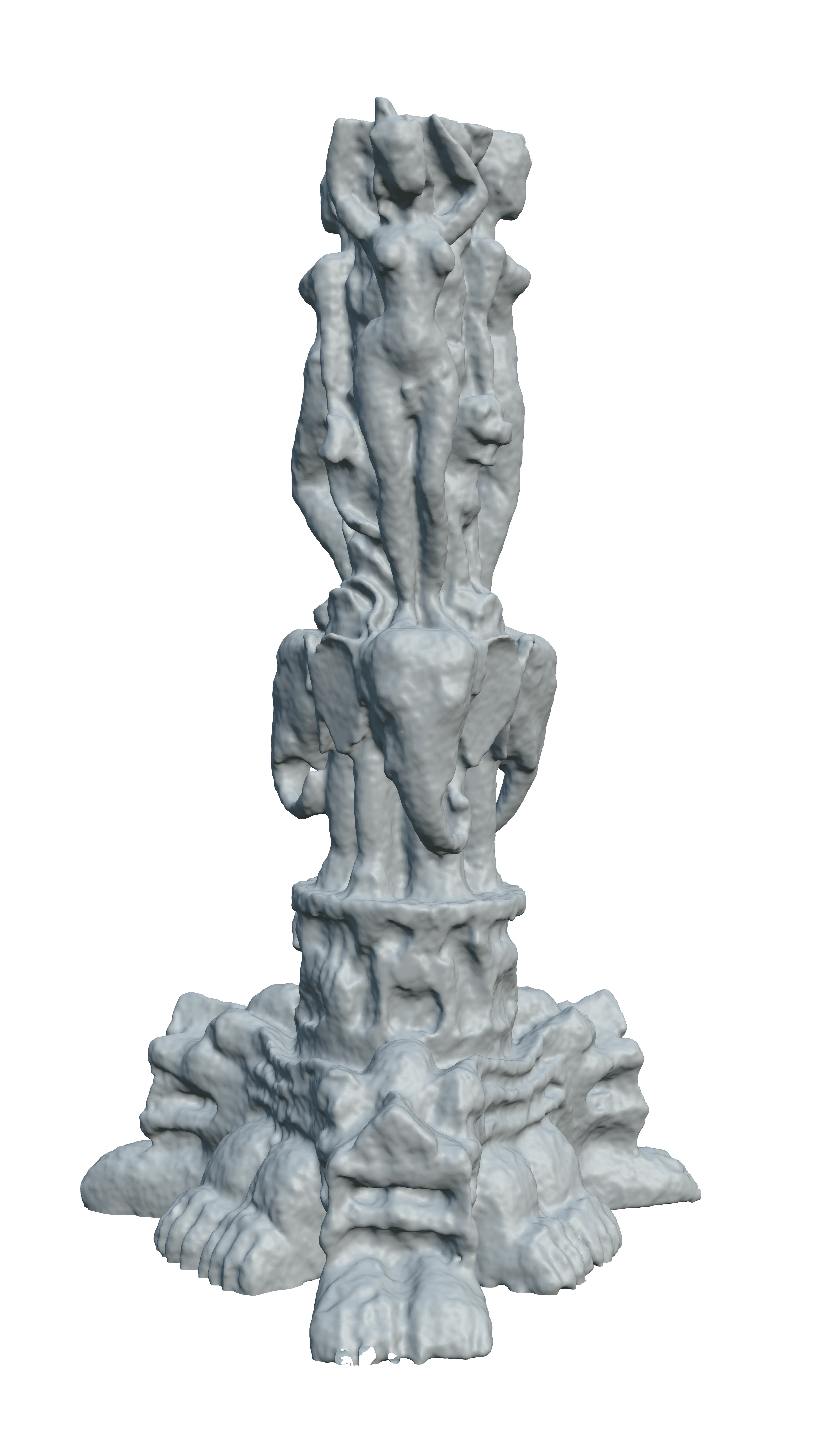} & \includegraphics[width=0.14\textwidth]{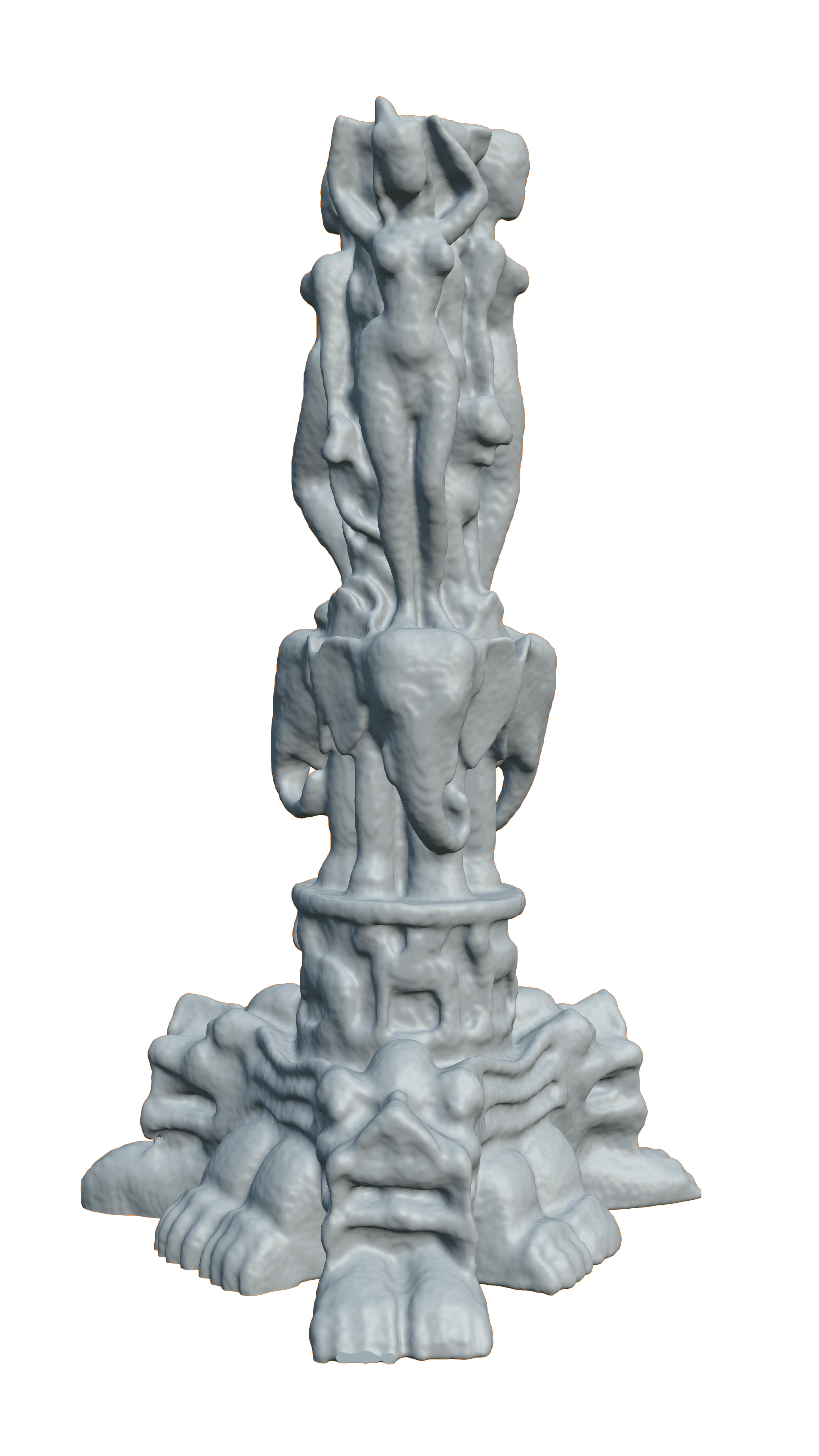} & 
    \includegraphics[width=0.14\textwidth]{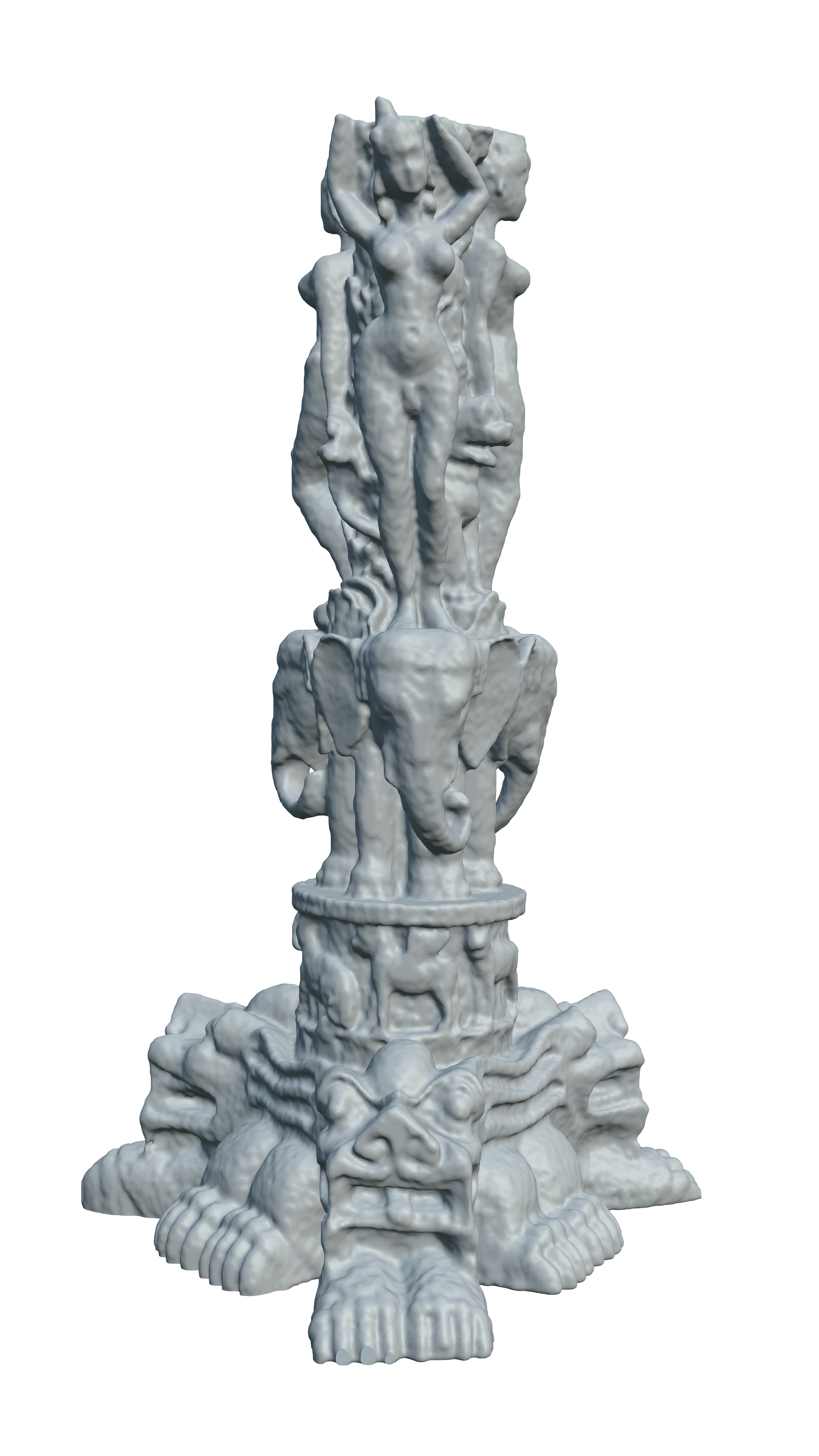} & 
    \includegraphics[width=0.14\textwidth]{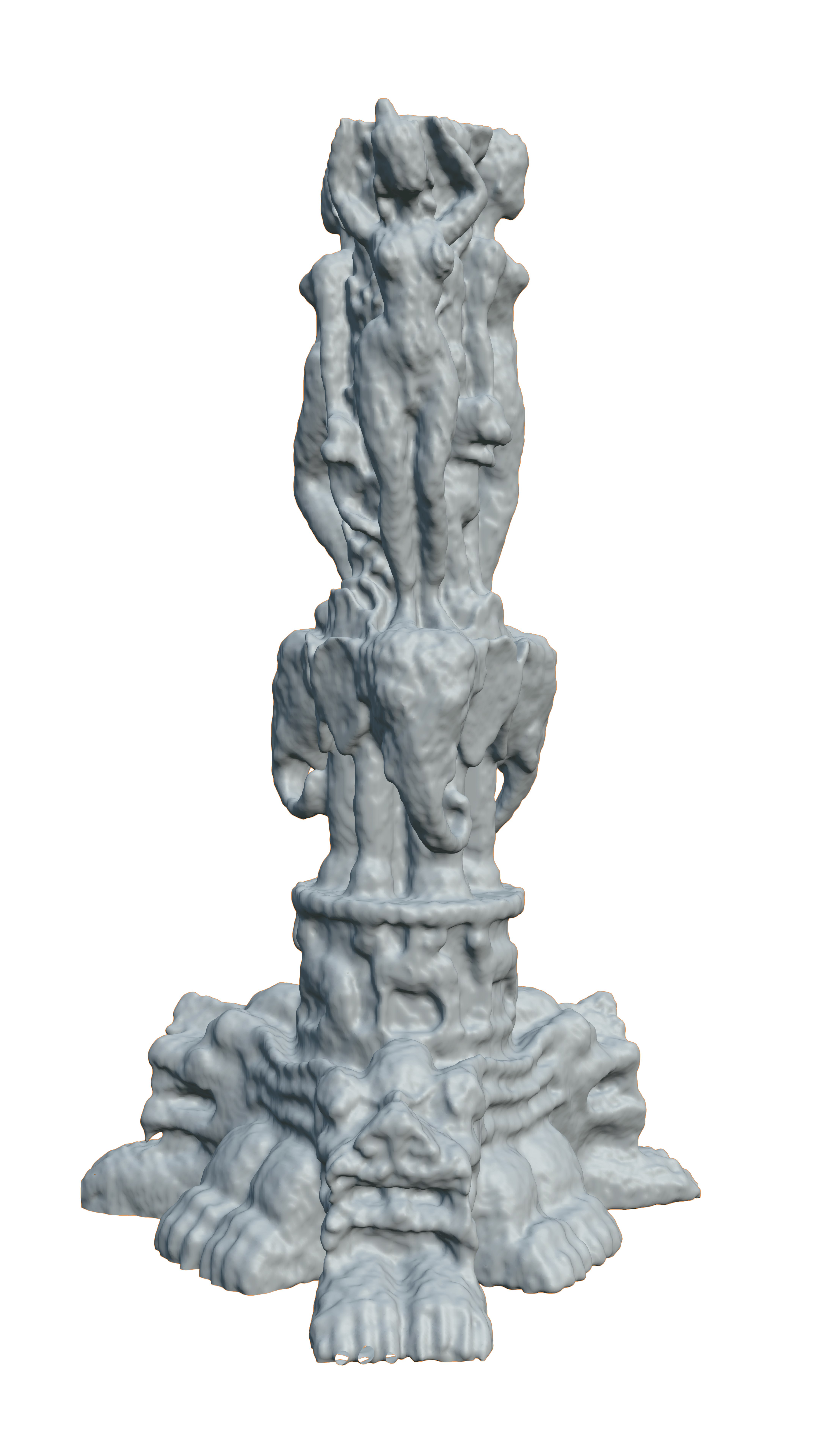} & 
    \includegraphics[width=0.14\textwidth]{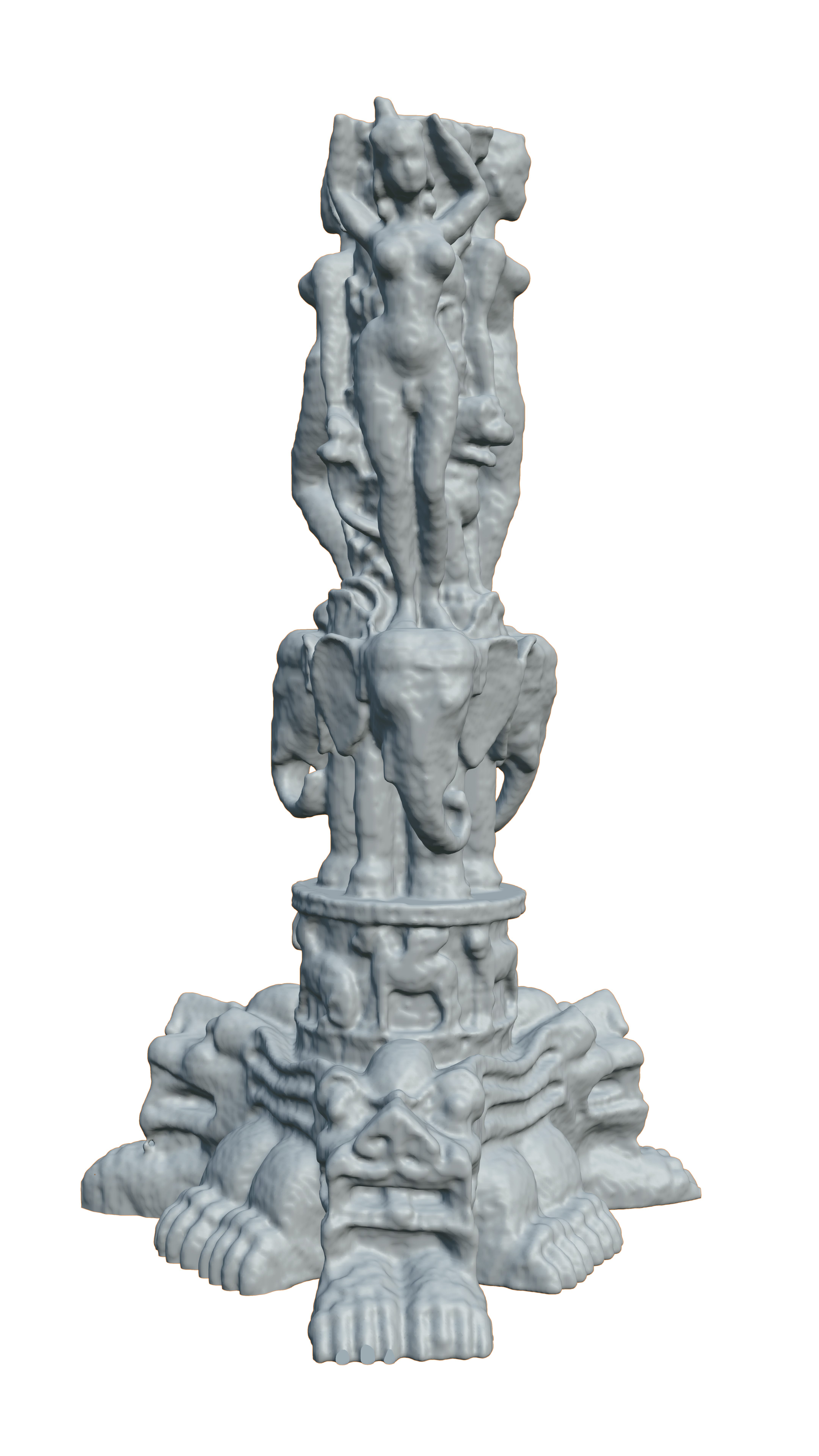}
  \end{tabular}

  \captionof{figure}{[Occupancy Volumes representation] The results above show meshes generated with occupancy volumes with various implicit neural representations. The numbers above the figures report IOU values.}
  \label{tab:occupancy_results}
\end{table}

Figure~\ref{tab:occupancy_volume_rep_zoom} presents zoomed-in plots for the occupancy volume representation, showcasing the better quality of NestNet’s model in detailing the ``Thai statue''. The NestNet model excels in preserving intricate details such as the muscular tonality and textures of the figures. It captures the smooth contours and sharp lines characteristic of the original sculpture, achieving a level of detail closely resembling the Ground Truth.


\begin{table}[!h]
  \centering
  \setlength{\tabcolsep}{2pt} 
  \begin{tabular}{ccccccc}
    Ground Truth & NestNet & WIRE & SIREN  & Gaussian & MFN & FFN \\ 
    \includegraphics[width=0.12\textwidth]{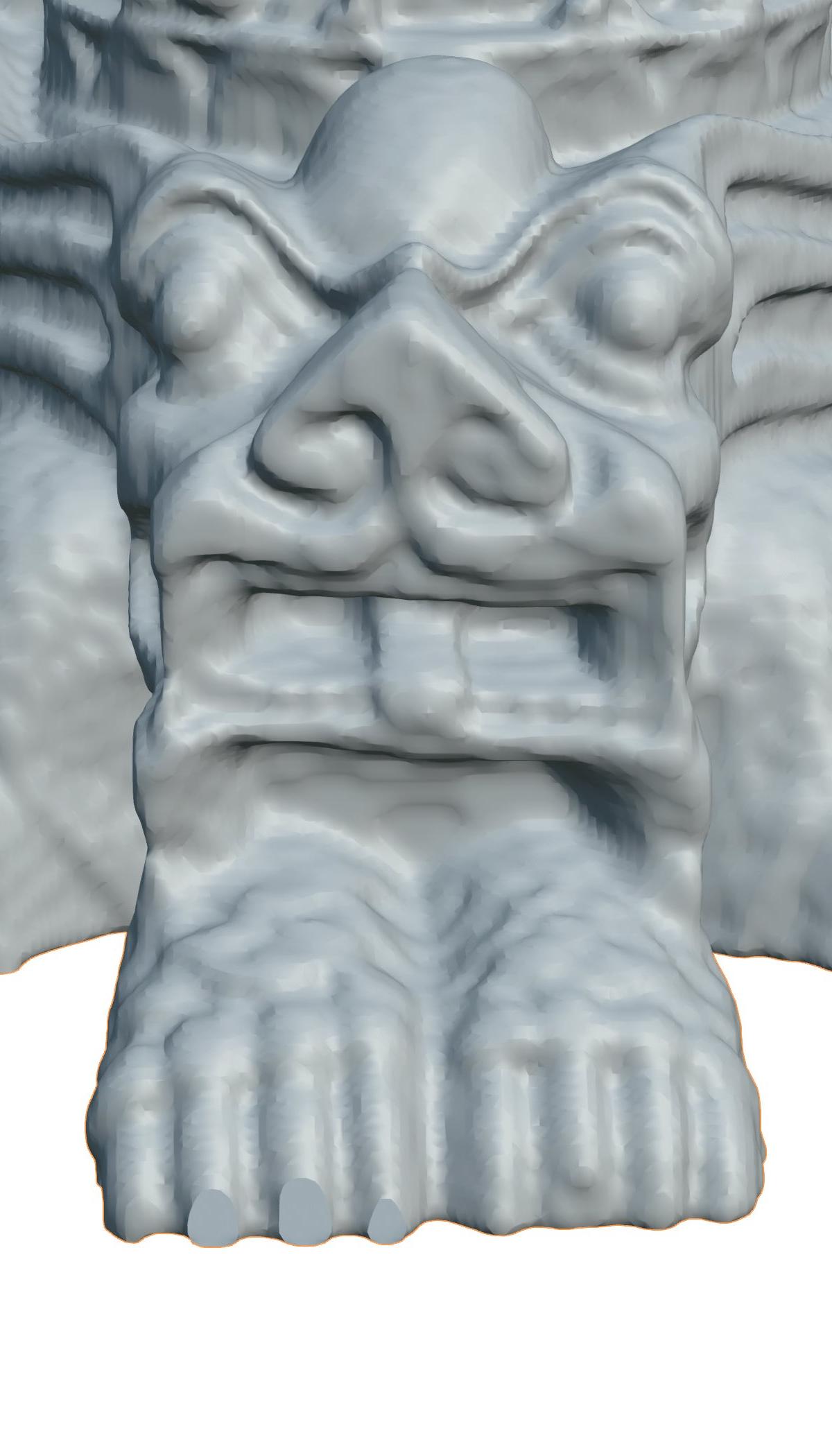} & 
    \includegraphics[width=0.12\textwidth]{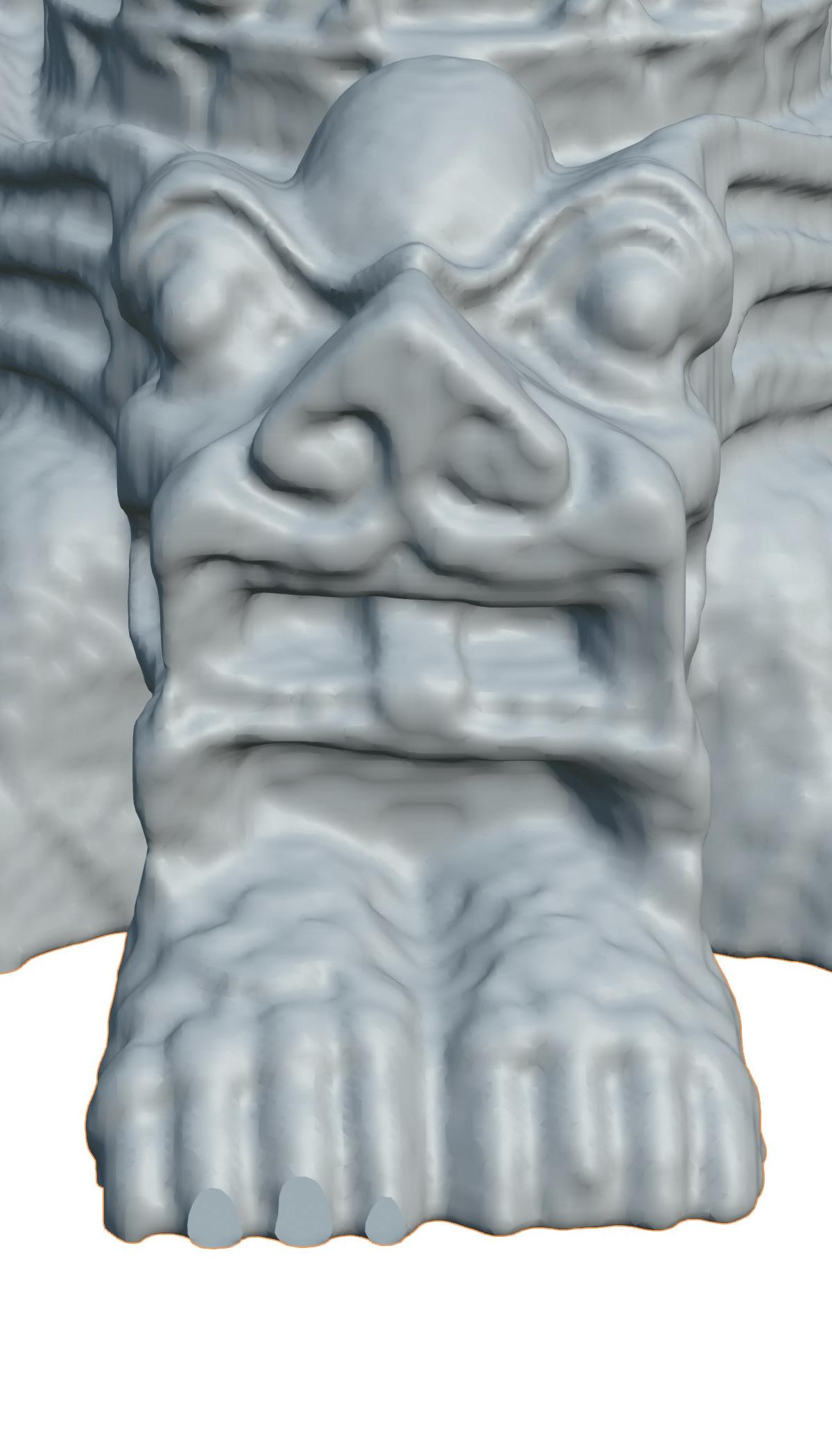} &
    \includegraphics[width=0.12\textwidth]{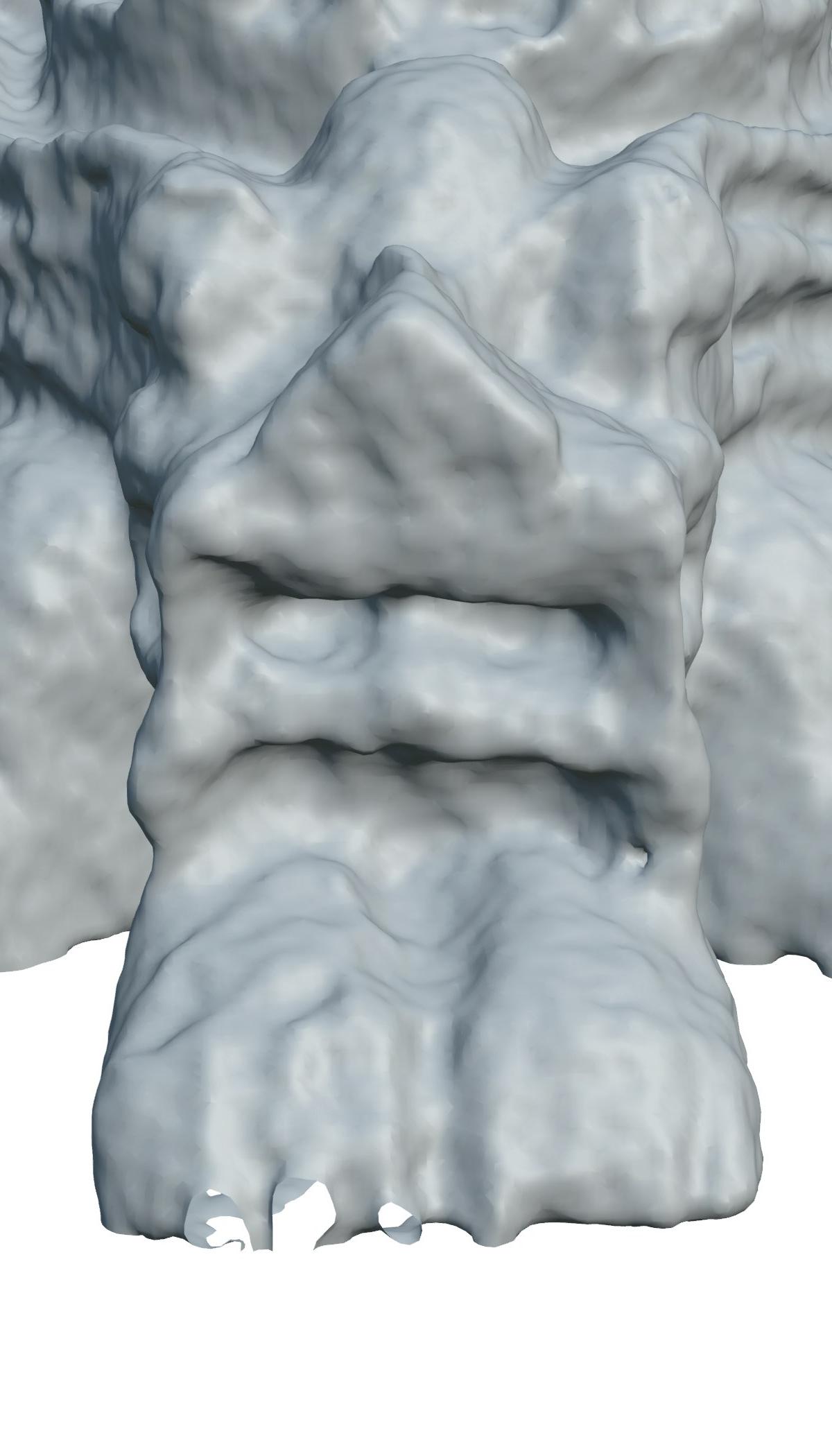} &
    \includegraphics[width=0.12\textwidth]{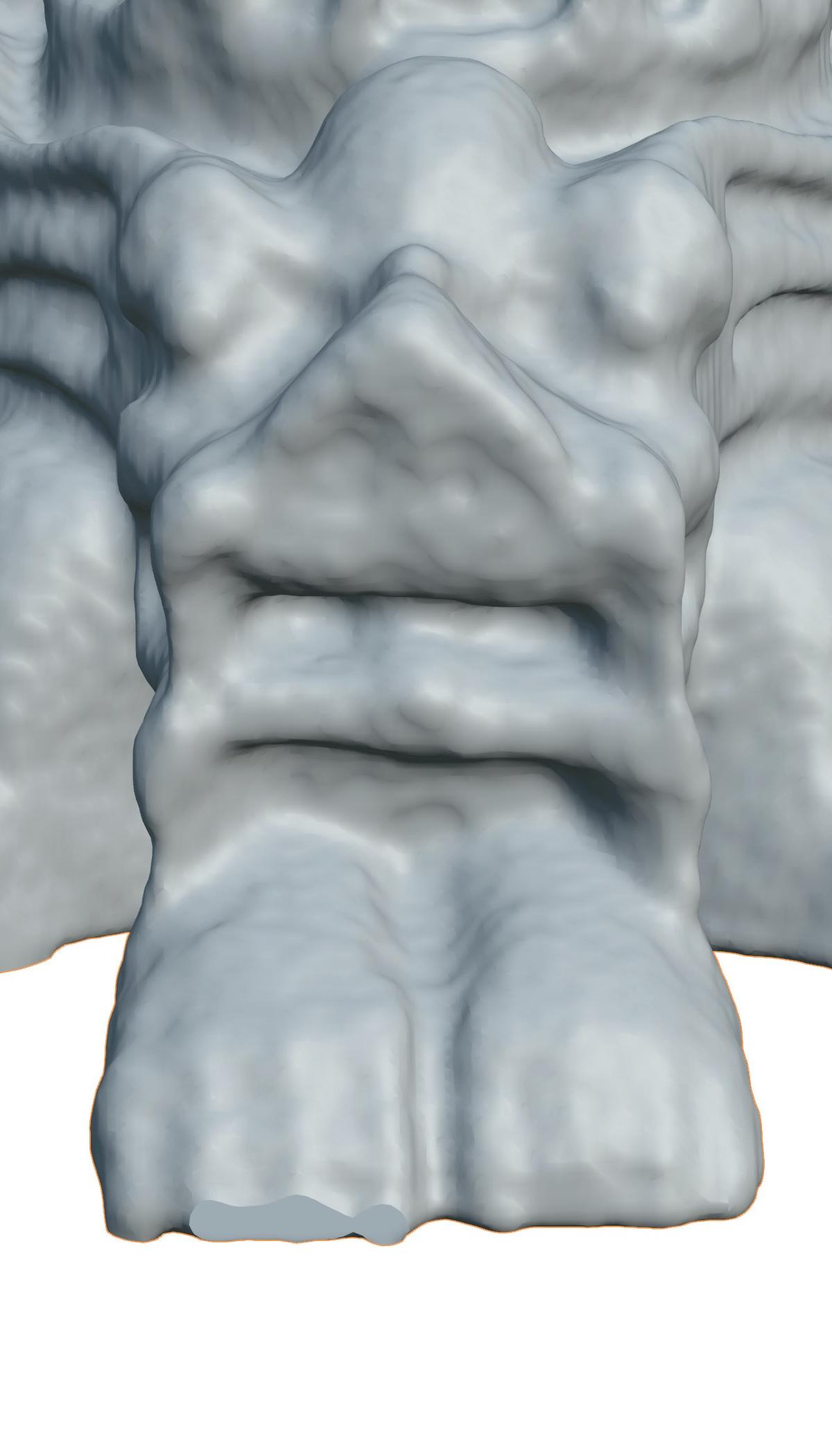} 
    
    & \includegraphics[width=0.12\textwidth]{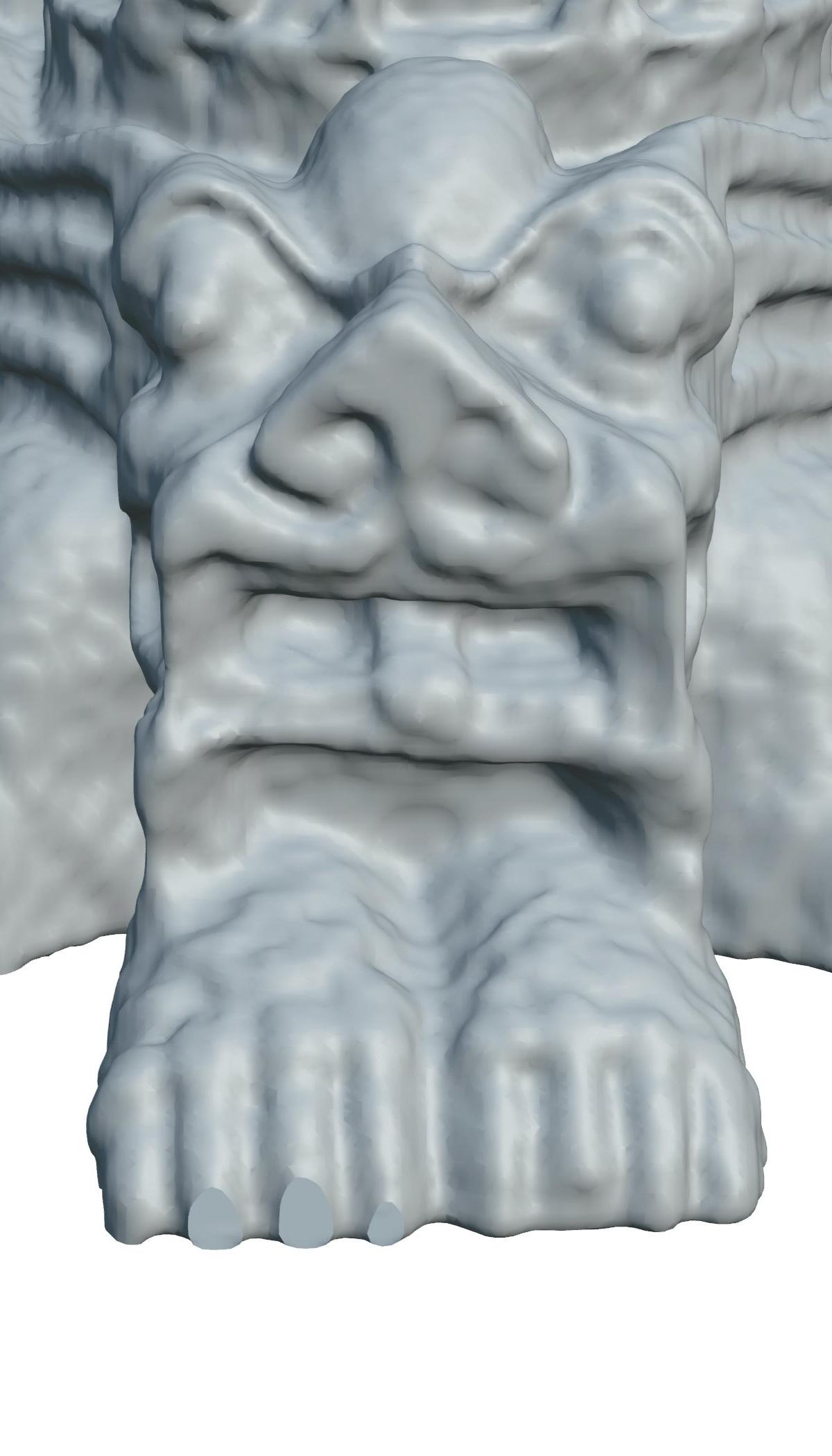} &
    \includegraphics[width=0.12\textwidth]{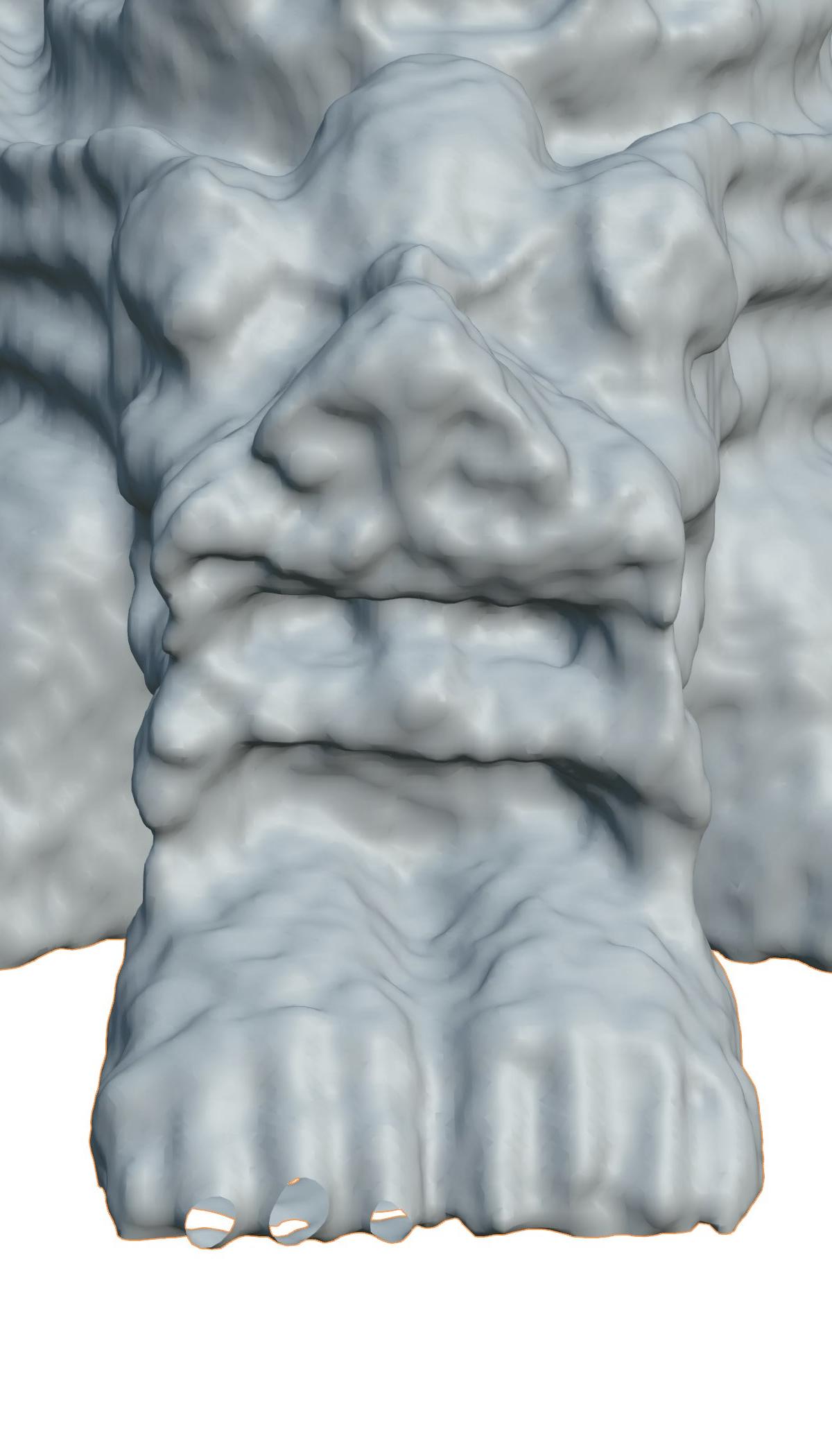} &
    \includegraphics[width=0.12\textwidth]{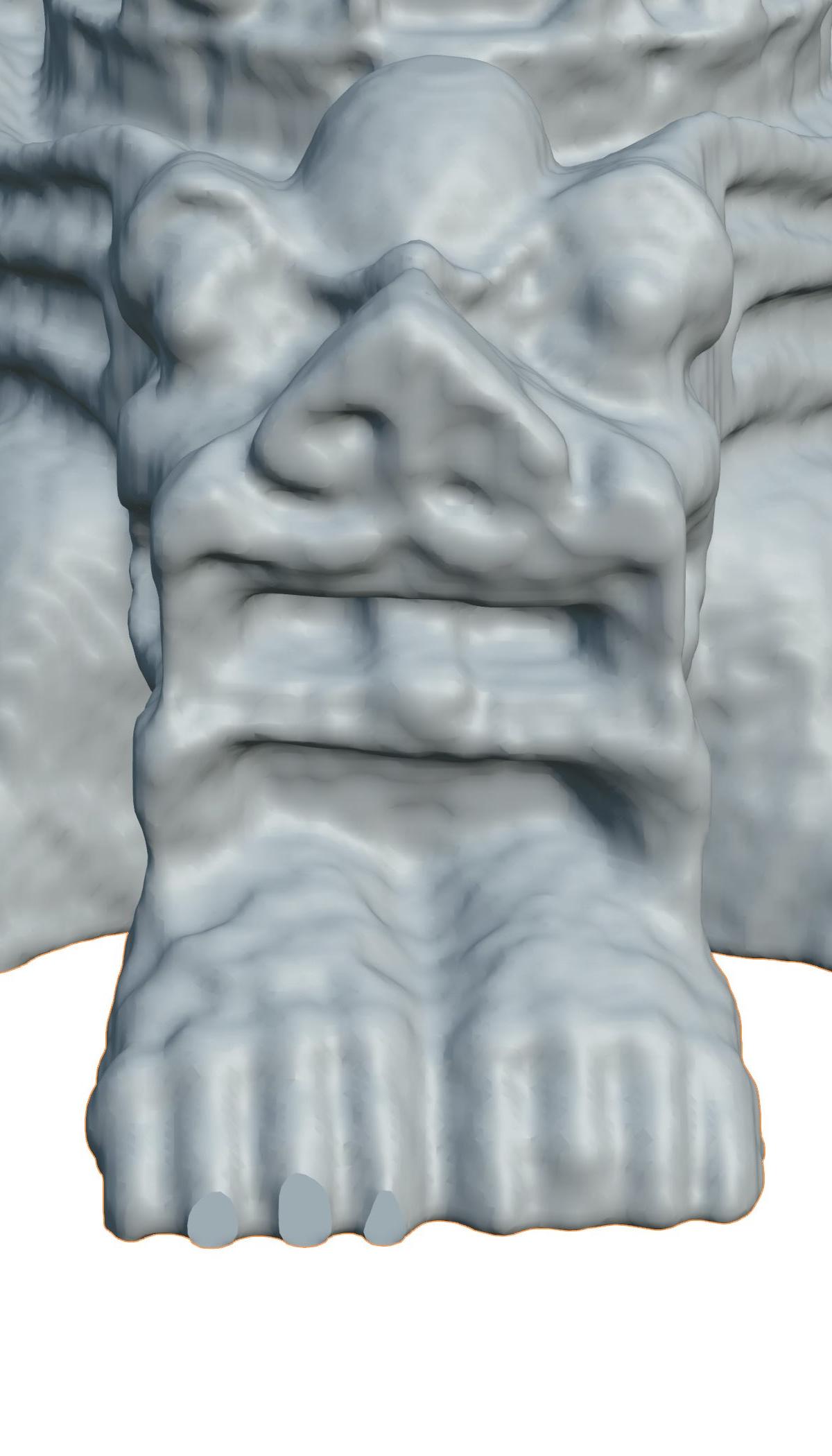}
  \end{tabular}\vspace{-5mm}
  \captionof{figure}{[Occupancy Volume Representation] Figures depict the zoomed-in comparison for all considered methods.}
  \label{tab:occupancy_volume_rep_zoom}
\end{table}

\subsection{Inverse Problems}
A next set of tests includes inverse problems from Computer Vision tasks ranging from super-resolution to image denoising to CT reconstruction. 
\subsubsection{Super-resolution}
\paragraph{Single image super-resolution.}
Following the procedure given in \citet{saragadam2023wire}, we downsample an image by a factor of 1/4. Then we implement 4$\times$ super-resolution by training an INR on the downsampled images and querying pixel densities on 4$\times$ refined mesh. A test image is chosen from the DIV2K dataset~\citep{agustsson2017ntire}, which is depicted in Figure~\ref{tab:SISR_results}.  For all methods, we train INRs for 2000 epochs with the initial learning rate 0.01. We set $s_0=6.0$ for Gaussian, $\omega_0=8$ for SIREN, and $s_0=6$, $\omega_0=8$ for WIRE.

In Figure~\ref{tab:SISR_results}, NestNet is shown to be very performant quantitatively, outperforming the baseline methods both in PSNR and SSIM. Also, qualitatively, the reconstruction of NestNet captures finer details of the high-resolution image (i.e., the clear segmentation object while presenting smooth and continuous representation within each segment). 

\begin{table}[h]
  \centering
  \setlength{\tabcolsep}{1pt}
  \begin{tabular}{cccc}
    Ground Truth & Bilinear Interpolation& NestNet (\textbf{27.5275}, \textbf{0.87}) & WIRE (26.9199, 0.85)\\
    \includegraphics[width=0.24\textwidth]{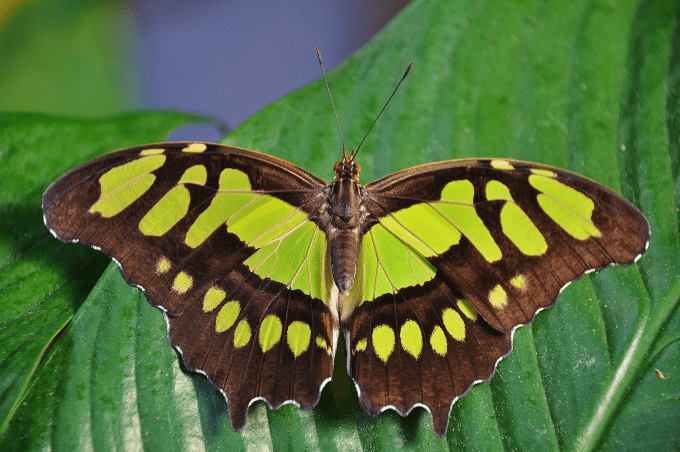} & 
    \includegraphics[width=0.24\textwidth]{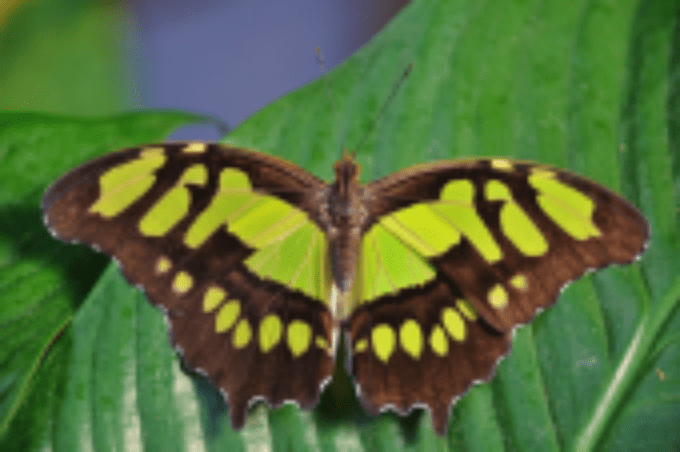} & \includegraphics[width=0.24\textwidth]{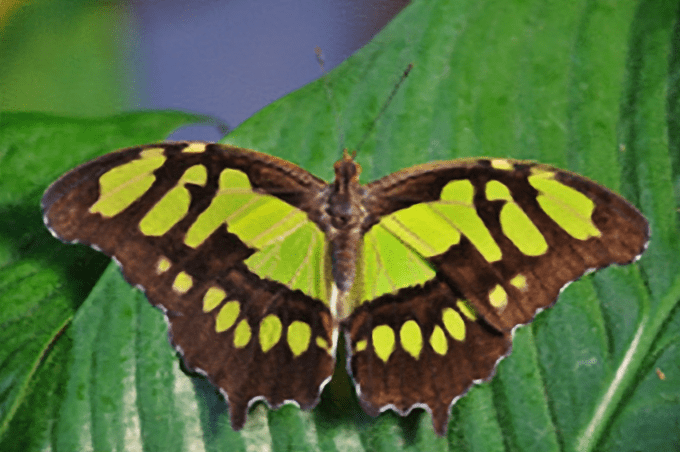} & 
    \includegraphics[width=0.24\textwidth]{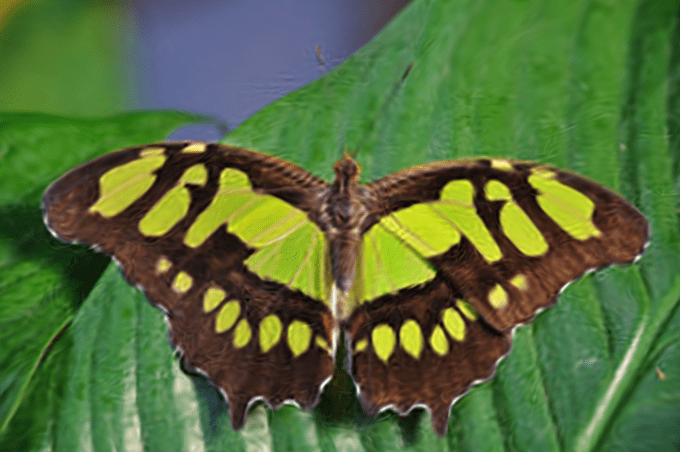}\\
    
    SIREN (24.6793, 0.79) & Gaussian (24.2776, 0.78)& MFN (24.0512, 0.69) & FFN (24.9249, 0.77)\\
    \includegraphics[width=0.24\textwidth]{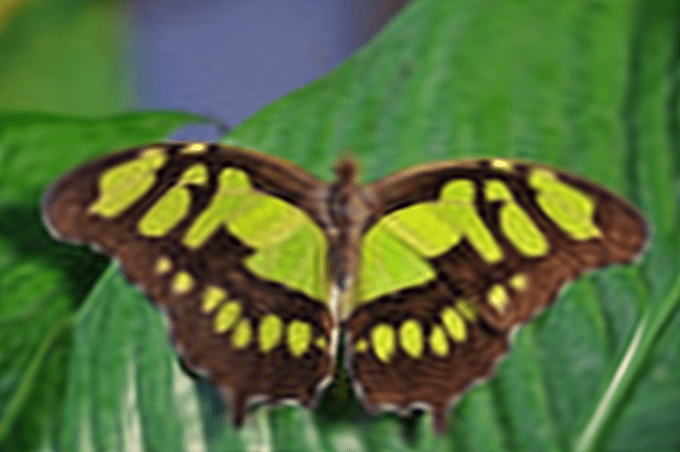} & 
    \includegraphics[width=0.24\textwidth]{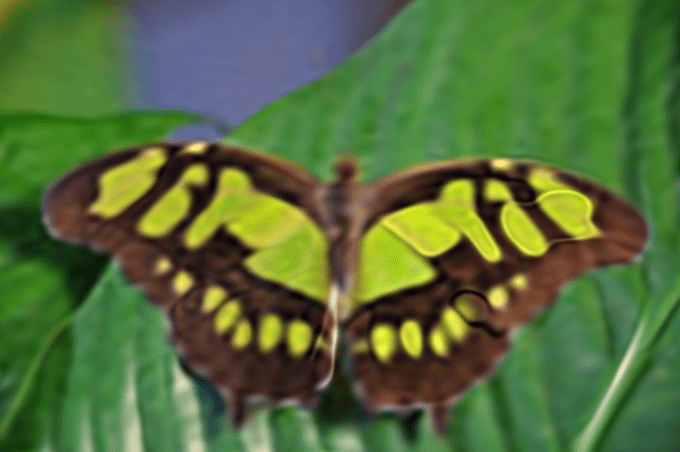} & \includegraphics[width=0.24\textwidth]{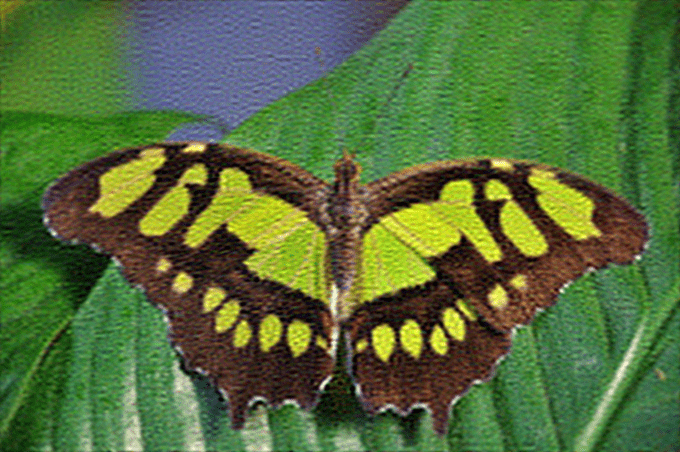} & \includegraphics[width=0.24\textwidth]{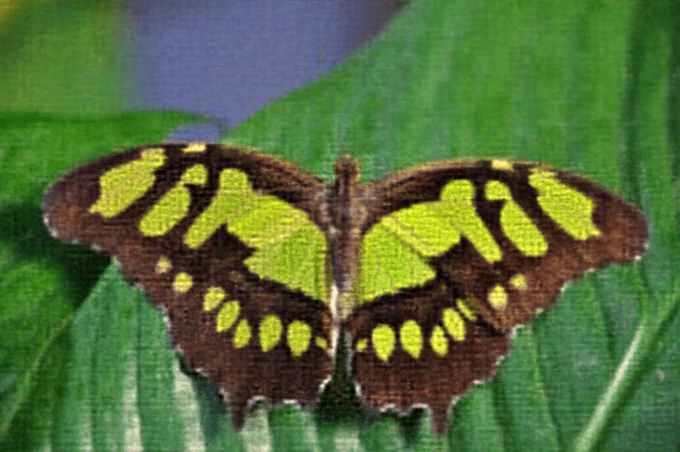}\\
  \end{tabular}
  \captionof{figure}{[Single image super resolution] Figures depict the results of 4$\times$ super resolution for all considered methods. The numbers report PSNR and SSIM (in the parenthesis).}
  \label{tab:SISR_results}
\end{table}

Figure~\ref{tab:SISR_results_app} presents zoomed-in plots for single image super-resolution of a butterfly image from the DIV2K dataset. This analysis highlights NestNet's excellent edge preservation and texture fidelity, ensuring both major and subtle features—such as the delineation of wing spots, lines, and the gradation of colors—are distinctly and naturally restored. While WIRE closely approaches the success of NestNet, it falls short in representing some finer details. For example, the red lines on the butterfly's wings appear blurred in comparison to the sharp, clear edges produced by NestNet. Additionally, the texture at the wing edges has a slightly muddled appearance with WIRE, unlike the sharpness observed with NestNet. WIRE also struggles to accurately capture the color gradient and the antennae details that exist in the ground truth image. Although NestNet has not fully perfected these elements, it remains the most successful among the methods compared.

\begin{table}[!htb]
  \centering
  \setlength{\tabcolsep}{2pt} 
  \begin{tabular}{cccccccc}
    Ground Truth & Bi. Interp & NestNet  & WIRE  &SIREN  & Gaussian & MFN & FFN \\ 
    \includegraphics[width=0.11\textwidth]{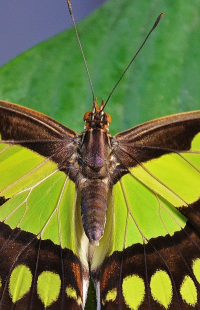} & 
    \includegraphics[width=0.11\textwidth]{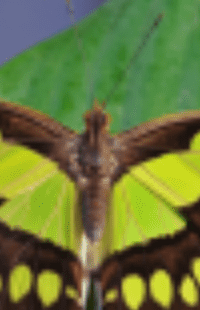} &
    \includegraphics[width=0.11\textwidth]{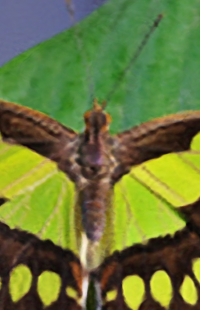} &
    \includegraphics[width=0.11\textwidth]{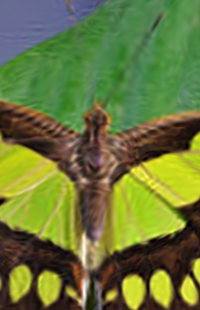} &
    \includegraphics[width=0.11\textwidth]{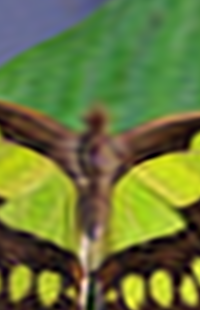} &
    \includegraphics[width=0.11\textwidth]{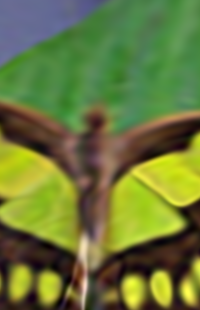} &
    \includegraphics[width=0.11\textwidth]{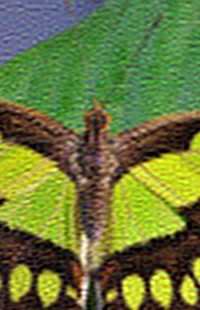} &
    \includegraphics[width=0.11\textwidth]{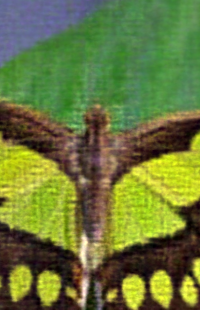} 
  \end{tabular}
  \captionof{figure}{[Single image super resolution] Figures depict the zoomed-in comparison for all considered methods.}
  \label{tab:SISR_results_app}
\end{table}

\paragraph{Multi-image super-resolution.} Next, we assess the performance of NestNets in multi-image super-resolution on the Kodak dataset. The main purpose of this test is to assess the model's capability in interpolating signals measured on an irregular grid and the input source is forged by creating 
multiple images by shifting and rotating with respect to each other from one sample image. This procedure is implemented by following \citet{saragadam2023wire}, which encodes a small sub-pixel motion between four images created by downsampling,  translating and rotating. For all methods, we train INRs for 2000 epochs with the initial learning rate 0.005. We set $s_0=5.0$ for Gaussian, $\omega_0=5$ for SIREN, and $s_0=5$, $\omega_0=5$ for WIRE.

The ground-truth image and results of super-resolution by each method are shown in Figure~\ref{tab:MISR_results} and corresponding PSNR and SSIM are reported. A NestNet results in an image with much improved PSNR (+1 compared to the next best result) and SSIM (+0.04 compared to the next best result). All these values are higher than the best results reported in the previous work \citet{saragadam2023wire} on the same image.

Figure~\ref{tab:MISR_results_app} displays zoomed-in plots for multi-image super-resolution on the Kodak dataset, focusing on an image of a biker with a green helmet. NestNet excels in preserving the textures and colors of the rider's gear, particularly producing a vibrant green color that closely matches the ground truth image. Additionally, NestNet surpasses other models in rendering sharp edges and well-defined backgrounds. Notably, the right-side zoomed-in plot reveals that Wire tends to create a muddled background.

\begin{table}[h]
  \centering
  \setlength{\tabcolsep}{1pt}
  \begin{tabular}{cccc}
    Ground Truth & Bicubic ($\times 4$) & NestNet (\textbf{24.3151}, \textbf{0.82}) & WIRE (23.2108, 0.78)\\
    \includegraphics[width=0.24\textwidth]{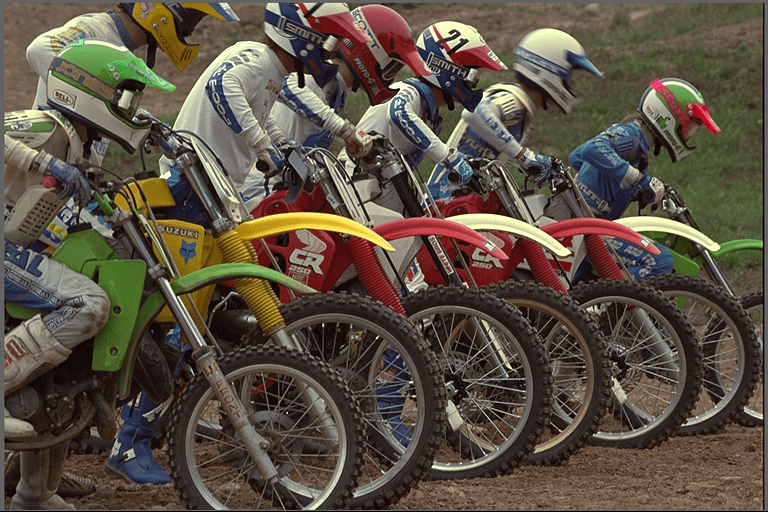} & 
    \includegraphics[width=0.24\textwidth]{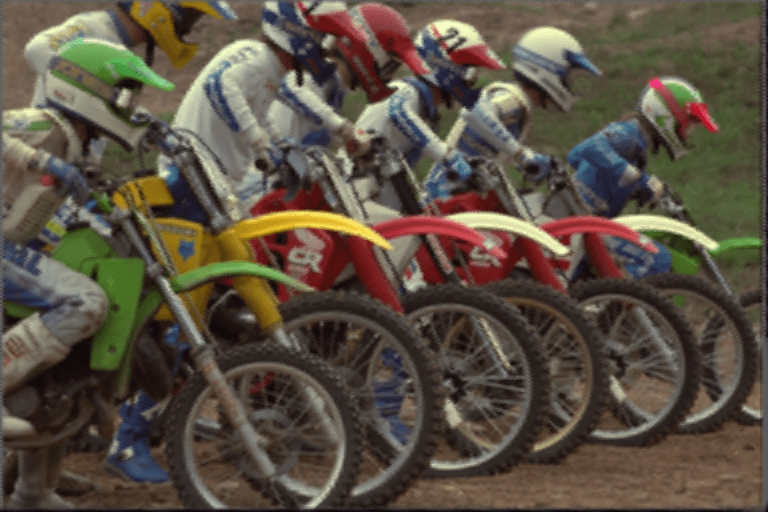} & \includegraphics[width=0.24\textwidth]{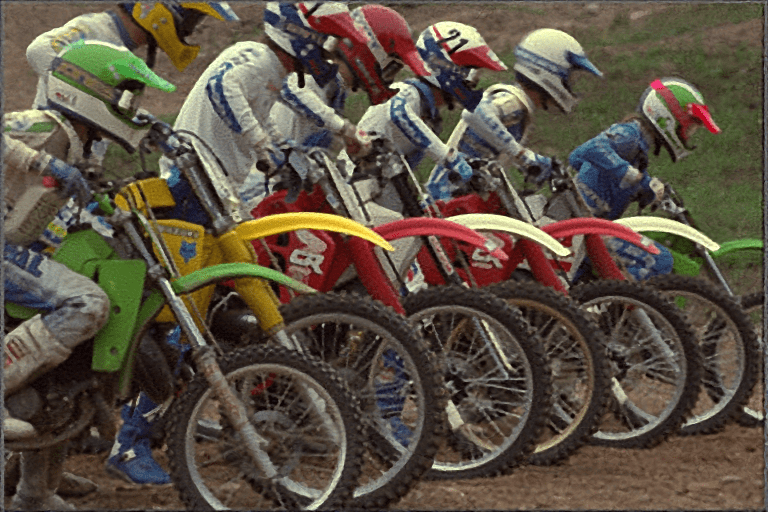} & 
    \includegraphics[width=0.24\textwidth]{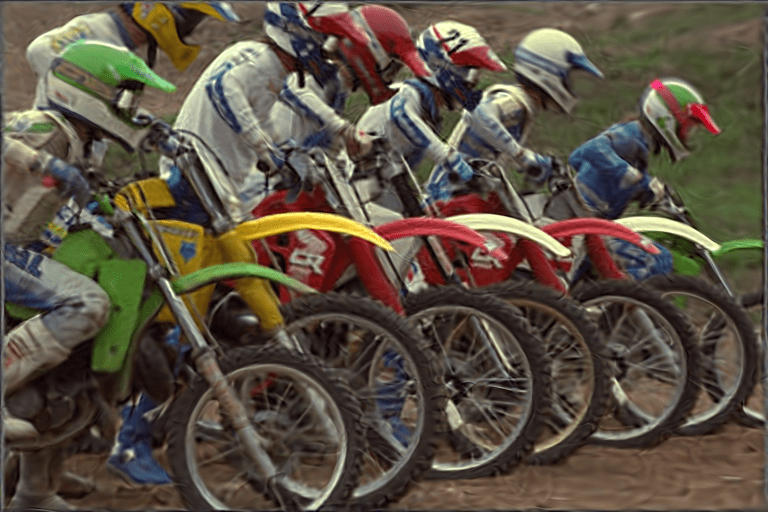}\\

    SIREN (21.7545, 0.69) & Gaussian (21.0305, 0.63)& MFN (19.7773, 0.63) & FFN (21.7824, 0.68)\\
    \includegraphics[width=0.24\textwidth]{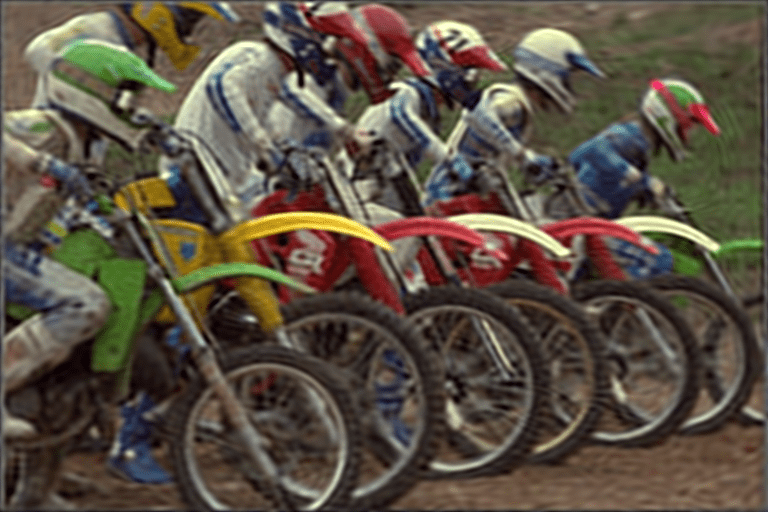} & 
    \includegraphics[width=0.24\textwidth]{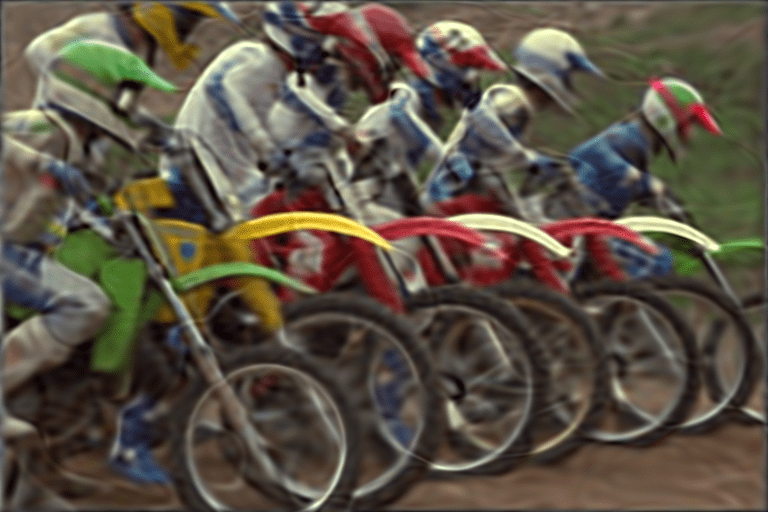} & \includegraphics[width=0.24\textwidth]{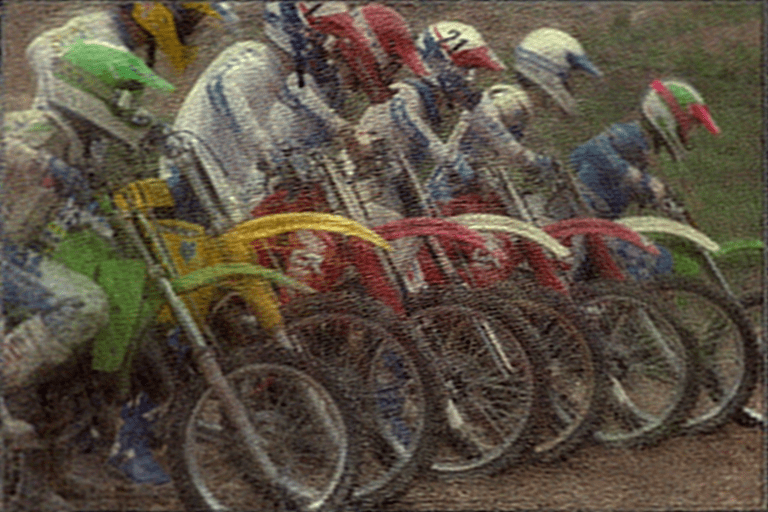} & \includegraphics[width=0.24\textwidth]{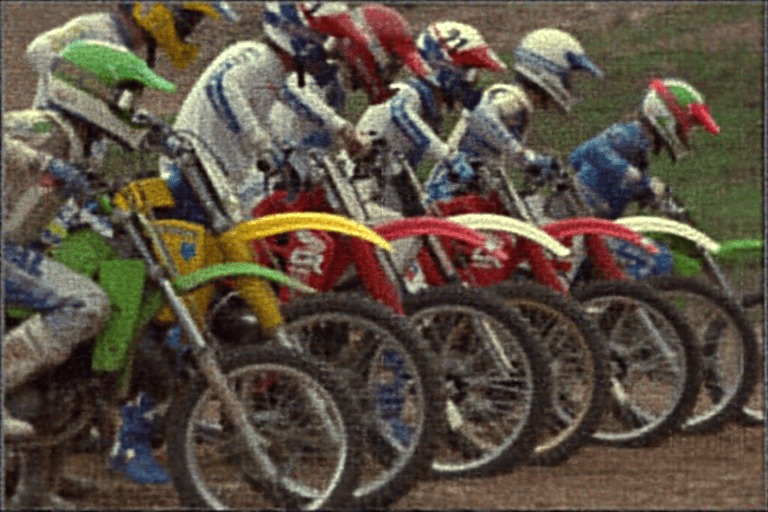}\\
  \end{tabular}
  \captionof{figure}{[Multi image super resolution] Figures depict the results of 4$\times$ super resolution from 4 images captured with varying subpixel shifts and rotations  for all considered methods. The numbers above the figures report PSNR and SSIM (in the parenthesis).}
  \label{tab:MISR_results}
\end{table}

\begin{table}[!h]
  \centering
  \setlength{\tabcolsep}{1pt} 
  \begin{tabular}{cccccccc}
    Ground Truth & Bicubic ($\times 4$) & NestNet  & WIRE & SIREN  & Gaussian & MFN & FFN \\
    \includegraphics[width=0.11\textwidth]{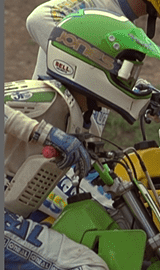}
     & 
    \includegraphics[width=0.11\textwidth]{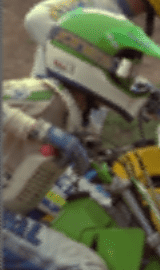}
     &
    \includegraphics[width=0.11\textwidth]{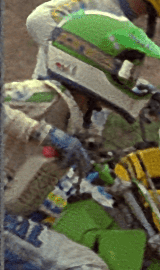}
     &
    \includegraphics[width=0.11\textwidth]{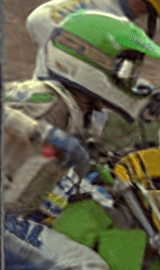} &
    \includegraphics[width=0.11\textwidth]{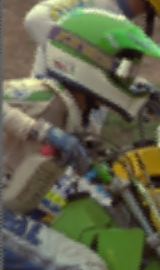}
     & 
    \includegraphics[width=0.11\textwidth]{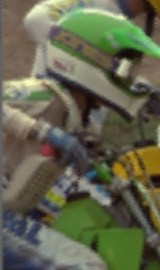}
     & 
    \includegraphics[width=0.11\textwidth]{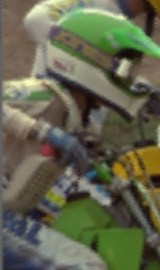}
    & 
    \includegraphics[width=0.11\textwidth]{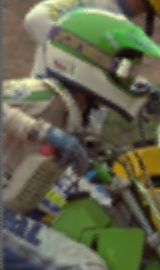}
  \end{tabular}
  \captionof{figure}{[Multi image super resolution] Figures depict the zoomed-in comparison for all considered methods.}
  \label{tab:MISR_results_app}
\end{table}

\subsubsection{Image-denoising}
The next type of inverse problem is image-denoising. We choose an image from the Kodak dataset and inject noise by sampling photon noise from an independently distributed Poisson distribution at each pixel with a maximum mean photon count of 30. The ground truth and the noisy image are shown in Figure~\ref{tab:img_denoise_results}. For all methods, we train INRs for 2000 epochs with the initial learning rate 0.005. We set $s_0=5.0$ for Gaussian, $\omega_0=5$ for SIREN, and $s_0=5$, $\omega_0=5$ for WIRE.

Figure~\ref{tab:img_denoise_results} also shows results of denoising performed by each INRs and NestNet produces improved accuracy in the denoised image, again outperforming all considered baselines (i.e., the highest PSNR and SSIM). Qualitatively, the NestNet reconstruction presents the most accurate fine details and less blurry presentation of the image.

\begin{table}[h]
  \centering
  \setlength{\tabcolsep}{1pt}
  \begin{tabular}{cccc}
    Ground Truth & Noisy image & NestNet (\textbf{25.6177}, \textbf{0.80}) & WIRE (25.3556, 0.78)\\
    \includegraphics[width=0.24\textwidth]{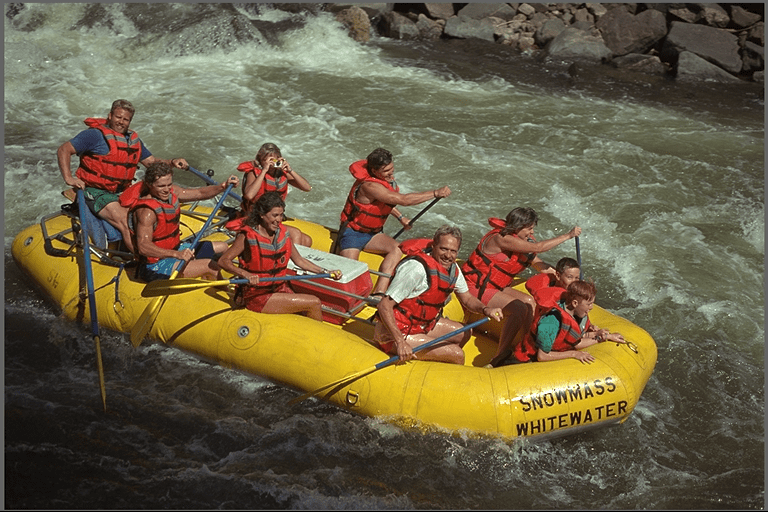} &
    \includegraphics[width=0.24\textwidth]{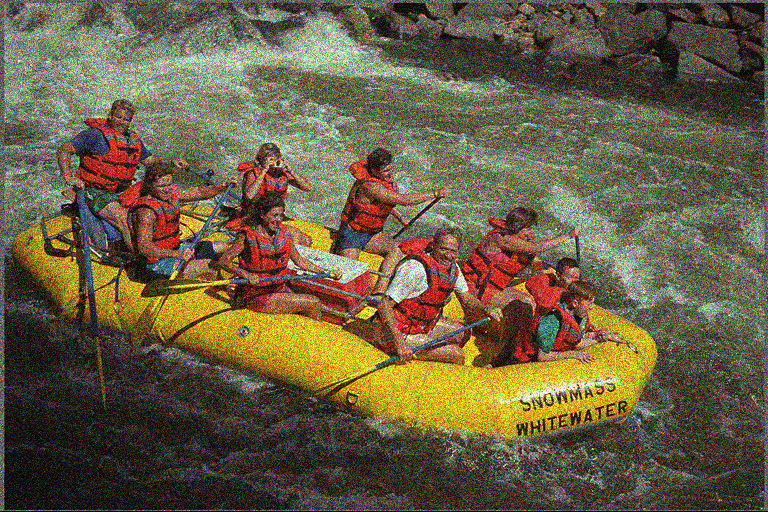} &
    \includegraphics[width=0.24\textwidth]{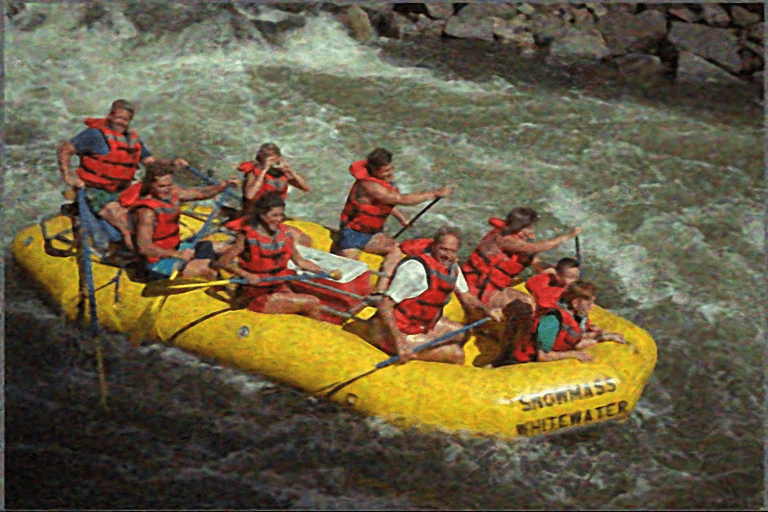} & \includegraphics[width=0.24\textwidth]{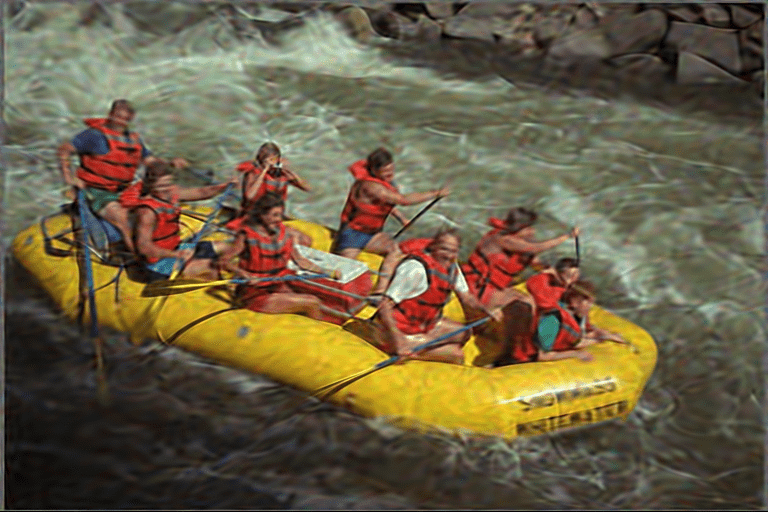} \\

    SIREN (20.8582, 0.60) & Gaussian (21.7875, 0.63)& MFN (24.9448, 0.77) & FFN (23.9626, 0.73)\\
    \includegraphics[width=0.24\textwidth]{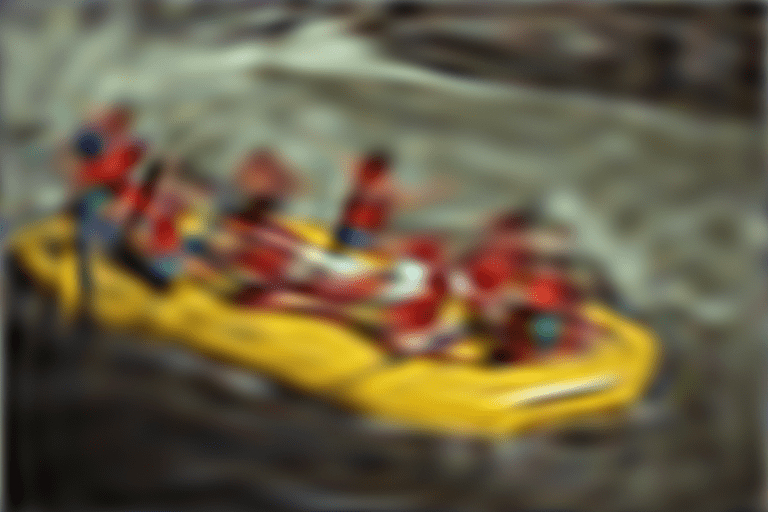} & 
    \includegraphics[width=0.24\textwidth]{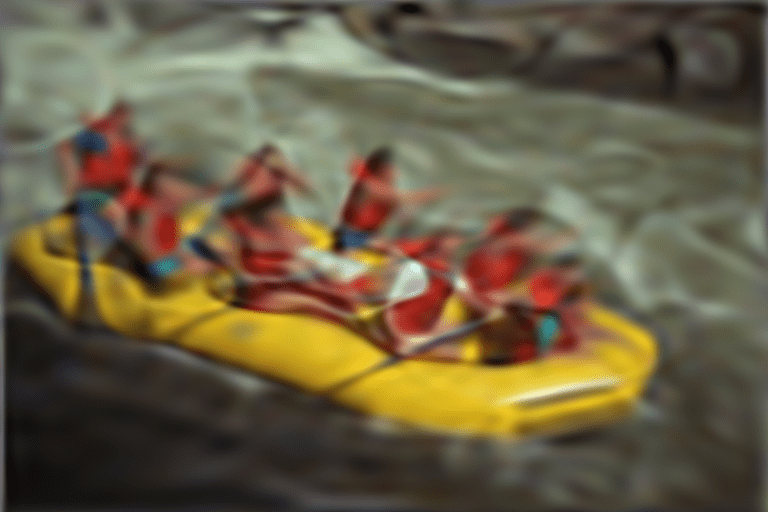} & \includegraphics[width=0.24\textwidth]{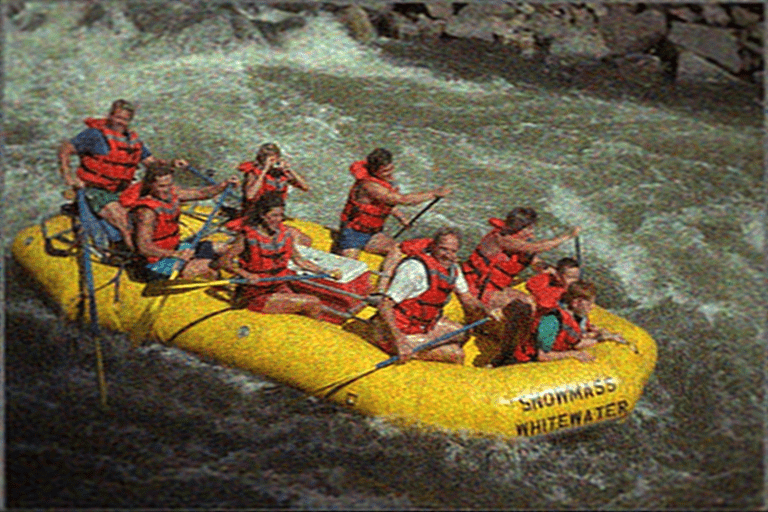} & \includegraphics[width=0.24\textwidth]{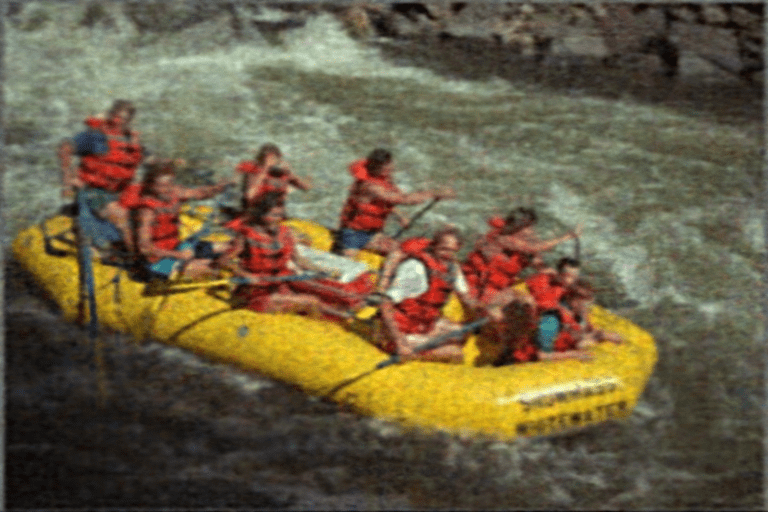}\\
    
  \end{tabular}
  \captionof{figure}{[Image denoising] The numbers above the figures report PSNR and SSIM (in the parenthesis).}
  \label{tab:img_denoise_results}
\end{table}

Figure~\ref{tab:img_denoise_results_app} presents zoomed-in plots for the image denoising task. NestNet has outperformed other methods in preserving sharp edges and fine details, maintaining color fidelity, and reducing noise without creating artifacts. Although WIRE, SIREN, and Gaussian successfully handled noise, they have excessively smoothed out finer textures, resulting in the loss of subtle textural details in less prominent areas, as evident in the zoomed-in images of the text on the front of the boat and the man’s facial features. MFN, although slightly better at retaining finer details, lacks in denoising ability.

\begin{table}[!h]
  \centering
  \setlength{\tabcolsep}{1pt} 
  \begin{tabular}{cccccccc}
    Ground Truth & Noisy image & NestNet  & WIRE  & SIREN  & Gaussian & MFN & FFN \\
    \includegraphics[width=0.12\textwidth]{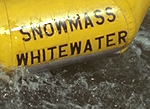}
     & 
    \includegraphics[width=0.12\textwidth]{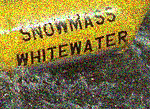}
     &
    \includegraphics[width=0.12\textwidth]{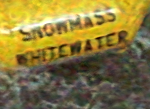}
     &
    \includegraphics[width=0.12\textwidth]{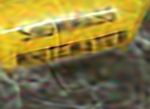} &
     
    \includegraphics[width=0.12\textwidth]{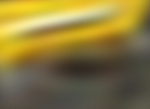}
     & 
    \includegraphics[width=0.12\textwidth]{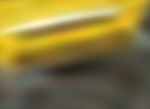}
     & 
    \includegraphics[width=0.12\textwidth]{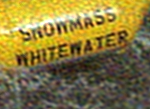}
    & 
    \includegraphics[width=0.12\textwidth]{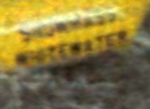}\\
    
    Ground Truth & Noisy image & NestNet  & WIRE  & SIREN  & Gaussian & MFN & FFN \\
    \includegraphics[width=0.12\textwidth]{figs/Denoise_GT_part_two.png}   & 
    \includegraphics[width=0.12\textwidth]{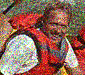} &
    \includegraphics[width=0.12\textwidth]{figs/Denoise_nestmlp_0__part_two.png} &
    \includegraphics[width=0.12\textwidth]{figs/Denoise_wire_3__orignial_part_two.png} & 
    \includegraphics[width=0.12\textwidth]{figs/Denoise_siren_1_part_two.png} &
    \includegraphics[width=0.12\textwidth]{figs/Denoise_gauss_0_part_two.png} &
    \includegraphics[width=0.12\textwidth]{figs/Denoise_mfn_0_part_two.png} &
    \includegraphics[width=0.12\textwidth]{figs/Denoise_posenc_4_part_two.png} 
  \end{tabular}
  \captionof{figure}{[Image denoising] Figures depict the zoomed-in comparison for all considered methods.}
  \label{tab:img_denoise_results_app}
\end{table}

\subsubsection{CT reconstruction}
As a last task for inverse problems, we test NestNets on computed tomography (CT) reconstruction. Again, by using the procedure considered in \citet{saragadam2023wire}, we generate 100 CT measurements with 100 different angles of 256$\times$256 x-ray colorectal images \citep{clark2013cancer}. For all methods, we train INRs for 5000 epochs with the initial learning rate 0.005. We set $s_0=10$ for Gaussian, $\omega_0=10$ for SIREN, and $s_0=10$, $\omega_0=10$ for WIRE.

Figure~\ref{tab:ct_results} shows the ground-truth CT image and reconstructed images via INRs. The reconstruction via NestNet achieves the highest accuracy again both in PSNR and SSIM. Both in PSNR and SSIM, the NestNet achieve better performance than the best results reported in \citet{saragadam2023wire} (i.e., 32.3 dB and 0.81); there is a remarkable jump in the SSIM (from 0.81 or 0.84 to 0.93). Qualitatively, NestNet presents the most sharp segmentation of objects and presents minimal ringing artifacts compared to the baselines. 

\begin{table}[h]
  \centering
  \setlength{\tabcolsep}{1pt}

  \begin{tabular}{cccc}
    Ground Truth &  NestNet (\textbf{32.3745}, \textbf{0.93}) & WIRE (29.3886, 0.84) & SIREN (27.5518,  0.84)\\
    \includegraphics[width=0.24\textwidth]{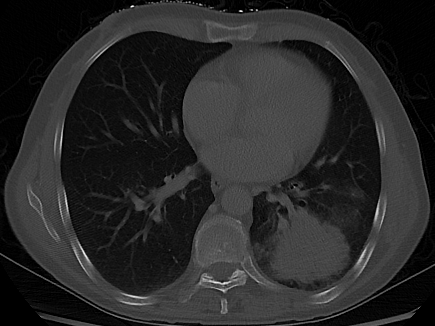} & 
    \includegraphics[width=0.24\textwidth]{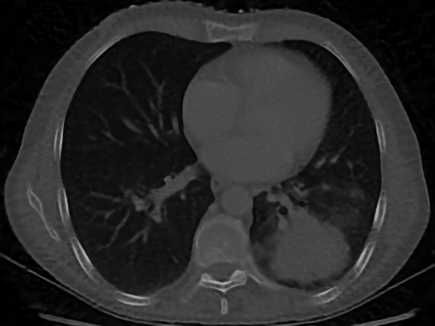} & \includegraphics[width=0.24\textwidth]{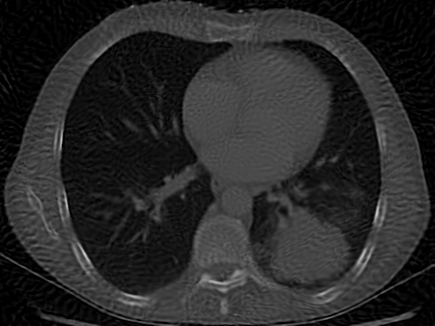} & \includegraphics[width=0.24\textwidth]{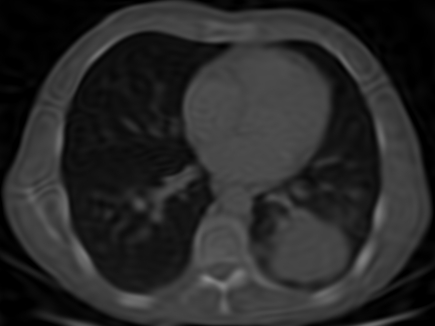}\\
    
    Gaussian (27.9947, 0.85) &  MFN (25.6977, 0.66) & FFN (26.9803, 0.82)\\
    \includegraphics[width=0.24\textwidth]{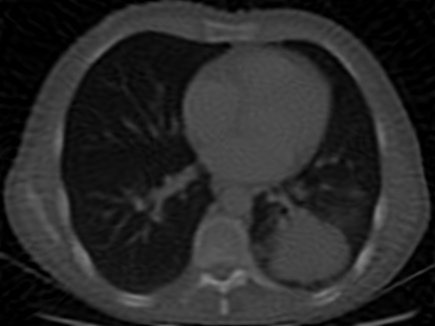} & 
    \includegraphics[width=0.24\textwidth]{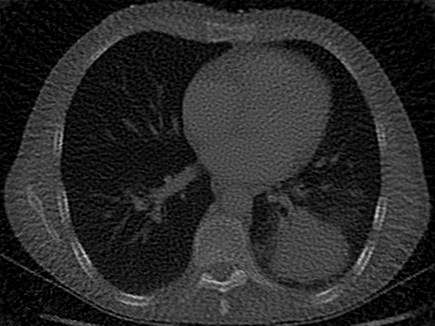} & 
    \includegraphics[width=0.24\textwidth]{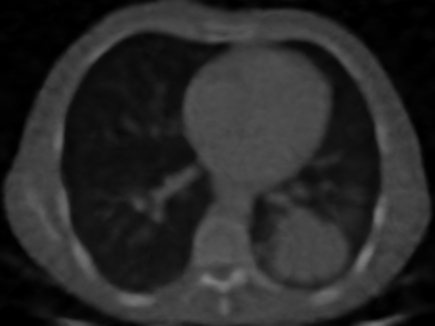}
  \end{tabular}

  \captionof{figure}{[CT Reconstruction] Figures depict the results of CT-based reconstruction with 100 angles for a 256 × 256 image for all considered methods. The numbers above the figures report PSNR and SSIM (in the parenthesis).}
  \label{tab:ct_results}
\end{table}

Figure~\ref{tab:ct_results_app} displays zoomed-in plots of CT reconstructions, highlighting variations in reconstruction quality. NestNet excels in preserving the integrity of pulmonary textures and spinal details, delivering high fidelity representations akin to actual scans. In contrast, models like WIRE and Gaussian, though partially successful, tend to obscure finer details and introduce noise artifacts that can mask diagnostic features. Meanwhile, SIREN and FFN, while avoiding noise artifacts, still produce somewhat blurred images, resulting in a loss of essential subtleties within the vertebral bodies and lung peripheries.

\begin{table}[!h]
  \centering
  \setlength{\tabcolsep}{1pt} 
  \begin{tabular}{ccccccc}
    Ground Truth & NestNet & WIRE & SIREN  & Gaussian & MFN & FFN \\ 
    \includegraphics[width=0.12\textwidth]{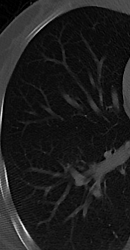}
     & 
    \includegraphics[width=0.12\textwidth]{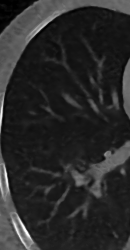}
     &
    \includegraphics[width=0.12\textwidth]{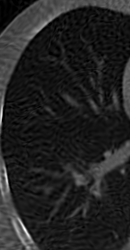}
     &
    \includegraphics[width=0.12\textwidth]{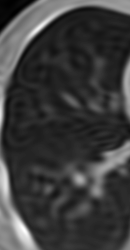} 
    & \includegraphics[width=0.12\textwidth]{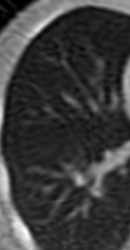}
     & 
    \includegraphics[width=0.12\textwidth]{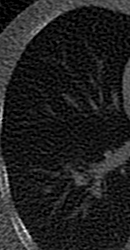}
     & 
    \includegraphics[width=0.12\textwidth]{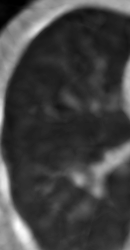}\\
  \end{tabular}
  \captionof{figure}{[CT Reconstruction] Figures depict the zoomed-in comparison for all considered methods.}
  \label{tab:ct_results_app}
\end{table}

\subsection{PDE Solution Approximation}
The solution approximated by using NestNet is evidently shown to be much
more accurate than those of the baselines and the accuracy
of the solution is further supported by the diverse error metrics shown in Table~\ref{tab:pinn_error}; the NestNet solution are an order
of magnitude accurate than the baselines and the explained
variance of NestNet is almost close to 1.

\begin{table}[t]
    \caption{Solution accuracy measured using various metrics.}
    \label{tab:pinn_error}
    \centering
    \setlength{\tabcolsep}{4pt}  
    \renewcommand{\arraystretch}{1.2} 
    \begin{tabular}{l|c c c}
        \hline
        & \textbf{Abs. Err.} & \textbf{Rel. Err.} & \textbf{Exp. Var.} \\
        \hline
        PINN   & 0.2015  & 0.3719  & 0.6147 \\
        SIREN  & 0.2367  & 0.4031  & 0.5564 \\
        FFN    & 0.5101  & 0.5918  & 0.0000 \\
        WIRE   & 0.4447  & 0.5525  & 0.1392 \\
        \hline
        NestNet  & \textbf{0.0017}  & \textbf{0.0342}  & \textbf{0.9967} \\
        \hline
    \end{tabular}
\end{table}

\section{Additional Experimentation}\label{app:extra_exps}
\begin{figure}[h]
    \centering
    \begin{subfigure}{0.49\columnwidth}
    \includegraphics[width=1\columnwidth]{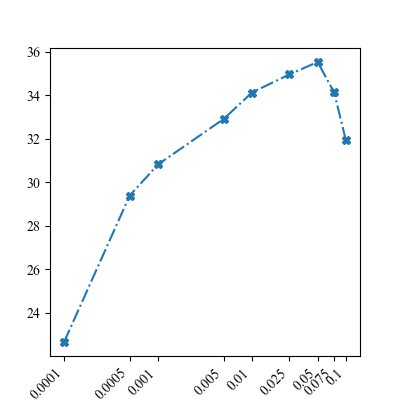}
    \caption{PSNR versus learning rate.}
    \label{fig:psnr_lr}
    \end{subfigure}
    \begin{subfigure}{0.49\columnwidth}
    \includegraphics[width=1\columnwidth]{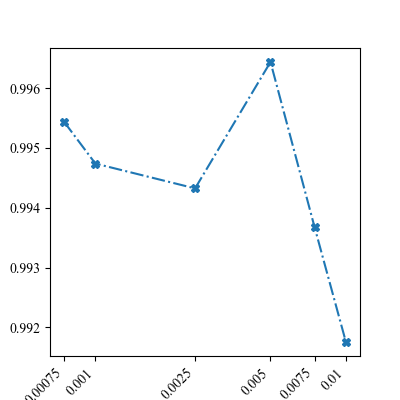}
    \caption{IOU versus learning rate.}
    \label{fig:iou_lr}
    \end{subfigure}
    \caption{[Signal representation] PSNR (left) and IOU (right) obtained via employing various learning rates for the image representation and occupancy volume representation tasks.}
    \label{fig:lr_study}
\end{figure}

\paragraph{Learning rate study for the image representation and the occupancy volume representation tasks.} Figure~\ref{fig:psnr_lr} shows the PSNR achieved by employing different learning rates.
For the image representation task, with the higher learning rate (e.g., [0.025,0.075]), the method is capable of achieving PSNRs higher than 34 and 35.5249 at the learning rate 0.05, which is more than 4 dB higher than that of the second best performing baseline (i.e,. WIRE). 

Figure~\ref{fig:iou_lr} reports the IOU of INRs trained with different learning rates. For all considered learning rates sampled from a wide range [0.0075,0.01], NestNets produce INRs with very high IOU values (over 0.992 for all considered learning rates).

\paragraph{Different scales for super-resolution tasks}

Table~\ref{tab:SISR_results_scale} further compares the performance of single-image super-resolution on varying downsampling rates, 1/2, 1/4 and 1/6 (denoted by $\times 2$, $\times 4$, and $\times 6$, respectively). Again, NestNet achieves the highest performance in all scales by producing images with PSNR that is $\sim$1 dB higher than the second best baseline in PSNR and notable increase in SSIM. 

\begin{table}[!h]
    \centering
    \begin{minipage}[t]{0.49\textwidth}
        \centering
        \setlength{\tabcolsep}{1.3pt}  
        \renewcommand{\arraystretch}{1.2} 
        \captionsetup{type=table}
        \caption{[Single image super resolution] The results of super resolution for all considered methods. The numbers above the figures report PSNR and SSIM (in the parenthesis).}
        \label{tab:SISR_results_scale}
        \begin{tabular}{l|c|c|c}
             & $\times 2$ & $\times 4$ & $\times 6$ \\
             \hline
            SIREN     & 24.97 (0.79) & 24.68 (0.79) & 24.21 (0.77) \\
            Gaussian  & 24.51 (0.79) & 24.28 (0.78) & 23.76 (0.77) \\
            MFN       & 29.78 (0.91) & 24.05 (0.69) & 20.17 (0.47) \\
            FFN       & 26.96 (0.84) & 24.92 (0.77) & 22.18 (0.65) \\
            WIRE      & 28.87 (0.89) & 26.92 (0.85) & 24.44 (0.77) \\
            \hline
            NestNet   & \textbf{29.42} (\textbf{0.90}) & \textbf{27.53} (\textbf{0.87}) & \textbf{25.22} (\textbf{0.81}) \\
            \hline
        \end{tabular}
        
    \end{minipage}%
    \hfill
    \begin{minipage}[t]{0.49\textwidth}
        \centering
        \setlength{\tabcolsep}{1.3pt}  
        \renewcommand{\arraystretch}{1.2} 
        \captionsetup{type=table}
        \caption{[Multi image super resolution] The results of super resolution for all considered methods. The numbers above the figures report PSNR and SSIM (in the parenthesis).}
        \label{tab:MISR_results_scale}
        \begin{tabular}{l|c|c|c}
             & $\times 2$ & $\times 4$ & $\times 8$ \\
             \hline
            SIREN     & 21.71 (0.68) & 21.75 (0.69) & 21.45 (0.67) \\
            Gaussian  & 20.40 (0.59) & 21.03 (0.64) & 21.36 (0.66) \\
            MFN       & 24.55 (0.84) & 19.78 (0.63) & 18.60 (0.49) \\
            FFN       & 22.28 (0.71) & 21.78 (0.68) & 21.10 (0.64) \\
            WIRE      & 23.01 (0.75) & 23.21 (0.78) & 21.77 (0.68) \\
            \hline
            NestNet   & \textbf{25.65} (\textbf{0.86}) & \textbf{24.32} (\textbf{0.82}) & \textbf{22.33} (\textbf{0.72}) \\
            \hline
        \end{tabular}
        
    \end{minipage}
\end{table}

Table~\ref{tab:MISR_results_scale} further compares the performance of multi-image super-resolution on varying downsampling rates, 1/2, 1/4 and 1/8 (denoted by $\times 2$, $\times 4$, and $\times 8$, respectively). NestNet achieves the highest performance in all scales by producing images with PSNR that is $\sim$1 dB higher than the second best baseline in PSNR and again some notable increase in SSIM.

\end{document}